\pdfoutput=1

\documentclass{article}
\usepackage{template_files/jfrExamplee}
\usepackage{graphicx}
\usepackage{setspace}
\usepackage{amsmath}
\usepackage{amssymb}
\usepackage[dvipsnames]{xcolor}
\usepackage{colortbl}
\usepackage{multirow}
\usepackage{paralist}
\usepackage{enumitem}
\usepackage{hyperref}
\hypersetup{
    colorlinks = true,
    allcolors = {blue},
}

\usepackage{subcaption}
\usepackage{algorithm,algorithmic}
\usepackage{titlesec}

\usepackage{tocloft}

\usepackage[framemethod=tikz]{mdframed}
\usepackage{lipsum}

\definecolor{mycolor}{rgb}{0.122, 0.435, 0.698}

\newmdenv[innerlinewidth=0.5pt, roundcorner=4pt,linecolor=mycolor,innerleftmargin=6pt,
innerrightmargin=6pt,innertopmargin=6pt,innerbottommargin=6pt]{mybox}

\usepackage{tikzscale}
\usepackage{pgfplots}
\pgfplotsset{compat=1.16}
\usepackage{tikz}

\usepackage[smaller,nohyperlinks]{acronym}

\newcommand{\comb}[1]{{\color{blue}#1}}

\newcommand{\ph}[1]{{\textbf{#1}:}}
\newcommand{\pr}[1]{\textbf{#1:}} %

\newcommand{\rev}[1]{#1}
\newcommand{\revv}[1]{#1}

\newcommand{\argmax}{\mathop{\mathrm{argmax}}}

\graphicspath{{../}{./}{./figures/}}

\title{\large NeBula: Quest for Robotic Autonomy in Challenging Environments;
\\ An Overview of TEAM CoSTAR's Solution \\at Phase I and II of DARPA Subterranean Challenge}

\author{
\small \normalfont
    \parbox{\linewidth}{\centering
Ali Agha\textsuperscript{1}, 
Kyohei Otsu\textsuperscript{1}, 
Benjamin Morrell\textsuperscript{1}, 
David D. Fan\textsuperscript{1}, 
Rohan Thakker\textsuperscript{1}, 
Angel Santamaria-Navarro\textsuperscript{1}, 
Sung-Kyun Kim\textsuperscript{1}, 
Amanda Bouman\textsuperscript{2}, 
Xianmei Lei\textsuperscript{1}, 
Jeffrey Edlund\textsuperscript{1}, 
Muhammad Fadhil Ginting\textsuperscript{1}, 
Kamak Ebadi\textsuperscript{1}, 
Matthew Anderson\textsuperscript{1}, 
Torkom Pailevanian\textsuperscript{1}, 
Edward Terry\textsuperscript{1}, 
Michael Wolf\textsuperscript{1}, 
Andrea Tagliabue\textsuperscript{3}, 
Tiago Stegun Vaquero\textsuperscript{1}, 
Matteo Palieri\textsuperscript{1,10}, 
Scott Tepsuporn\textsuperscript{1}, 
Yun Chang\textsuperscript{3}, 
Arash Kalantari\textsuperscript{1}, 
Fernando Chavez\textsuperscript{1}, 
Brett Lopez\textsuperscript{1}, 
Nobuhiro Funabiki\textsuperscript{1}, 
Gregory Miles\textsuperscript{1}, 
Thomas Touma\textsuperscript{1}, 
Alessandro Buscicchio\textsuperscript{1,10}, 
Jesus Tordesillas\textsuperscript{1}, 
Nikhilesh Alatur\textsuperscript{1}, 
Jeremy Nash\textsuperscript{1}, 
William Walsh\textsuperscript{1}, 
Sunggoo Jung\textsuperscript{7}, 
Hanseob Lee\textsuperscript{7}, 
Christoforos Kanellakis\textsuperscript{8}, 
John Mayo\textsuperscript{1}, 
Scott Harper\textsuperscript{1}, 
Marcel Kaufmann\textsuperscript{1}, 
Anushri Dixit\textsuperscript{2}, 
Gustavo J. Correa\textsuperscript{9}, 
Carlyn Lee\textsuperscript{1}, 
Jay Gao\textsuperscript{1}, 
Gene Merewether\textsuperscript{1}, 
Jairo Maldonado-Contreras\textsuperscript{1}, 
Gautam Salhotra\textsuperscript{4}, 
Maira Saboia Da Silva\textsuperscript{1}, 
Benjamin Ramtoula\textsuperscript{1}, 
Seyed Fakoorian\textsuperscript{1}, 
Alexander Hatteland\textsuperscript{1}, 
Taeyeon Kim\textsuperscript{1}, 
Tara Bartlett\textsuperscript{1}, 
Alex Stephens\textsuperscript{1}, 
Leon Kim\textsuperscript{1}, 
Chuck Bergh\textsuperscript{1}, 
Eric Heiden\textsuperscript{4}, 
Thomas Lew\textsuperscript{1}, 
Abhishek Cauligi\textsuperscript{1}, 
Tristan Heywood\textsuperscript{1}, 
Andrew Kramer\textsuperscript{5}, 
Henry A. Leopold\textsuperscript{1}, 
Hov Melikyan\textsuperscript{1},
Hyungho Chris Choi\textsuperscript{7},
Shreyansh Daftry \textsuperscript{1},
Olivier Toupet\textsuperscript{1}, 
Inhwan Wee\textsuperscript{7}, 
Abhishek Thakur\textsuperscript{1}, 
Micah Feras\textsuperscript{1}, 
Giovanni Beltrame\textsuperscript{6}, 
George Nikolakopoulos\textsuperscript{8}, 
David Shim\textsuperscript{7}, 
Luca Carlone\textsuperscript{3}, 
Joel Burdick\textsuperscript{2} %
    }%
  \\\\
\textsuperscript{1} NASA Jet Propulsion Laboratory, California Institute of Technology\\
\textsuperscript{2} California Institute of Technology\\
\textsuperscript{3} Massachusetts Institute of Technology\\
\textsuperscript{4} University of Southern California \\
\textsuperscript{5} University of Colorado Boulder \\
\textsuperscript{6} École Polytechnique de Montréal \\
\textsuperscript{7} Korea Advanced Institute of Science and Technology \\
\textsuperscript{8} Luleå University of Technology \\
\textsuperscript{9} University of California, Riverside \\
\textsuperscript{10} Polytechnic University of Bari \\
}

\begin{document}

\maketitle
\begin{abstract}
This paper presents and discusses algorithms, hardware, and software architecture developed by the TEAM CoSTAR (Collaborative SubTerranean Autonomous Robots), competing in the DARPA Subterranean Challenge. Specifically, it presents the techniques utilized within the Tunnel (2019) and Urban (2020) competitions, where CoSTAR achieved 2nd and 1st place, respectively. We also discuss CoSTAR's demonstrations in Martian-analog surface and subsurface (lava tubes) exploration. The paper introduces our autonomy solution, referred to as NeBula (Networked Belief-aware Perceptual Autonomy). NeBula is an uncertainty-aware framework that aims at enabling resilient and modular autonomy solutions by performing reasoning and decision making in the belief space (space of probability distributions over the robot and world states). We discuss various components of the NeBula framework, including: (i) geometric and semantic environment mapping; (ii) a multi-modal positioning system; (iii) traversability analysis and local planning; (iv) global motion planning and exploration behavior; (v) risk-aware mission planning; (vi) networking and decentralized reasoning; and (vii) learning-enabled adaptation. We discuss the performance of NeBula on several robot types (e.g. wheeled, legged, flying), in various environments. We discuss the specific results and lessons learned from fielding this solution in the challenging courses of the DARPA Subterranean Challenge competition. 
\end{abstract}

\newpage
\tableofcontents

\clearpage
\section{Introduction}
Robotics and Artificial Intelligence (AI) are transforming our lives, with a growing number of robotic applications ranging from self-driving cars~\cite{yurtsever2020survey}, search \& rescue~\cite{balta2017integrated}, healthcare~\cite{qin2020temporal} and humanitarian missions~\cite{santana2007sustainable}, to robots under water~\cite{kinsey2006survey} and robots beyond our home planet on Mars~\cite{sasaki2020map,autonomyformarsrovers} and the moon~\cite{fordlunarpits}. Autonomy and AI are empowering these robots to carry out missions autonomously, increasing efficiency with reduced human risk, saving lives, and accomplishing tasks that are often in hazardous environments too dangerous for human.

\begin{figure}
    \centering
    \subfloat[]{\includegraphics[height=0.3\textwidth]{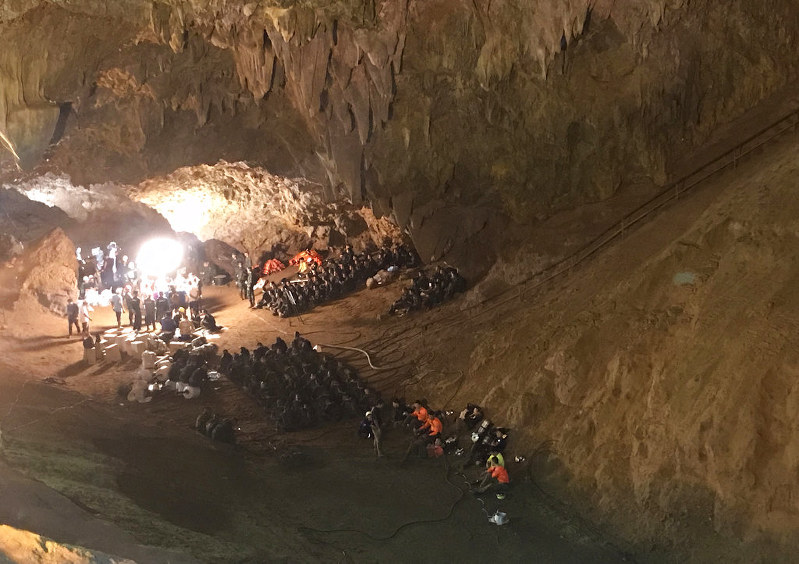}}
    \quad
    \subfloat[]{\includegraphics[height=0.3\textwidth]{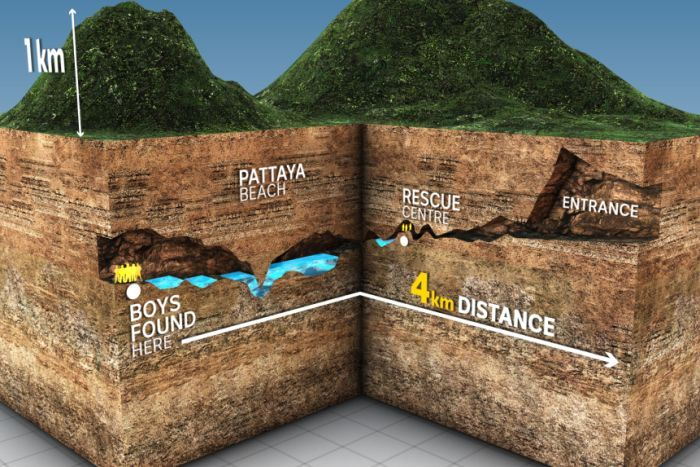}}
    \caption[Tham Luang cave rescue mission]{Tham Luang cave rescue mission. Figures from \cite{thaicave_pic} and \cite{thaicave_schematic}.}
    \label{fig:IT_tham_luang_cave}
\end{figure}

\ph{Extreme environments} Underground environments are an important example of the type of terrain that imposes a lot of risk for humans, with a wide range of terrestrial and planetary applications. On Earth, autonomous underground exploration is a crucial capability in search and rescue missions after natural disasters, in mining, oil and gas industry, and in supporting spelunkers and cave rescue missions. One prominent example is the Tham Luang cave rescue~(\autoref{fig:IT_tham_luang_cave}), where the international community aimed at rescuing thirteen members of a football team from 4 km inside a partially flooded cave. Drones equipped with thermal cameras have been flown over Tham Luang to detect possible access points, and an underwater robot was deployed to send information back on the water depth and condition of the cave. However, at that time, no technology existed to autonomously reach the people, map the cave, and scan for people deep underground.


\ph{Planetary applications} Beyond our home planet, the research community has identified more than 200 lunar and 2000 Martian cave-related features~\cite{cushing2012marscaves}. Caves and subsurface voids, in general, are of utmost importance in space exploration for several reasons. First, their stable, radiation-shielding environment, and potential to act as volatile traps makes them an ideal habitat candidate for future human exploration~\cite{kesner2007mobility,timothy2010cavesInSolarSystem}. Second, planetary cave environments may harbor life due to their shielding from cosmic rays, and if there is life beyond Earth, deep planetary caves are one of the most likely places to find it. Third, exploring caves provides an unprecedented opportunity for scientists to study planetary volcanic processes and the geological history of planetary bodies. These reasons, among many others, have made subsurface exploration one of the main next frontiers for space exploration~\cite{stamenkovic2019next,thomas-agu}.

While autonomy and AI technologies are growing fast, challenges still remain for operations in extreme environments. Technical challenges include: perceptual degradation (darkness, obscurants, self-similarity, limited textures) in wholly unknown and unstructured environments; extreme terrain that tests the limits of traversability; mission execution under constrained resources; and high-risk operations where robot failure or component degradation is a real possibility. Most challenging of all, however is the combination of the above features. Further work is needed to push the state-of-the-art to enable systems that can robustly and consistently address these challenges simultaneously.

\ph{Contributions} In this paper, we discuss the NeBula autonomy solution and Team CoSTAR’s contributions towards addressing some of the challenges in robotic exploration of unknown extreme surface and subsurface environments. We discuss these technologies in the context of the DARPA Subterranean challenge~\cite{subt_webpage}, where Team CoSTAR won the Urban competition and ranked 2nd in the Tunnel competition. The videos in ~\cite{youtubeurbanandtunnelvideo,youtubespotpaper,youtubeAutonomousCaveExploration,youtubeBDLavaTube, youtubeMarsDog} depicts some highlights of these runs. As we will discuss in~\autoref{sec:subt_intro}, this competition pushes the state-of-the-art boundaries in extreme environment exploration in mobility, perception, autonomy, and networking. Specifically, we will discuss Team CoSTAR’s contributions in advancing the autonomy in the following fronts:
\begin{enumerate}[topsep=0pt,itemsep=-1ex,partopsep=1ex,parsep=1ex]
    \item Resilient, modular, and uncertainty-aware autonomy architecture (\autoref{sec:nebula_architechture})
    \item State estimation in perceptually-degraded environments (\autoref{sec:state_estimation})
    \item Large scale positioning and 3D mapping (\autoref{sec:lamp})
    \item Semantic understanding and artifact detection (\autoref{sec:artifacts})
    \item Risk-aware traversability analysis (\autoref{sec:traversability})
    \item Global motion planning and coverage/search behaviors (\autoref{sec:global_planning})
    \item Multi-robot networking (\autoref{sec:multirobot_networking})
    \item Mission planning and system health management (\autoref{sec:mission_planning})
    \item Mobility systems and hardware integration (\autoref{sec:hardware}).
\end{enumerate}

\section{DARPA Subterranean Challenge} \label{sec:subt_intro}
\begin{figure}[!htbp]
    \centering
    \includegraphics[width=0.7\columnwidth]{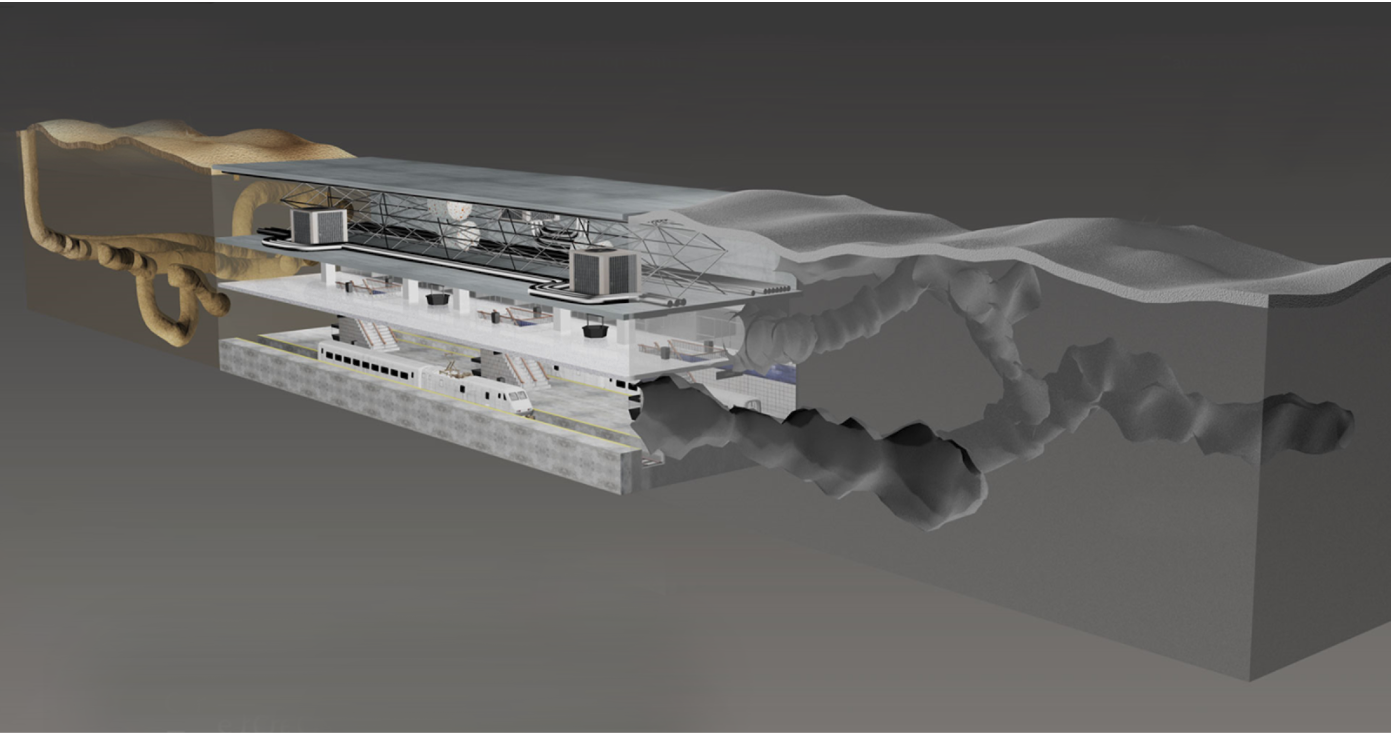}
    \caption{The three subdomains in the DARPA Subterranean Challenge: Tunnel systems, urban underground, and cave networks}
    \label{fig:IS}
\end{figure}

The DARPA Subterranean or “SubT” Challenge \cite{subt_webpage} is a robotic competition that seeks novel approaches to rapidly map, navigate, and search underground environments  (\autoref{fig:IS}). The competition spans a period of three years. The teams participating in the systems track develop and implement physical systems for autonomous traversal, mapping, and search in various subterranean environments, \rev{including mines, industrial complexes, and natural caves.}

\ph{Illustrative scenarios} The primary scenario of interest for the competition is providing autonomous and rapid situational awareness in unknown and challenging subterranean environments. The layout of the environment is unknown, could degrade or change over time (i.e., dynamic terrain), and is either impossible or too high-risk to send in personnel. Potential representative scenarios range from planetary cave exploration to rescue efforts in collapsed mines, post-earthquake search and rescue missions in urban underground settings, and cave rescue operations for injured or lost spelunkers. Additional scenarios include missions where teams of autonomous robotic systems are sent to perform rapid search and mapping in support of follow-on operations in advance of service experts, such as astronauts and search/rescue personnel. These scenarios present significant challenges and dangers that would preclude employing a human team, such as collapsed and unstable structures or debris, a presence of hazardous materials, lack of ventilation, and potential for smoke and/or fire.

\subsection{Competition Rules}
\ph{Competition structure} Each team has a fixed time window (1 hour) to carry out the mission. Each team deploys their robots to provide rapid situational awareness through mapping of the unknown environment and localization of artifacts (e.g., CO$_2$ gas source, survivors, electrical boxes). As the systems explore the environment, these situational awareness updates need to be communicated back to a base station, outside the challenging area, in as close to real-time as possible. The urgency in completing the course objectives and providing near-real-time situational awareness updates is a consistent focus of the competition.

\begin{figure}[!htbp]
    \centering
    \includegraphics[width=0.95\columnwidth]{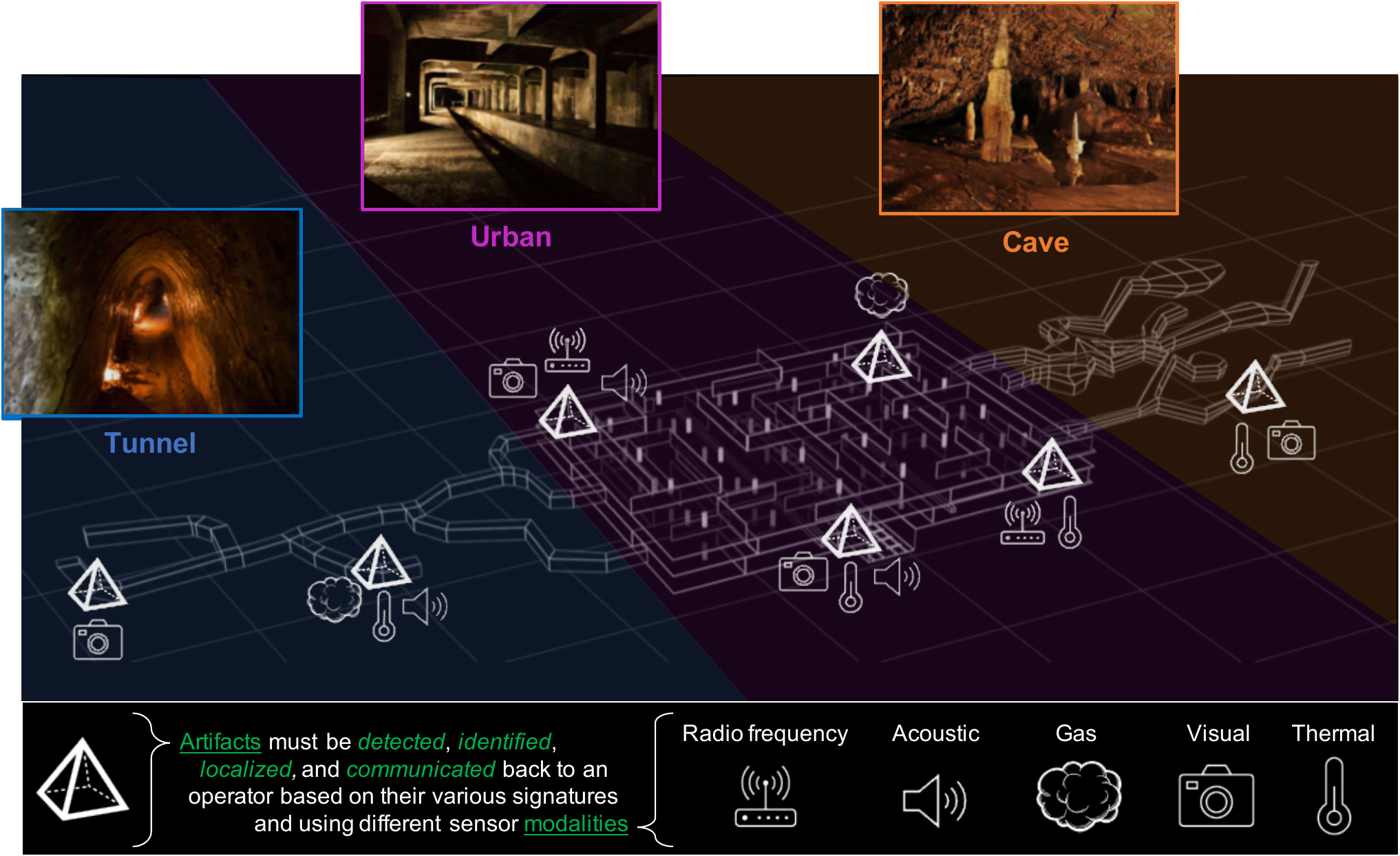}
    \caption{\rev{The final event will comprise elements from all subdomains in a course to demonstrate the versatility of solutions developed. To motivate the efficient exploration and search of the environment, competitors score points by accurately reporting the type and position of artifacts distributed along the course.}}
    \label{fig:IR}
\end{figure}

\ph{Scoring} The detailed scoring metrics are discussed in \cite{scoring_metric}. At a high-level, each team gets 1 point per artifact, if the team can (i) reach, detect, and recognize the artifact, (ii) localize the object in global coordinates with less than 5 meters error, and (iii) report the object to outside the course during the 1-hour mission period. There are 20 artifacts distributed throughout the course, see \autoref{fig:IR}, and each team has only \rev{25} chances to report the artifacts. Hence, teams need to be careful in using their reporting budget. In addition to artifacts, the quality of a 3D map, environment coverage, and the traverse length are quantified as interim measures of success.

\ph{High-level rules} There is no prior map of the environment. No team member is allowed to enter the course prior or during the competition. The mission length is limited to an hour. There is one single human supervisor outside the course, who is allowed to see the data coming from the robot and interact with them, if and when a communication link is established. Communication between the operator and robots is typically limited to the areas near the course entrance, due to the scale, complexity, and communication-denied nature of the course.

\subsection{Technical Challenge Elements} \label{subsec:subt_intro_challenge}
The competition is intended to push the boundaries of the state-of-the-art and state-of-the-practice across various challenge elements, including: austere navigation, degraded sensing, severe communication constraints, terrain obstacles, and endurance limits.

\ph{All-terrain Traversability} The environment includes multiple levels, loops, dead-ends, slip-inducing terrain interfaces, and a range of features and obstacles that challenge robot’s mobility. Examples of terrain elements and obstacles include constrained passages, sharp turns, large drops/climbs, inclines, steps, ladders, mud, sand, and/or water. The environments may also include organic or human-made materials; structured or unstructured clutter; and intact or collapsed structures and debris.

\ph{Degraded Perception and Sensing} The environment includes elements that range from constrained passages to large openings, lighted areas to complete darkness, and wet to dusty conditions. Such environments with limited visibility, difficult and expansive terrain, and/or sparse features can lead to significant localization error and drift over the duration of an extended run. Perception and proprioceptive sensors will need to reliably operate in these low-light, obscured, and/or scattering environments while having the dynamic range to accommodate such varying conditions. Dust, fog, mist, water, and smoke are among the challenging elements.

\ph{Constrained Communication} Limited line-of-sight, radio frequency (RF) propagation challenges, and effects of varying geology in subterranean environments impose significant impediments to reliable networking and communications links. Teams in this competition consider innovative approaches to overcome these constraints, including novel combinations of hardware, software, waveforms, protocols, distributed or dispersed concepts, and/or deployment methods.

\ph{Endurance and Power Limits} To succeed in accomplishing the mission goals, teams need to be capable of a team-aggregated endurance of 60-120 minutes. This aggregate endurance requires novel deployment concepts, energy-aware planning, heterogeneous agents of varying endurance, energy harvesting or transfer technologies, and/or a combination of various approaches to overcome the various challenge elements.

\section{Concept of Operations}\label{sec:conops}
\begin{figure}[ht]
  \centering
  \includegraphics[trim={0 10cm 0 2cm},clip,width=\linewidth]{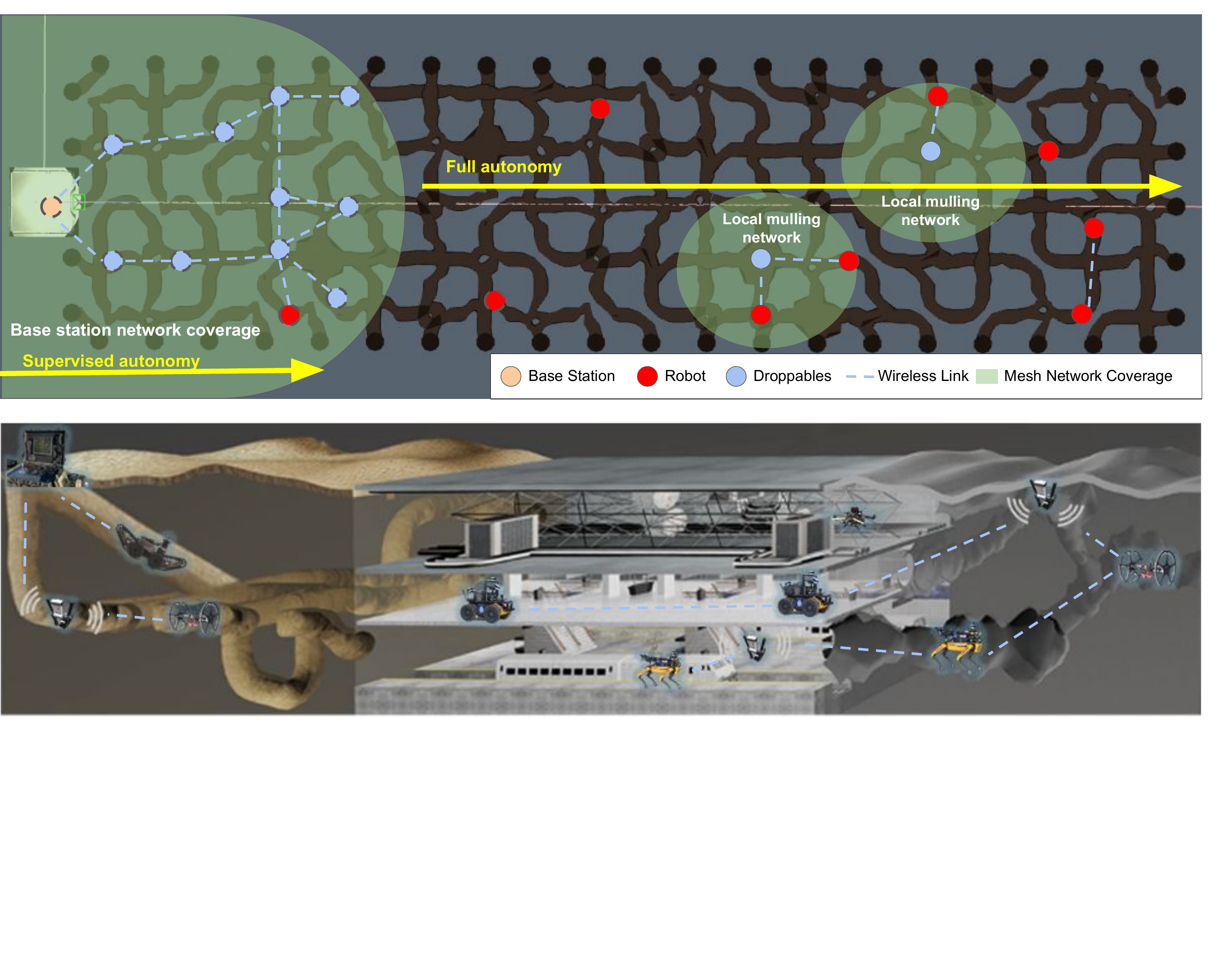}
  \caption{NeBula's Concept of Operation. Top: Bird’s eye view of autonomy in cave. Bottom: Perspective view with our robots and different environments. }
  \label{fig:conops_diagram}
\end{figure}

In this section, we briefly go over NeBula’s ConOps (\autoref{fig:conops_diagram}) for exploring unknown extreme terrains under time constraints. In order to simultaneously address various challenges associated with exploring unknown challenging terrains (\autoref{subsec:subt_intro_challenge}), we rely on a team of heterogeneous robots with complementary capabilities in mobility, sensing, and computing.

\begin{figure}[t]
  \centering
  \includegraphics[trim={0 3cm 0 2cm},clip,width=0.9\linewidth]{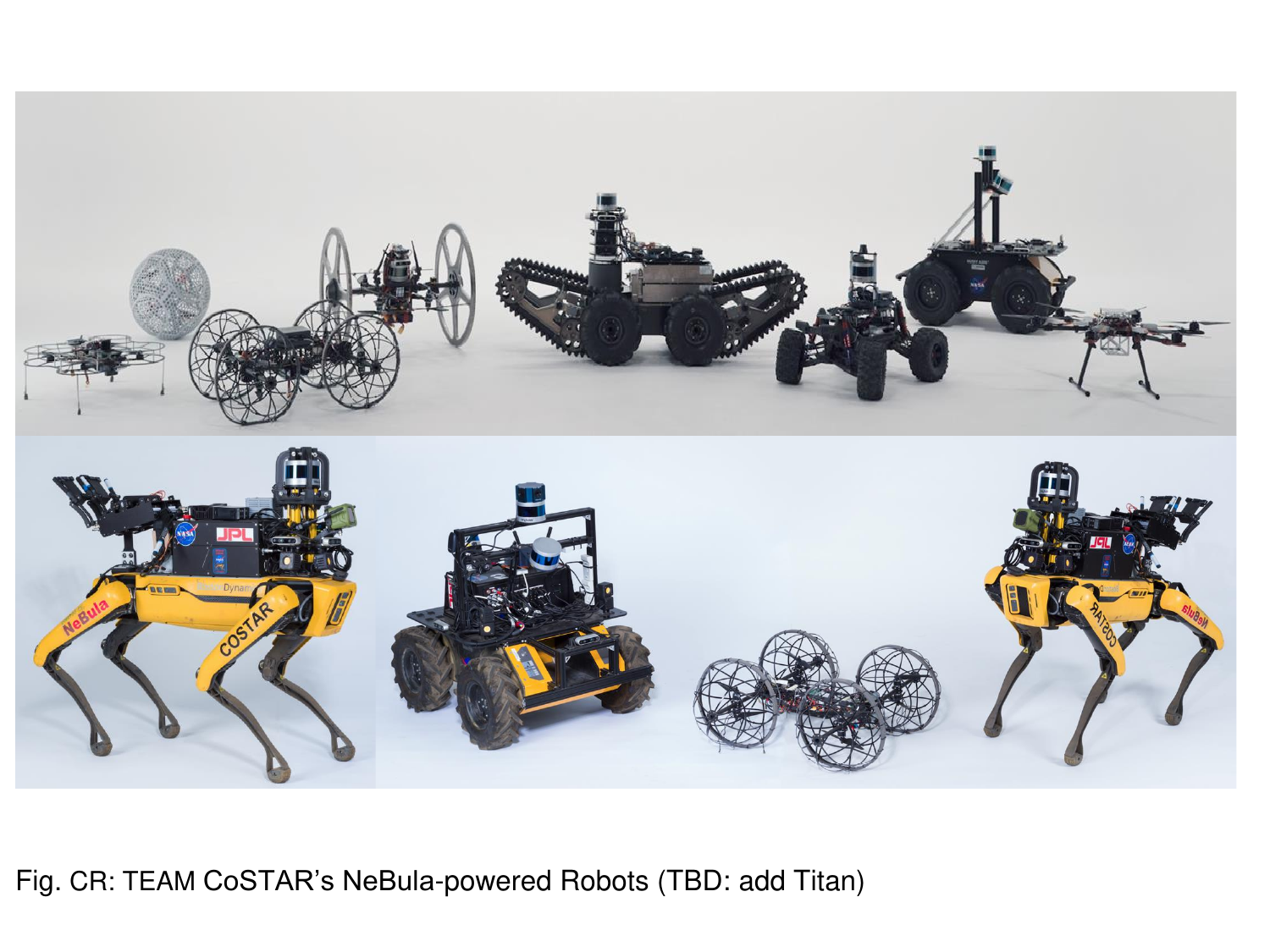}
  \caption{Team CoSTAR's NeBula-powered robots}
  \label{fig:nebula_robots}
\end{figure}

\ph{Robot Capabilities}
\autoref{fig:nebula_robots} shows the robots we have deployed. The capabilities of these robots drive the ConOps design process. Tables \ref{table:robot_mobility_modes}, \ref{table:nebula_sensor_stack}, \ref{table:nebula_processors} summarize our heterogeneous robot capabilities from mobility, sensory, and computing perspectives. Our ConOps induces specific combinations of mobility-sensor-computer, defining the robots we deploy in the environment to satisfy the mission objectives. The payload capacity of each robot is directly correlated with the sensory capacity; larger payload capacity allows for a larger sensory suite. Also, the energy/battery capacity, and desired endurance on each robot is correlated with their processing capabilities; typically, larger robots are able to carry larger batteries and more powerful computing resources. Some of these mobility-sensor-computer combinations are discussed in \autoref{sec:hardware}.

\begin{table}[t]
\begin{centering}
\caption{Heterogeneous NeBula-powered Mobility Modes}
\label{table:robot_mobility_modes}
\resizebox{\columnwidth}{!}{
\begin{tabular}{|c||c|c|c|c|c|c|c|}
\cline{1-8}
\hline 
\textbf{Robot Type} &\textbf{Deployed In} &\textbf{Energy capacity} &\textbf{Payload capacity} &\textbf{Comm} &\textbf{Speed} &\textbf{Mobility} &\textbf{Endurance}  \tabularnewline
\hline 
\hline
\textbf{Legged robots} &Urban  &Mid &Mid &Mid &Mid &Mid &Mid \tabularnewline
\hline 
\textbf{Hybrid (ground/aerial)} &STIX  &Low &Low &Low &High &Mid-High &Low-Mid \tabularnewline
\hline 
\textbf{Wheeled} &STIX, Mine, Urban  &High &High &High &Low &Low &High \tabularnewline
\hline 
\textbf{Drones} &STIX, Mine  &Low &Low &Low &High &High &Low \tabularnewline
\hline 
\textbf{Tracked} &STIX  &High &Mid &Mid &Low &Low-Mid &Mid-High \tabularnewline
\hline 
\textbf{Fast small rovers} &Mine  &Mid &Mid &Low &High &Low &Mid \tabularnewline
\hline 
\textbf{Aggregated robot team} &All events  &Shared/Synergistic &Shared/Synergistic &Shared/Synergistic &Aggregated &Aggregated &Aggregated \tabularnewline
\hline 
\end{tabular}
} %
\par\end{centering}
\begin{centering}
\caption{Heterogeneous NeBula Sensing Modalities (the values are based on our specific ConOps) }
\label{table:nebula_sensor_stack}
\resizebox{\columnwidth}{!}{
\begin{tabular}{|c||c|c|c|c|c|c|c|c|c||c|c|c||}
\cline{2-13}
\multicolumn{1}{c||}{} & \multicolumn{9}{c||}{
\textbf{Exteroceptive}}
& \multicolumn{3}{c||}{\textbf{Proprioceptive}}\tabularnewline
\hline 
\textbf{Sensors} &Lidar &Vision &Radar &Thermal &Sonar &IR Depth &CO2/Gas &Wi-Fi &Sound &Contact/Force &Encoder &IMU\tabularnewline
\hline 
\hline
\textbf{Accuracy}&High &Mid &Low &Low &Low &High &Low &Low &Low &Low &Mid &High\tabularnewline
\hline 
\textbf{Power efficiency} &Low &High &High &Mid &Mid &High &High &High &High &High &High &High\tabularnewline
\hline 
\textbf{Size/weight efficiency} &Low &High &High &Low &Mid &High &High &High &Mid &Mid &Mid &High \tabularnewline
\hline
\textbf{Range and FOV} &Mid &High &Low &High &Low &Low &Low &High &Mid &- &- &- \tabularnewline
\hline
\textbf{Dark/fog/smoke/dust} &Mid &Low &High &High &Mid &Mid &- &- &- &- &- &-\tabularnewline
\hline
\end{tabular}
} %
\par\end{centering}
\begin{centering}{
\caption{NeBula Processors}
\label{table:nebula_processors}
\resizebox{5in}{!}{
\begin{tabular}{|c||c|c|c|c|c|}
\cline{1-6}
\hline 
\textbf{Processors} &\textbf{Micro-controllers} &\textbf{Snapdragon} &\textbf{Intel NUC} &\textbf{Nvidia Xavier} &\textbf{AMD} \tabularnewline
\hline 
\hline
\textbf{Compute} &Low &Low &Mid &High &High \tabularnewline
\hline 
\textbf{Power consumption} &Low &Low &Mid &High &High 
\tabularnewline
\hline 
\textbf{Size efficiency} &High &High &Mid &Low &Low \tabularnewline
\hline
\end{tabular}
} %
}
\par\end{centering}
\end{table}

\ph{ConOps}
Our Concept of Operations (ConOps) utilizes a heterogeneous set of platforms (see \autoref{table:robot_mobility_modes} and \autoref{fig:nebula_robots}). In the following, we describe several steps of an example illustrative mission ConOps.

\vspace{-5mm}
\begin{enumerate}[leftmargin=*]
    \item \underline{Vanguard Operations}: As the robots enter the environments, they explore the frontier with vanguard robots with highly capable sensing for mapping and artifact detection.
    
    \item \underline{Mesh Network Expansion}: As robots start the mission they aim at maintaining communication with the human supervisor by creating and extending a wireless mesh network inside the environments of networking. Ground robots do so by deploying communication pucks like breadcrumbs, and aerial scouts can self-deploy for either comms relays or added sensing. Mission autonomy will decide where and how to deploy these breadcrumbs.
    
    \item \underline{Leaving the Mesh Network}: The environment is highly communication-denied. Due to the large scale, complexity of the environment, and capacity of robots to carry communication nodes, the mesh network reach is typically limited to the parts of the environments in the vicinity of the base station (i.e., environment entrance). Hence, the robots will leave the communication network range soon and, for the most part, carry out a fully autonomous mission.
    
    \item \underline{Autonomous Mission}: Robots perform search, mapping, and exploration. Autonomous mission guides them to the rendezvous points to exchange information with each other, or they come back to the mesh network to exchange information with the base station.
   
    \item \underline{Dynamic Task Allocation}: Robots continue simultaneous frontier exploration. They autonomously monitor the
    \begin{enumerate}
        \item state of the robot team that includes: $(i)$ health, battery, and functionality level of the assets, $(ii)$ robot locations, $(iii)$ the information value (e.g., the numbers of detected artifacts) on each robot.
        \item state of the world that includes what robots learn about the environment, e.g., geometric and semantic maps,
        \item state of the mission that includes: $(i)$ the remaining mission time, and $(ii)$ margin to the desired mission output, $(iii)$ acceptable risk thresholds.
        \item state of communication: $(i)$ network connectivity, throughput, etc., $(ii)$ how long each robot is out of the comm range, $(iii)$ location of comm nodes
    \newline 
    \newline
    Given these states, the mission planner will decide to deploy new robots, re-task or re-position active robots in the environment.
    \end{enumerate}
    
    \item \underline{Team Behaviors}: Vehicles and team formation are configured during the mission. Examples include:
    \begin{enumerate}
        \item Return to Base, when a battery swap is needed, optimal, and possible at base,
        \item Return to Mesh Network, to ensure the data are communicated, then continue,
        \item Explore Frontier, continue as is, aggressively prioritizing coverage,
        \item Act as a Data Mule: to retrieve data from a vehicle who cannot come back to the mesh network (due to limited battery, health, speed, etc), faster and healthier vehicles can act as data mules to carry the information between others agents and the mesh network.

    \end{enumerate}
    
    \item \underline{Heterogeneous Coverage}: These behaviors continue until the entire course is explored. Due to the heterogeneous capabilities of the robots from mobility, sensing, and computation perspectives, the autonomy might dispatch different robots to the same parts of the course. This is to increase the confidence and coverage in mapping and artifact detection by providing multimodal information (e.g., thermal, radar, etc) about the environment elements. For example, the drone might have reached and searched parts of the course but given its limited sensing capabilities, autonomy will dispatch a ground robot to get a second vote on an artifact before submitting it to the server. All the data is submitted to the server prior to the end of mission time.

\end{enumerate}
\section{NeBula Autonomy Architecture}
\label{sec:nebula_architechture}

Resiliency is a key requirement to enable a repeatable and consistent robotic autonomy solution in the field. \emph{To enable a resilient autonomy solution, NeBula takes uncertainty into account} to cope with unmodeled and unknown elements during the mission. 

\ph{Motivation/Insight}
An important (if not the most important) cause of the brittleness of today's autonomy solutions is the disjoint design of various subsystems. Traditionally, when designing or advancing the performance of a certain module, the typical assumption is that the rest of the system functions properly and nominally. When it comes to real-world deployment, these assumptions typically break due to the discrepancy between the computational models and real-world models. This introduces uncertainty in the perception, inferences, decision-making, and execution, potentially leading to suboptimal behaviors.

\ph{Key principle for resilient autonomy architectures}
Focused on fielding autonomy in challenging environments, NeBula is built on a fundamental principle that: ``To achieve resilient autonomy, the decision-making, inference, and perception modules must be reciprocal and tightly co-designed.'' This implies reasoning over joint probability distributions across various component-level states as opposed to marginal distributions over a set of disjoint system states. Contrary to the typical sense $\rightarrow$ infer $\rightarrow$ plan $\rightarrow$ act sequence in autonomy solutions, NeBula architecture is built on a plan-to-(sense $\rightarrow$ infer $\rightarrow$ act) loop, where the planner dynamically \emph{plans for} the acquisition of sufficient sensory information and \emph{plans for the quality of representation} required to enable resilient and uninterrupted missions within the prescribed mission risk thresholds. NeBula’s joint perception-planning is formulated as an uncertainty-aware belief space planning problem. Belief captures the probability distribution over system’s states including the robot pose, environment state, measurements, team coordination state, health state, communication state, etc. Planning over joint beliefs and taking cross-component uncertainty into account (which describes the interaction of connected modules), NeBula allows for each module to not only be robust to uncertainties within its own subsystem but also resilient to uncertainties in the integration process, resulted from imperfections and off-nominal performance of connected modules.

\begin{figure}[ht]
  \begin{minipage}[c]{0.64\textwidth}
     \caption{Illustrative example of joint inference and decision-making in NeBula's low-level navigation system. Denoting the state domain of the localization by pose $x$, mapping by map state $m$, and planing by policy parameters $u$, SLAM, SLAP, and SMAP aim at solving for (i.e., estimating or predicting) the joint distributions $p(x,m)$, $p(x,u)$, and $p(m,u)$. The full joint problem, SPLAM (simultaneous planning, localization, and mapping) solves for the probability distribution $p(x,m,u)$.} \label{fig:illustrative_ex}
  \end{minipage}
  \begin{minipage}[c]{0.33\textwidth}
    \includegraphics[width=\textwidth]{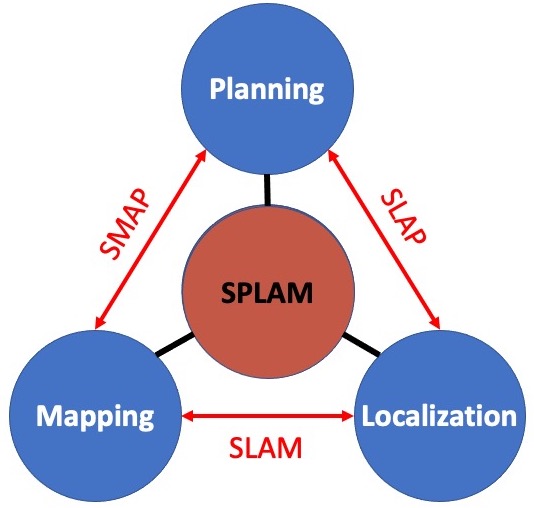}
  \end{minipage}\hfill
\end{figure}

\ph{Illustrative example (Perception-aware planning under uncertainty)}
Simultaneous localization and mapping (SLAM) is a fundamental problem in robotics that aims at simultaneously solving the localization problem (``where is the robot?") and  the mapping problem (``what does the environment look like?"). This is a very well studied problem and it is well-known that solving SLAM (i.e., incorporating joint probability distributions between localization states and environment states) typically leads to optimal and resilient inference, whereas solving localization and mapping separately and putting their solutions together is suboptimal and can lead to a brittle inference system. 
Analogous to SLAM philosophy, NeBula extends this concept from pure inference to “joint inference and decision-making” (Fig. \ref{fig:illustrative_ex}). For example, NeBula develops solutions where mapping and planning are solved simultaneously using SMAP (simultaneous mapping and planning) to achieve resilient traversability and risk-awareness. Similarly NeBula develops SLAP (simultaneous localization and planning) solutions where localization uncertainty is taken into account in the planning phase using belief space planners. Solving these joint problems typically leads to behaviors where the autonomous system is intelligently planning proactive actions to improve the “inference quality” (e.g., world model or robot model) and reduce uncertainty to the levels necessary to achieve mission goals within the prescribed risk thresholds. This is in contrast to typical solutions where the relationship is one-way and the inference module serves the decision-making modules, and decision-making components react to inference output.

\ph{Modularity and scalability}
In addition to resiliency, NeBula focuses on a modular and scalable framework. This requirement is driven by missions carried out by a team of networked heterogeneous robots. Each robot has different mobility (e.g., wheeled, legged, aerial, hybrid), sensing, and computational capabilities (e.g.,~\autoref{sec:hardware}). NeBula provides appropriate abstraction to allow for re-usability and agnosticism to the specific robot and hardware. Any low-level hardware-specific modules should be properly isolated to increase the re-usability of software. Further, NeBula supports networked systems where agents are intermittently losing and re-establishing communications, and sharing knowledge with each other and with the base station, enabling large-scale environment exploration with a limited number of robots. Each robot has a level of local autonomy to act individually when it is disconnected from the rest of the team.

\begin{figure*}[!t]
  \centering
  \includegraphics[width=0.8\textwidth,angle=-90]{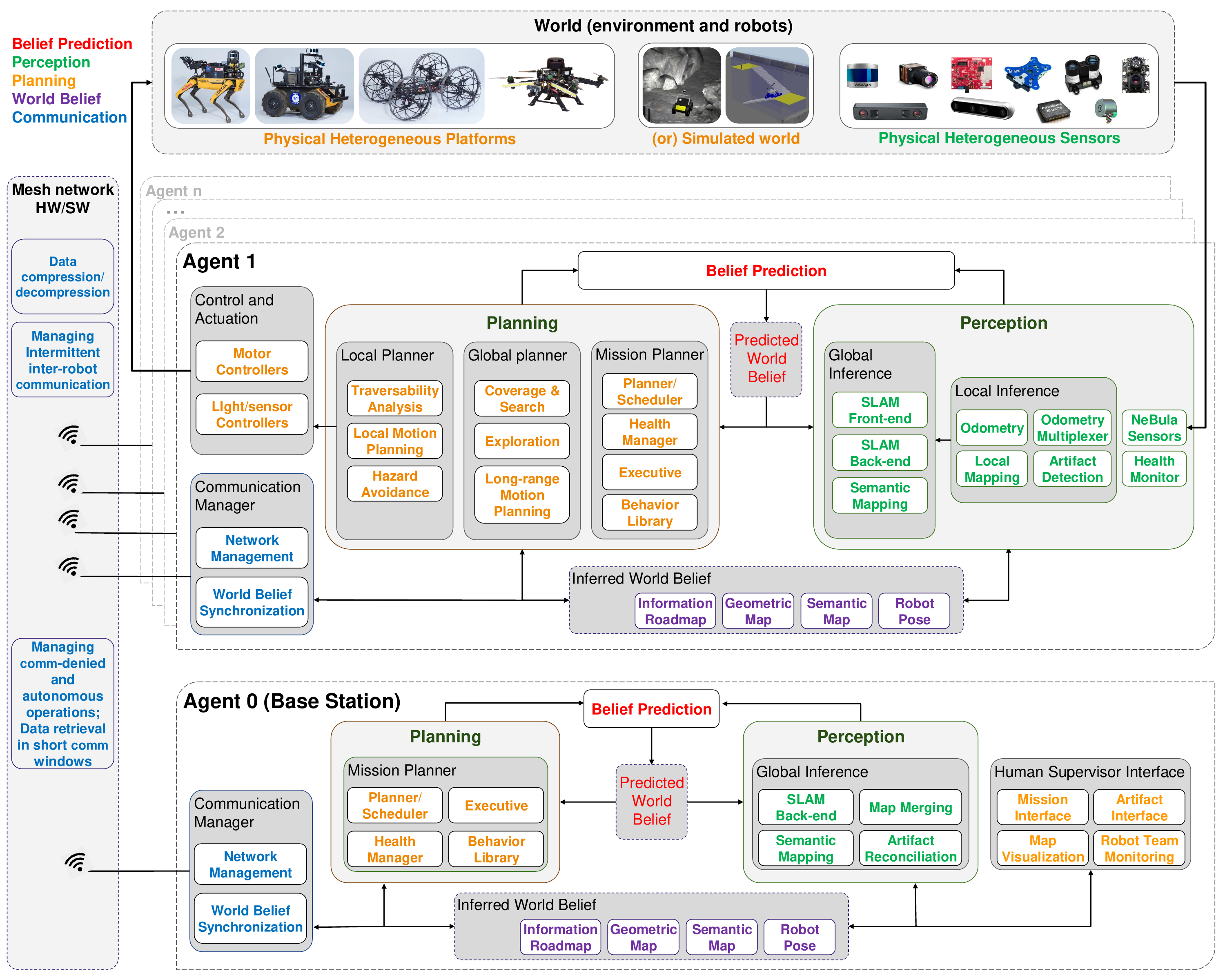}
  \caption{NeBula functional block diagram}
  \label{fig:nebula_architecture}
\end{figure*}

\ph{System architecture}
\autoref{fig:nebula_architecture} illustrates the NeBula’s high-level functional block diagram, and will serve as a visual outline of the sections of this paper. The system is composed of multiple assets: mobile robots, stationary comm nodes, and a base station, each of which owns different computational and sensing capabilities. The base station acts as the central component to collect data from multiple robots and distribute tasks, if and when a communication link to the base station is established. In the absence of the communication links the multi-asset system performs fully autonomously. The fundamental blocks are:
\vspace{-15pt}
\begin{itemize}[leftmargin=20pt]
    \item \textit{Perception} (\autoref{sec:state_estimation},~\ref{sec:lamp},~\ref{sec:artifacts}): Perception modules are responsible to process the sensory data and create a world model belief. The local perception modules (\autoref{sec:state_estimation}) provide the odometry and state estimation information needed for local navigation, such as state (pose, velocity) and traversability maps. The global SLAM (simultaneous localization and mapping) module, in~\autoref{sec:lamp}, tracks the robot's position within a globally consistent frame while building a 3D map of the environment. The semantic understanding and artifact detection module (\autoref{sec:artifacts}) adds semantic information to the map and finds objects of interest from the environment, and in conjunction with the global localization module reports their location.
    \item \textit{Planning} (\autoref{sec:traversability},~\ref{sec:global_planning},~\ref{sec:mission_planning}): Planning modules will make onboard decisions based on the current world belief. The planning modules are composed of multiple layers. The highest layer is the mission planning module (\autoref{sec:mission_planning}) which runs a mission according to its specifications, generates global goals for each robot, and allocates tasks to different robots. The second layer is the global motion planning layer (\autoref{sec:global_planning}), responsible for exploration, search, and coverage behaviors in global scale and large environments. It makes plans to safely move the robot to a goal assigned by the mission planner. It also enables autonomous exploration of the environment in order to increase the knowledge and confidence about the world-belief. The third layer is the traversability and local navigation component (\autoref{sec:traversability}), responsible for analyzing how and with what velocities different terrain elements can be traversed. It quantifies the motion risk, and optimizes/replans local trajectories with high frequency to enable aggressive traversability in obstacle-laden and challenging  environments, while ensuring the risk levels stay within the prescribed mission specifications. NeBula abstracts motion models, enabling the planning stack to be robot-agnostic and to support heterogeneous mobility platforms.
    \item \textit{World Belief}: This block includes a probabilistic model of the world. It is jointly constructed by perception and planning modules, and enables a tight-integration between these modules leading to perception-aware behaviors. World belief extends the traditional state database concept to a belief database, where we maintain probability distributions over various state domains as well as joint probability distributions across multiple domains. It includes belief over robot pose, environment map, mission state, system health, information roadmap, among other state domains. There are multiple variations of the world belief: (i) local to each robot, (ii) shared belief across robots, and (iii) predicted belief to assess future risk and performance to enable making perception-aware and uncertainty-aware decisions. During exploration tasks, robots develop their own local world models based on what they perceived with their limited sensor input. They generate the world model as an abstract representation of spatial and temporal information of their surrounding environment (e.g., maps, hazards) and internal state (e.g., pose, health). This world belief is internally predicted to enable uncertainty-aware decisions and actions based on this predicted model. The shared world model is synchronized among the robots and the base station using asynchronous bidirectional messaging with the publish/subscribe paradigm. The discussion of world belief is distributed across all sections of the paper.
    \item \textit{Communications} (\autoref{sec:multirobot_networking}): When possible, communication modules synchronize the shared world models across the robots and the base station. To cope with the dynamic and unstable nature of the underlying mesh network, the communication manager is responsible to provide reliable message transfer with buffering, compression, and re-transmission. The modules also provide capabilities to maintain a mobile ad-hoc network using radio devices. Static communication nodes can be dropped from particular robots to help forming a network.
    \item \textit{Operations} (\autoref{sec:mission_planning}): Operation modules aid the human supervisor to effectively monitor the system performance and interact with it if and when communication links are established. One of the main roles is the visualization of complex world belief in a human-recognizable form. In the nominal operation scenarios, the human operator only interacts with the system by updating the world belief, when needed and when possible.
\end{itemize}
\vspace{-3mm}
Over time, each layer adapts to the collected data as well as to improvements of models in other layers.

\ph{Current Implementation Status} NeBula is a growing and evolving framework. Its current version (in 2020) has been deployed in several large projects for terrestrial and planetary applications. It has enabled autonomous operations on various vehicle types including 1) wheeled rovers, 2) legged robots, 3) flying multicopters, 4) hybrid aerial/ground vehicles, 5) 1/5th scale race cars 6) tracked vehicles, and 7) full-size passenger vehicles. We refer to Section \ref{sec:hardware} for a detailed description of our implementation on these platforms.

\section{State Estimation}\label{sec:state_estimation}
One of the fundamental components of the NeBula architecture, shown in~\autoref{fig:nebula_architecture}, is reliable state estimation under perceptually-degraded conditions. This includes environments with large variations in lighting, obscurants (e.g., dust, fog, and smoke), self-similar scenes, reflective surfaces, and featureless/feature-poor surfaces. NeBula relies on a resilient odometry framework that fuses a set of heterogeneous sensors to handle these various challenges.
This section briefly describes this odometry solution. 
For more details, please see~\cite{hero2019isrr,LOCUS,RIO2020,SeyedMCCKF20,SeyedAMCCKF21,LION,lew2019contact}. %

\begin{figure}[!ht]
  \centering
  \subfloat[Main HeRO modules and interconnections]{\includegraphics[width=0.75\linewidth]{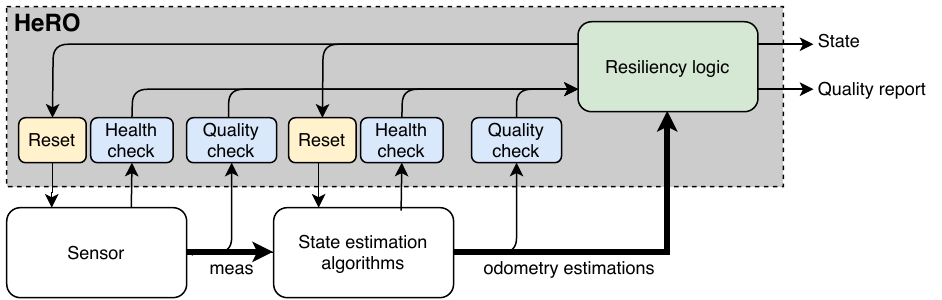}}\\
  \subfloat[Examples of HeRO sensors, algorithms (\emph{e.g.}, LIO: LiDAR-inertial, VIO: Visual-inertial, TIO: thermal-inertial, RIO: RaDAR-inertial, OF: Optical flow, CIO: contact-inertial or WIO: Wheel-inertial odometries) and observable state components]{\includegraphics[width=\textwidth]{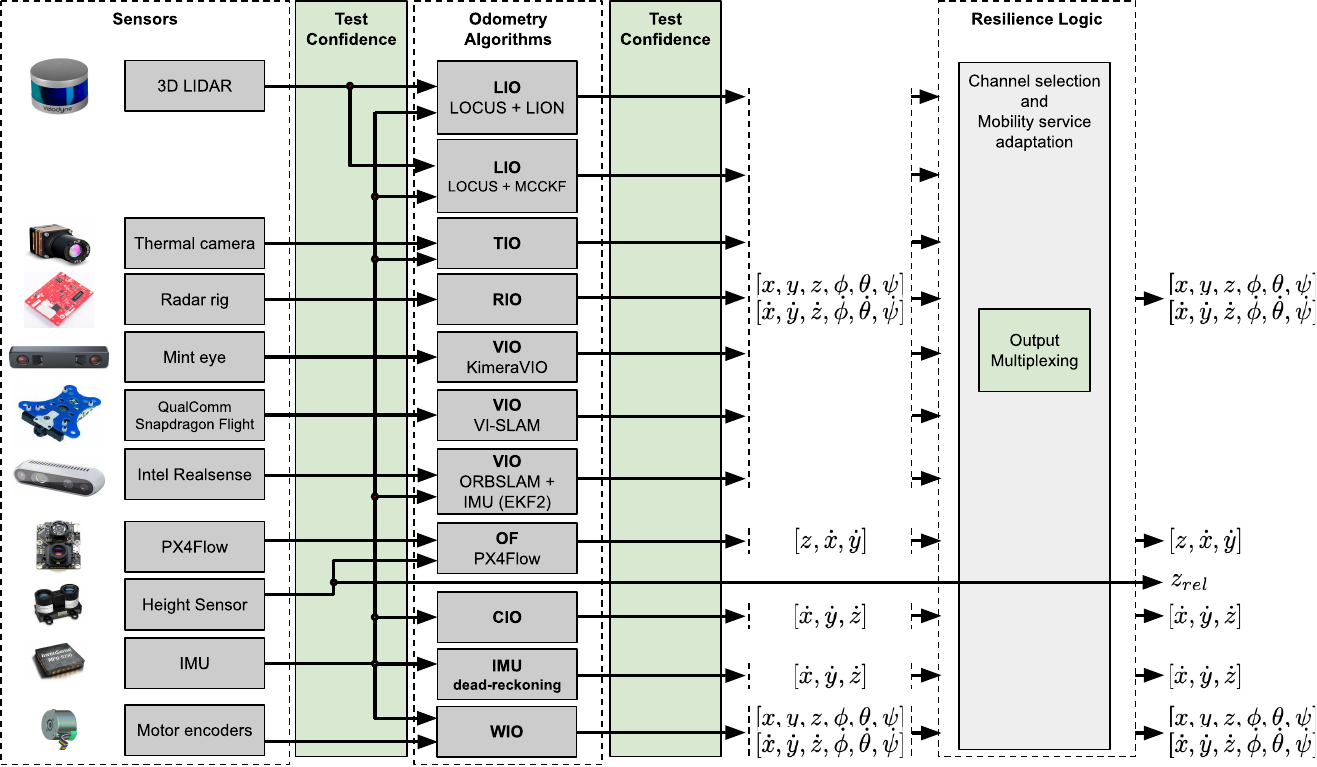}}
  \caption{NeBula's state estimation architecture
  }
  \label{fig:hero}
\end{figure}

\ph{Objective}
The objective of the state estimation pipeline is to utilize multi-modal sensing to determine the robot's state, producing resilient, high-rate and smooth estimates in a probabilistic sense.
A key aspect to the proposed approach is assigning each sensor output a \emph{quality measure} that can be used to identify healthy or unhealthy measurements \emph{before} they are fused.
We start by introducing our notation: $\vec{p} \in \mathbb{R}^3$ (global position, $x, y, z$); $\vec{R} \in SO(3)$ (global orientation, which can be described with minimal representation $\phi, \theta, \psi$); $\vec{v} \in \mathbb{R}^3$ (body linear velocity); $\vec{\omega} \in \mathbb{R}^3$ (body angular velocity); $\vec{a} \in \mathbb{R}^3$ (body linear acceleration); $\vec{\alpha} \in \mathbb{R}^3$ (body angular acceleration); and $Q_i \in \{Good, Bad\}$ (quality of $i$, with $i \in \{\vec{p}, \vec{R}, \vec{v}, \vec{\omega}, \vec{a}, \vec{\alpha}\}$).

Note that we restricted the quality of the state to binary values although it can be easily generalized to higher resolutions and even continuous representations.

\ph{HeRO Architecture}
The proposed architecture, shown in~\autoref{fig:hero}, considers redundancy and heterogeneity in both sensing and estimation algorithms.
It is designed to expect and detect failures while adapting the behavior of the system to ensure safety.
To this end, we present HeRO, Heterogeneous and Resilient Odometry estimator~\cite{hero2019isrr}: a framework of estimation algorithms running in parallel supervised by a resiliency logic.
Resilience logic has three main functions: a) perform confidence tests in data quality (measurements and individual estimations) and check health of sensors and algorithms; b) re-initialize those algorithms that might be malfunctioning; c) generate a smooth state estimate by multiplexing the inputs based on their quality.
The output of this resiliency logic, which includes a \emph{state quality} measure, is used by the guidance and control system to determine the best mode of operation that ensures safety (see, for instance, the NeBula interconnections between Perception and Planning modules in~\autoref{fig:nebula_architecture}).
For example, guidance and control could switch to pure velocity control if the position estimates are unhealthy or issue a stop command if both position and velocity estimates are unreliable. 

\ph{Heterogeneous complementary algorithms} 
In addition to selecting heterogeneous sensing modalities, HeRO uses heterogeneous odometry algorithms, \emph{e.g.}, LiDAR-inertial (LIO), visual-inertial (VIO), thermal-inertial (TIO), kinematic-inertial (KIO), contact-inertial (CIO) or RaDAR-inertial (RIO), running in parallel to decrease the probability of a state estimation failure.
The key idea behind HeRO is that any single state estimation source can carry errors, either due to failures in sensor measurements, algorithms or both, but having a complete failure becomes increasingly rare as the number of heterogeneous parallel approaches increases.
HeRO is front-end agnostic, accepting a various algorithmic solutions and with the ability to incorporate either tightly or loosely coupled approaches. 
However, to take advantage of all possible mobility modes, there is a need for estimating position, orientation, velocity and, ideally, acceleration.
HeRO is tailored to incorporate a vast variety of estimation algorithms.
\autoref{fig:hero} depicts the main sensor and algorithmic solutions developed and used by team CoSTAR in the DARPA Subterranean Challenge.
\rev{Our solution considers software-synchronized sensors (common clock synchronization after initialization), with extrinsic calibrations roughly obtained from the robot model designs and fine tuned used optimization approaches such as Kalibr (camera-imu)~\cite{FurgaleRehderEtAl2013,RehderNikolicEtAl2016} or LiDAR-align (LiDAR-LiDAR)~\cite{ethz_lidar_align}; or by aligning with the robot manufacturer frames.}
\rev{We rely on a variety of in-house and open-source algorithms for sensor fusion. A few examples are as follows: WIO uses an Extended Kalman Filter (EKF) to fuse the measurements of the wheel encoders and those from an IMU. 
CIO also takes advantage of an EKF but this time including the modelling of the contacts~\cite{lew2019contact}.
The optical flow approach (OF), VIO, and thermal imagery fusion (TIO) leverage a combination of opensource and commercial solutions, including \revv{PX4Flow \cite{px4Flow}, ORBSLAM \cite{murORB2}, Qualcomm VI-SLAM~\cite{qualcommtechnologiesinc.MachineVisionSDK}, and the MiT KimeraVIO \cite{Rosinol20icra-Kimera}, ROTIO~\cite{khattak19}}, among others.
The RIO algorithm is our own development presented in~\cite{RIO2020}.
Finally, the fusion of LiDAR scans with IMU measurements is done by combining our LOCUS scan matching~\cite{LOCUS}  with IMU measurements, either using our Kalman filter variant (AMCCKF~\cite{SeyedAMCCKF21}) or a factor graph optimization.}
As an example, in the following we describe the latter LiDAR-inertial odometer.

\subsection{LiDAR-inertial odometry estimation}

LiDAR is one of the key sensors offering high range, accuracy. LiDARs also perform well in low-light conditions. Moreover, the combination with an IMU provides essential dynamic information and the registration of the gravity vector, which is of high importance for planning modules (e.g., computing traversability regions).
There exist several methods exploiting LiDAR-IMU data fusion achieving remarkable accuracy~\cite{liosam2020shan,ye2019tightly,hess2016real}.
However, they do not consider potential failures of the fused sensing modalities, which are likely to be observed in real-world field deployments and can result in catastrophic degradation of the odometry performance if not robustly handled. 

To enable reliable operation in extreme settings, our proposed LiDAR-inertial odometry estimation consists of three modules: $(i)$ A LiDAR front-end, which provides ego-motion estimation by analyzing LiDAR scans; $(ii)$ an IMU front-end based on pre-integration techniques; and $(iii)$ a data fusion module, formulated as a factor graph optimization problem, \rev{which fuses the data provided by the front-ends and also analyzes the LiDAR observability.}
This observability analysis is then used by HeRO to take informative decisions about using or not the LIO estimation.
These modules are briefly described hereafter.

\subsubsection{LiDAR front-end}
\begin{figure}[!ht]
  \centering
  \includegraphics[width=0.6\linewidth]{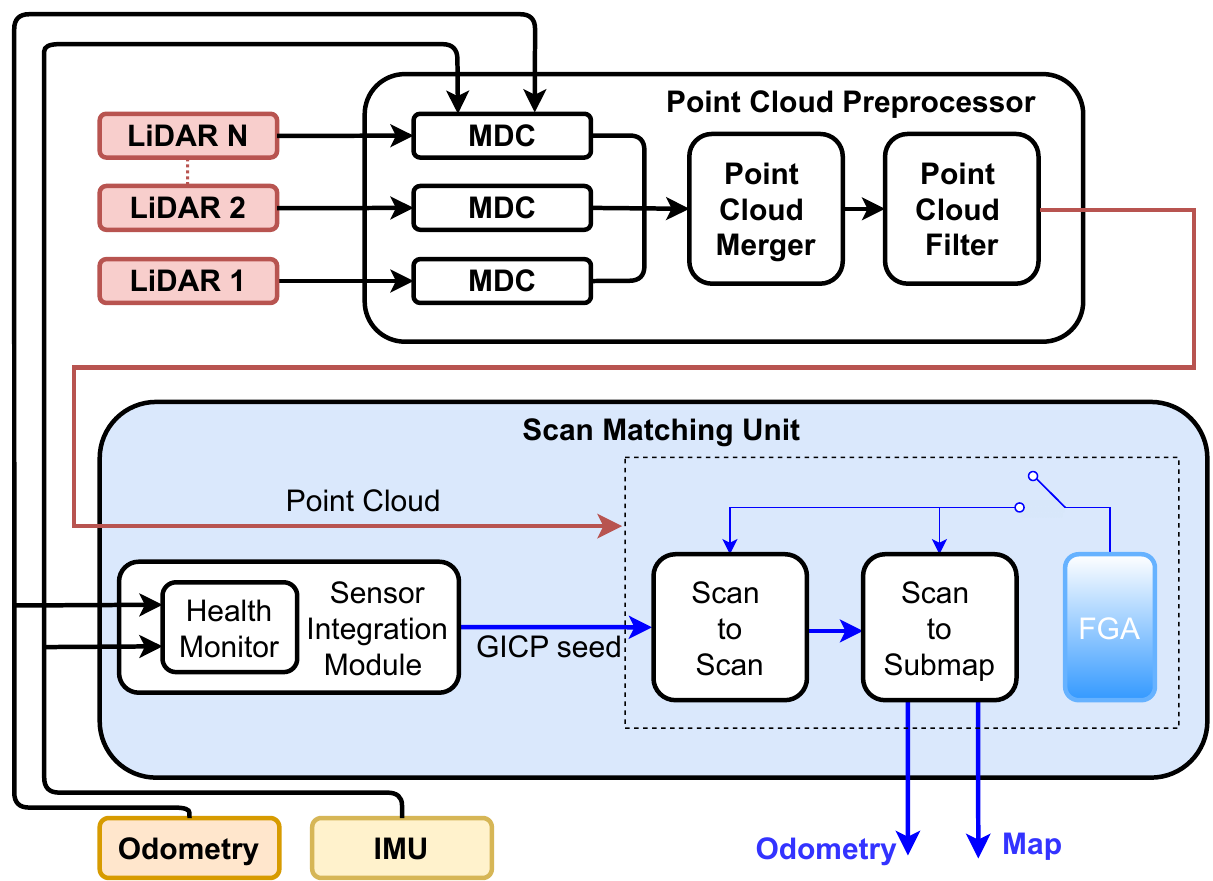}
  \caption{Architecture of the proposed system. In the block diagram MDC stands for Motion Distortion Correction, while FGA stands for Flat Ground Assumption.}
  \label{fig:locus_architecture}
\end{figure}

\ph{LOCUS}
NeBula's LiDAR-centered front-end, referred to as LOCUS (Lidar Odometry for Consistent operations in Uncertain Settings)~\cite{LOCUS}, is a multi-sensor LiDAR-centric solution for high-precision odometry and 3D mapping in real-time. LOCUS provides a Generalized Iterative Closest Point (GICP)~\cite{segal2009generalized} based multi-stage scan matching unit equipped with a health-aware sensor integration module for robust fusion of additional sensing modalities in a loosely coupled scheme. The architecture of the proposed system, depicted in~\autoref{fig:locus_architecture}, is made of three main components: $(i)$ point cloud pre-processor, $(ii)$ scan matching unit, $(iii)$ sensor integration module.

\begin{figure}[!ht]
  \centering
  \includegraphics[width=1.0\linewidth]{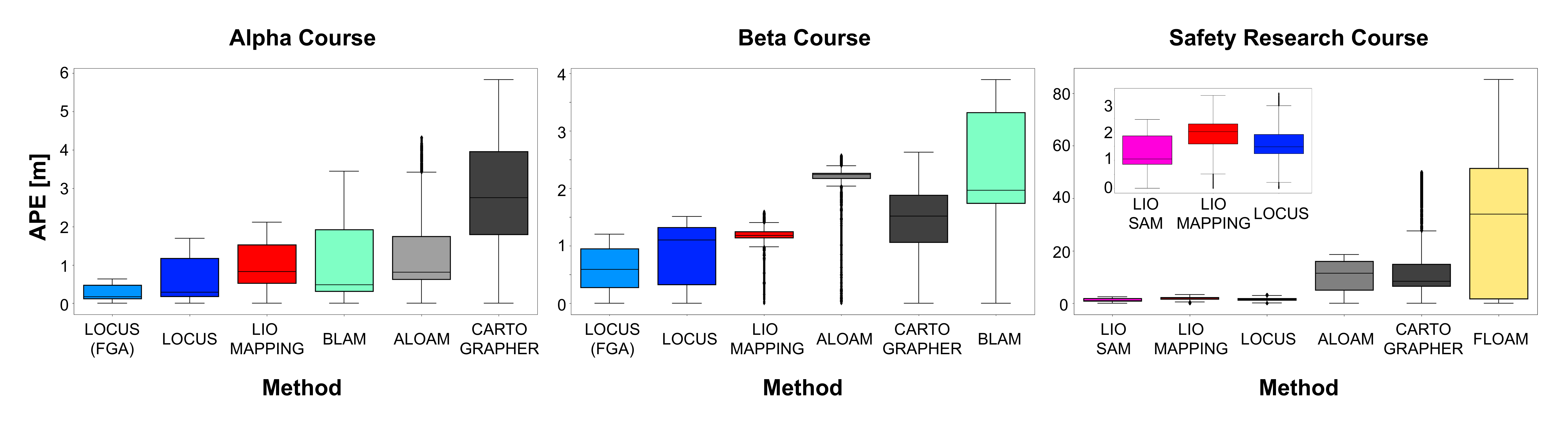}  \caption{\rev{Boxplot visualization of the absolute position error (APE) computed for the different LiDAR-based methods on the perception-degraded underground test datasets: for clarity, only the best six algorithms in each dataset are shown. A larger drift is observed in the Safety Research Course as in this dataset points are not motion corrected and the course presents many perceptually-challenging conditions such as harsh, unstructured terrains and long, feature-less corridors: for this run, the inset reports a zoomed version of the performance of the best three algorithms.} }
  \label{fig:locus_results}
\end{figure}

The point cloud pre-processor is responsible for the management of multiple input LiDAR streams (e.g. syncing, motion-correction, merging, filtering) to produce a unified 3D data product that can be efficiently processed in the scan matching unit. 
The scan matching unit then performs a cascaded GICP-based scan-to-scan and scan-to-submap matching operation to estimate the 6-DOF motion of the robot between consecutive LiDAR acquisitions. 

The sensor integration module is a key-component of the system to enable joint optimization of accuracy and robustness. In robots with multi-modal sensing, when available, LOCUS can use an initial transform estimate from a non-LiDAR source to ease the convergence of the GICP in the scan-to-scan matching stage, by initializing the optimization with a near-optimal seed that improves accuracy and reduces computation, enhancing real-time performance. Multiple sources of odometry (e.g VIO, KIO, WIO) and raw IMU measurements available on-board are fed into a sensor integration module which selects the output from a priority queue of the inputs that are deemed healthy by a built-in health-monitor, which prioritizes the order based on the expected accuracy of the methods. If the highest priority input is not healthy, then the next highest priority is used. If all sensors fail, the GICP is initialized with identity pose and the system reverts to pure LiDAR odometry. Notice how the confidence tests (health monitoring) depicted in~\autoref{fig:hero} are here incorporated within this sensor integration module. \cite{LOCUS} provides more details on the system functioning.

\ph{LOCUS Comparative Results} We compare LOCUS with state of the art LiDAR odometry methods~\cite{liosam2020shan,ye2019tightly,hess2016real,LOAM,BLAM} in extreme, perceptually-degraded subterranean environments and demonstrate high localization accuracy along with substantial improvements in robustness to sensor failures. For the evaluation,  we  use the  data  collected in the Tunnel and Urban Circuit rounds of the DARPA Subterranean Challenge, from a wheeled ground robot carrying 2 Velodyne LiDARs, an IMU and running WIO onboard. To assess accuracy, we compute the absolute position error (APE) of the estimated robot trajectory against the ground-truth reference, for the different methods over the different runs, and report in \autoref{fig:locus_results} a summary of the results. Throughout all the operations, LOCUS achieves highly accurate performance. To assess robustness, we analyze the flexibility of the various methods with respect to sudden failures of an input source by testing the following failure scenarios: $(i)$ failure of WIO/IMU; $(ii)$ failure of WIO; $(iii)$ failure of LiDAR. In these scenarios, tightly-coupled approaches and methods designed with synchronized callbacks stop operating when an input is missing. In contrast, LOCUS consistently achieves reliable ego-motion estimation and mapping, demonstrating efficient handling of sensor failures in a cascaded fashion, behaviour that is desirable to accommodate the unforeseen challenges posed by real-world operations where hardware failures are likely to happen, or sensor sources can become unreliable (see~\cite{LOCUS} for details).


\subsubsection{IMU front-end}
\ph{IMU pre-integration}
The IMU front-end is based on a pre-integration technique of the inertial measurements.
This module leverages the state-of-the-art on-manifold pre-integration theory to summarize the high-rate IMU measurements into a single motion constraint~\cite{forster2015imu,forster2015manifold} for the subsequent pose-graph optimization performed in the LiDAR-IMU data fusion algorithms.
IMU pre-integration is also used to guarantee a pose and velocity estimate at high-rate and low latency, regardless of the time taken by the optimizer used in the back-end of the sensor fusion algorithm.

\subsubsection{LiDAR-IMU data fusion}
\begin{figure}
    \centering
    \begin{minipage}[b]{0.48\textwidth}
    \centering
    \includegraphics[width=\textwidth]{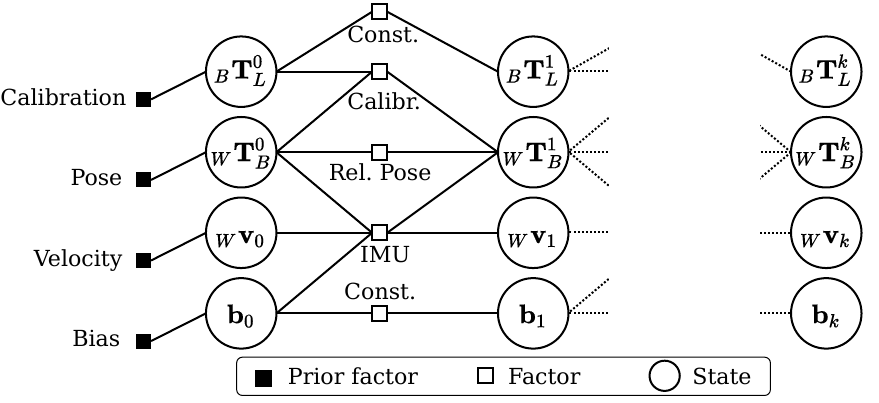}
    \caption{Factor graph representing state and measurements used by LION back-end. Biases are denoted with $\mathbf{b}_j:=\{_{B}\mathbf{b}^{a}, _{B}\mathbf{b}^{g}\}_j$.} %
    \label{fig:factor_graph_colored}
    \end{minipage}
  \hfill
  \begin{minipage}[b]{0.48\textwidth}
  \centering
    \includegraphics[width=\linewidth]{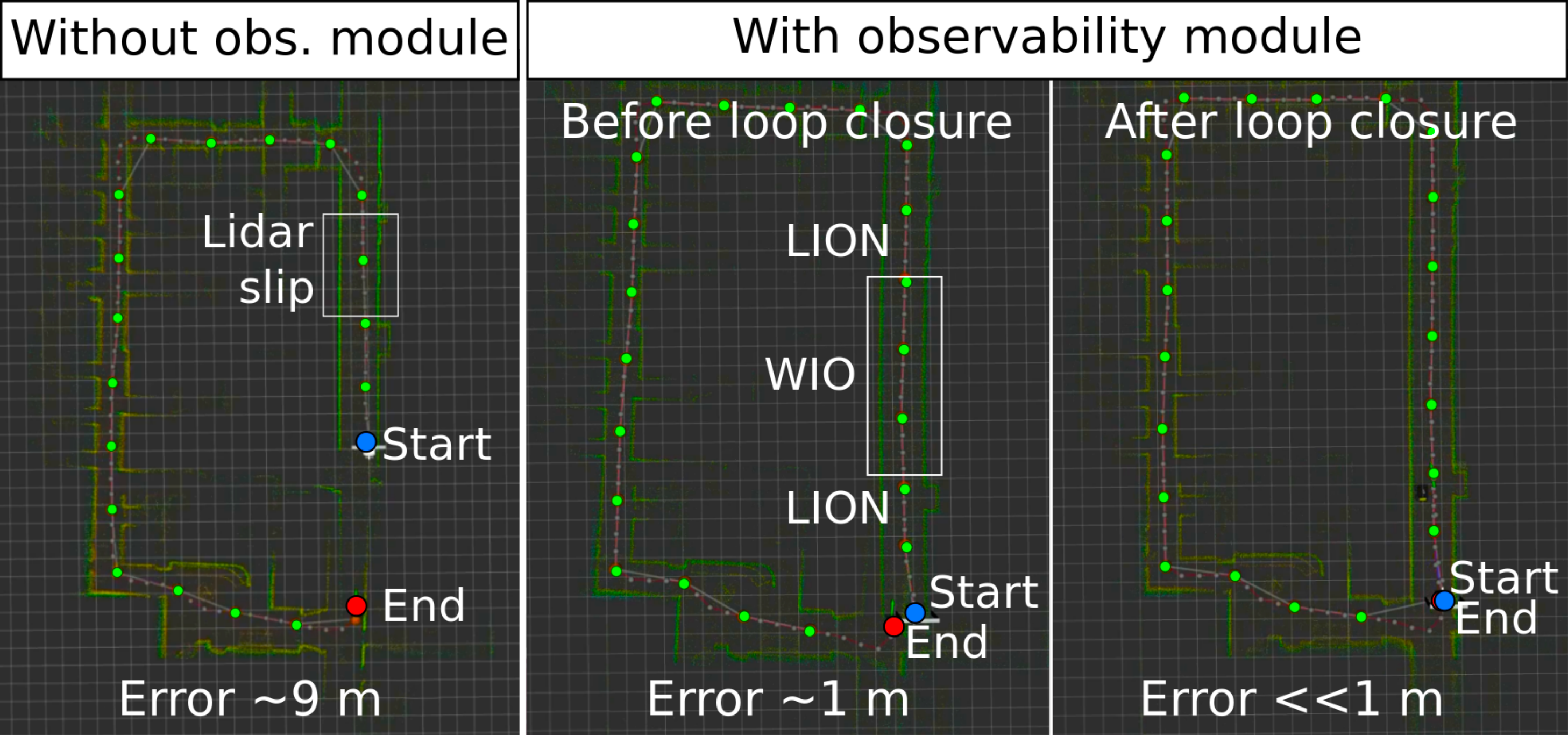}
  	\caption{Comparison of the translation error in an office-like environment, with and without the observability module integrated in HeRO.}
    \label{fig:slip_office}
  \end{minipage}
\end{figure}

\ph{Smoothing Framework (LION)} The fusion of the relative ego-motion estimations, obtained from LOCUS and IMU pre-integration front-ends, is performed via a fixed-lag smoother using a factor graph, as described in~\cite{LION}, where we introduce LION (LiDAR-Inertial Observability-aware Navigator for vision-denied environments).
The state estimated by the proposed smoother consists of: 
 \begin{inparaenum}[(a)]
 \item the pose (position and attitude) $\:_{W}\mathbf{T}_{B}$ of the IMU-fixed reference frame $B$ expressed in a slowly drifting inertial reference frame $W$;  
 \item the linear velocity $\:_{W}\mathbf{v}$;
 \item the IMU biases ($_{B}\mathbf{b}^{a}$ for the accelerometer and $_{B}\mathbf{b}^{g}$ for the gyroscope),  and 
 \item the extrinsic calibration $\:_{B}\mathbf{T}_{L}$ between the LiDAR-fixed frame $L$ and the IMU frame $B$, introduced to reduce the effects of error in mounting the sensors, as well as to address the challenges in offline LiDAR-extrinsic calibration. 
 \end{inparaenum} 
A representation of the states and factors used in the factor graph with a window of $k$ time steps can be found in~\autoref{fig:factor_graph_colored}. 
Following \cite{forster2015manifold}, we model the smoothing problem using GTSAM \cite{dellaert2012factor} and we solve the associated optimization with iSAM2 \cite{kaess2012isam2}.

\ph{Observability Module}
It is crucial for LiDAR-based estimation algorithms to determine if the geometry of the scene can well constrain the estimation of the translational motion, since long shafts and corridor-like structures can severely impact motion-observability. 
Following \cite{gelfand2003geometrically,bonnabel2016covariance}, \rev{we propose an observability metric, computed within LION architecture, which can inform the HeRO switching logic \autoref{fig:hero}} about potential unreliability in the odometry output of LiDAR-based estimators. Such metric is based on the condition number $\kappa(\boldsymbol{A}_{tt}):=\left|\lambda_{max}(\boldsymbol{A}_{tt})\right|\left|\lambda_{min}(\boldsymbol{A}_{tt})\right|^{-1}$ of the translational part $\boldsymbol{A}_{tt}$ of the Hessian of the cost 
minimized by the Point-To-Plane ICP algorithm.
The eigenvector associated with the smallest eigenvalue of $\boldsymbol{A}_{tt}$ is the \textit{least observable} direction for translation estimation. As a consequence, larger the condition number $\kappa(\boldsymbol{A}_{tt})$ is, the more poorly constrained the optimization problem is in the translational part. More details are provided in our related work ~\cite{LION}.

\ph{Evaluation} 
We report the performance of the LiDAR-IMU fusion technique during the two tracks (A and B) of the Tunnel Competition at the DARPA Subterranean Challenge.
The LOCUS output (selected, for this evaluation, to be the pure scan-to-scan matching from the Generalized ICP \cite{segal2009generalized}) was fused at 10 Hz \comb{with the IMU}, and consequently the fused output of LION, could be provided at up to 200 Hz.
The sliding window of LION used here is 3 seconds and the factor graph optimization was tuned to use approximately 30\% of one CPU core of an i7 Intel NUC.
The root-mean-squared error (RMSE) for position ($\boldsymbol{t}$(m)) and attitude estimation ($\boldsymbol{R}$(rad)) and the percentage drift ($\boldsymbol{t}$(\%)) are reported in~\autoref{table:tunnel_stats}, where we include a comparison with WIO and LOCUS (Scan-to-Scan), with the global localization algorithm LAMP~\cite{LAMP} (presented in the following section) as ground truth.
The results highlight that fusing inertial data with the odometry from the front-end significantly reduces the drift of LiDAR's pure scan-to-scan matching. Additionally, LION reliably estimates the attitude of the robot, and guarantees a gravity-aligned output provided at IMU rate.
Last, in~\autoref{fig:slip_office}, we report the performance of the observability module in an indoor, office-like environment, characterized by long corridors.
The results show that when the observability module is not used (\autoref{fig:slip_office}, \textit{left}), motion-unobservability creates a \textit{LiDAR slip}, producing a position estimation error of $\approx 9$ m. When the observability module is used (\autoref{fig:slip_office}, \textit{left}), the switching-logic in HeRO switches to WIO instead of LION for the section of the corridor without LiDAR features.
The total error is $\approx 1$ m (``Before loop closure''). Improved state estimation (reduced drift in the output of HeRO) benefits the global mapping solution (\autoref{sec:lamp}), which can now correctly detect a loop closure (\autoref{fig:slip_office}  ``After loop closure''), further reducing the drift.

\begin{table}
\begin{centering}
\resizebox{\columnwidth}{!}{
\begin{tabular}{|c||c|c|c|c|c|c||c|c|c|c|c|c||}
\cline{2-13} \cline{3-13} \cline{4-13} \cline{5-13} \cline{6-13} \cline{7-13} \cline{8-13} \cline{9-13} \cline{10-13} \cline{11-13} \cline{12-13} \cline{13-13} 
\multicolumn{1}{c||}{} & \multicolumn{6}{c||}{\textbf{Track A}} & \multicolumn{6}{c||}{\textbf{Track B}}\tabularnewline
\hline 
 & \multicolumn{3}{c|}{\textbf{Run 1} (685 m, 1520 s)} & \multicolumn{3}{c||}{\textbf{Run 2} (456 m, 1190 s)} & \multicolumn{3}{c|}{\textbf{Run 1 }(467 m, 1452 s)} & \multicolumn{3}{c||}{\textbf{Run 2 }(71 m, 246 s)}\tabularnewline
\hline 
\hline 
\textbf{Algorithm} & $\boldsymbol{t}$(m) & $\boldsymbol{t}$(\%) & $\boldsymbol{R}$(rad) & $\boldsymbol{t}$(m) & $\boldsymbol{t}$(\%) & $\boldsymbol{R}$(rad) & $\boldsymbol{t}$(m) & $\boldsymbol{t}$(\%) & $\boldsymbol{R}$(rad) & $\boldsymbol{t}$(m) & $\boldsymbol{t}$(\%) & $\boldsymbol{R}$(rad)\tabularnewline
\hline 
\textbf{Wheel-Inertial}& 130.50&	19.05&	1.60&	114.00&	25.00&	1.28&	78.21&	16.75&	0.99&	6.91&	9.79&   0.12\tabularnewline
\hline 
\textbf{Scan-To-Scan} & 105.47&	15.40&	0.90&	18.72&	4.11&	0.18&	56.6&	12.14&	0.79&	4.55&	6.45&	0.27\tabularnewline
\hline 
\textbf{LION} & 56.92&	8.31&	0.36&	7.00&	1.53&	0.10&	17.59&	3.77&	0.27&	3.78&	5.36&	0.05\tabularnewline
\hline 
\end{tabular}
} %
\par\end{centering}
\caption{Estimation error of Wheel-Inertial Odometry, Scan-To-Scan matching and LION for two runs of the two tracks of the Tunnel competition, computed for one of the robots deployed.}
\label{table:tunnel_stats}
\end{table}

\subsection{Other Odometry Sources}
Apart from the LiDAR-inertial odometry estimator, NeBula consists of other robust and resilient estimation algorithms developed to provide a robust state estimation while navigating in perception-challenging environments.
Some examples are, for instance, a Contact-inertial odometry estimation~\cite{lew2019contact}, where contacts are exploited to produce zero velocity updates into a Kalman filter that is integrating IMU measurements during a dead-reckoning situation; or a Radar-inertial odometry~\cite{RIO2020}, which provides reliable ego-motion estimations even in the presence of obscurants thanks to the Radar signal properties.
The parallel combination of these heterogeneous estimation sources within the HeRO architecture provides a qualitative and robust state estimation that can be refined with a back-end algorithm providing global localization, as described in \autoref{sec:lamp}.
\section{Large-Scale Positioning and 3D Mapping}\label{sec:lamp}
 NeBula's Simultaneous Localization and Mapping (SLAM) solution, called LAMP (Large-scale Autonomous Mapping and Positioning) achieves low drift, multi-robot, multi-sensor SLAM over large scales in perceptually degraded conditions. LAMP produces a consistent global representation of an unknown environment, along with the associated covariances to enable uncertainty-aware solutions across the NeBula system (\autoref{fig:nebula_architecture}). In the context of the DARPA Subterranean Challenge, LAMP achieves the requirement for artifact localization error of less than $5$ m over multiple kilometers of traverse. In this section, we will outline the architecture of LAMP, and then describe our approach to multi-sensor SLAM. Finally, we present results from representative field tests.

\subsection{Subsystem Overview}
\begin{figure}[!ht]
  \centering
  \includegraphics[width=1.0\textwidth]{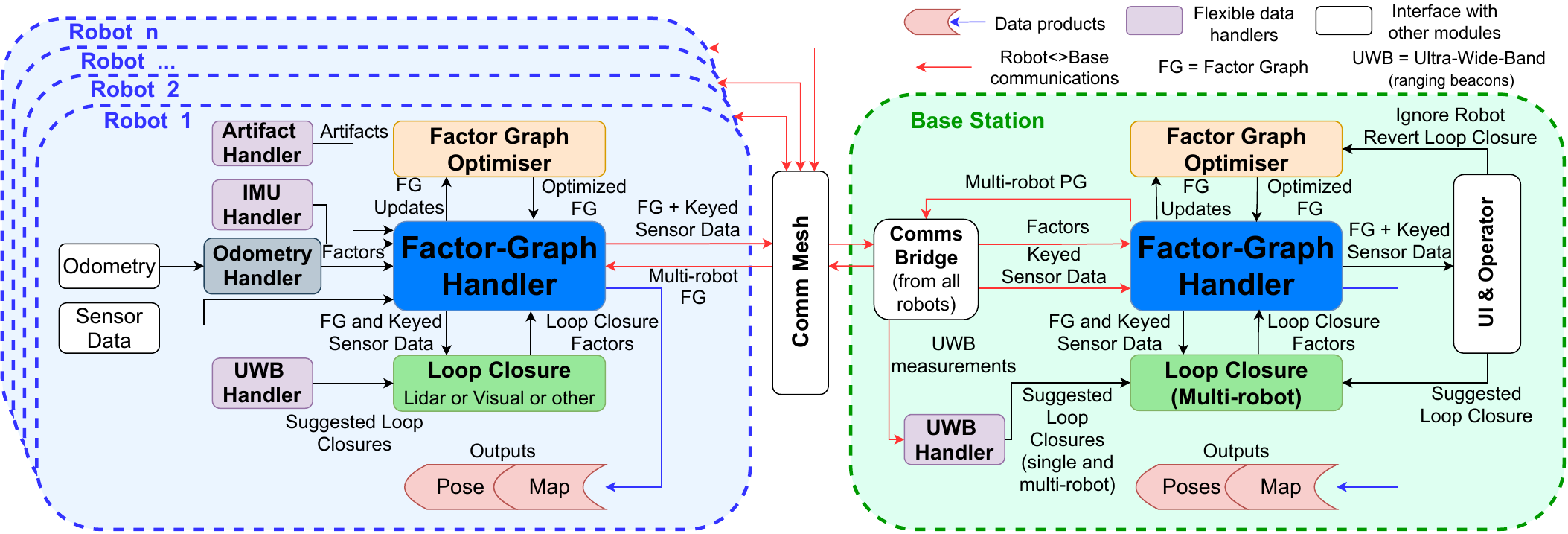}
  \caption{LAMP Architecture. Each robot maintains their own factor graph (FG), which can then be fused with a multi-robot team on the base station, using a centralized architecture. The LiDAR and camera data that is used for loop closures and map building (white "Sensor Data" box) is labelled Keyed Sensor Data, being linked with a specific pose-node. Handlers of artifact observations, IMU measurements, and Ultra-Wide-Band (UWB) signals all process data to add constraints to the factor graph. Our robust optimization approach runs in parallel to optimize both the robot and multi-robot factor graphs.}
  \label{fig:lamp_architecture}
\end{figure}

As outlined in~\autoref{fig:lamp_architecture}, LAMP is a factor-graph based SLAM solution, with the following key components: a) an adaptable odometry input that can process individual or fused odometry sources, such as HeRO~(\autoref{sec:state_estimation}), b) a multi-modal loop closure module, based on LiDAR, visual or semantic features, c) an outlier-resilient optimization of the factor graph, including multi-sensor inputs. 

The flow of data starts with the odometry and sensor inputs, which add factors to the graph on the robot. Parallel processes then run loop closure searches and factor-graph optimization. Next, the graph is sent to the base station. The base station merges graphs from each robot into a common multi-robot graph that is further optimized with the addition of inter-robot loop closures. The main output products of LAMP are a set of poses describing the robot trajectory and the location of artifacts, as well as a point-cloud map.

\begin{figure}[t]
  \begin{minipage}[c]{0.6\textwidth}
    \includegraphics[width=\textwidth]{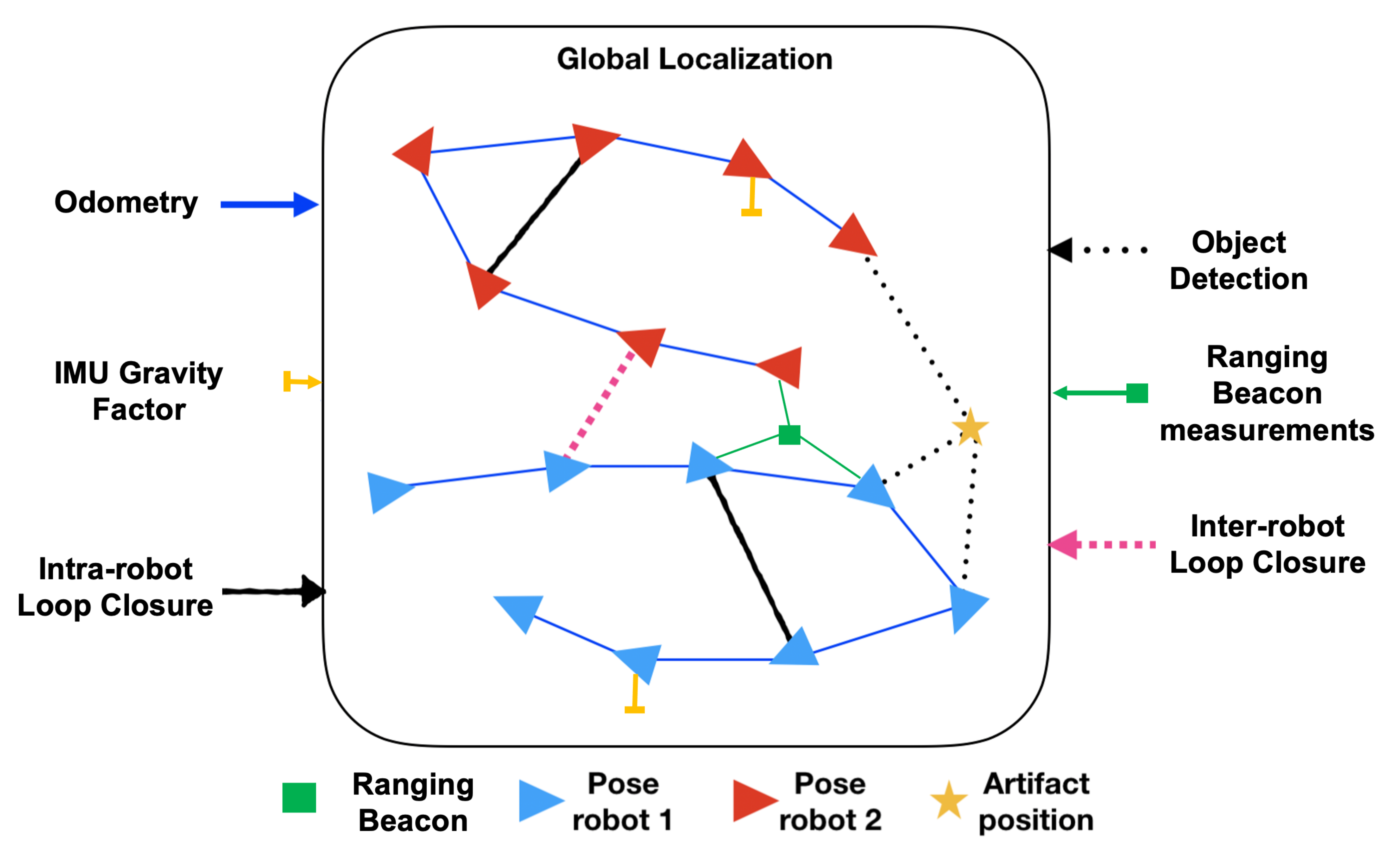}
  \end{minipage}\hfill
  \begin{minipage}[c]{0.35\textwidth}
    \caption{Graphical representation of the LAMP multi-robot pose-graph, including odometry factors, inter- and intra- robot loop closures, gravity factors, ranging beacon measurements and shared observations of landmarks (e.g. artifacts).}
  \label{fig:lamp_pose_graph}
  \end{minipage}
\end{figure}

\ph{Pose Nodes and Adaptable Odometry Input}
To make the factor-graph optimization computationally tractable over large-scale, long-term multi-robot exploration, LAMP utilizes a sparse graph of pose-nodes and edges (\autoref{fig:lamp_pose_graph}). The edges are obtained from an accumulation of odometry measurements between two consecutive nodes (odometry edges) or from translation and rotation estimates between non-consecutive nodes (loop closures edges, described below). A new pose-node and linking odometry edge is created after travelling more than a threshold translation or rotation. %
To address the challenge of perceptual degradation for these odometry edges, we use HeRO (\autoref{sec:state_estimation}) as the input odometry source. 

\ph{Multi-Modal Loop Closures}
A crucial capability to reduce the accumulated error in the robot trajectory is loop closure detection: the ability to correctly assert when a robot revisits a previously visited location or known landmark. Loop detection enables the computation of rigid body 3D transformations between non-consecutive pose-nodes that can be added as loop closure edges in the factor graph (\autoref{fig:lamp_pose_graph}). The multi-modal architecture of LAMP's loop closure module (\autoref{fig:lamp_architecture}) enables a robust and reliable system through the use of different sensing modalities. These loop closure sensing modalities include using lidar data~\cite{LAMP}, visual data~\cite{rosinol2020kimera} and semantic data~\cite{DARE-SLAM}. %

\ph{Multi-robot Fusion}
NeBula addresses the problem of multi-robot exploration of unknown environments by relying on LAMP's multi-robot architecture. This architecture is centralized, to leverage agents with greater computational resources (such as a base station), however for decentralized applications of NeBula we utilize the techniques described in~\cite{choudhary2017distributed}. 
In the centralized architecture the agent with the greatest computational capability serves as the central agent to fuse factor graphs constructed by individual robots into a consistent multi-robot graph, along with the associated sensor data (\autoref{fig:lamp_architecture}). The factor graphs are fused using the same multi-modal loop closure modules as on the single robot, but instead of searching for intra-robot loop closures, these modules search for inter-robot loop closures. To further improve localization accuracy, we leverage the computational power of the central agent to perform batch loop closure analysis across the entire graph. This analysis identifies and computes additional inter- and intra-robot loop closures to add to the multi-robot graph. The updated multi-robot global graph is then optimized, and periodically sent back to the robots, for each agent to have a consistent global representation of the environment for global planning (\autoref{sec:global_planning}).

\ph{Factor-Graph Optimization}
Our factor-graph optimization (Kimera-RPGO~\cite{rosinol2020kimera}) uses a robust outlier rejection approach to reject the erroneous loop closures that can occur when operating in perceptually degraded conditions, such as with obscurants and self-similar environments. Kimera-RPGO rejects erroneous loop closures by finding the largest consistent set of loop closures for each set of single robot and inter-robot loop closures, using a consistency graph and max clique search (an adaptation of~\cite{mangelson2018pairwise}). The loop closures that are not in the consistent set are discarded prior to optimization (see \cite{LAMP} and~\cite{lajoie2020door} for details). The updated factor graph is then optimized with a Levenberg–Marquardt solver that is implemented in GTSAM (Georgia Tech Smoothing and Mapping~\cite{dellaert2012factor}). 

\subsection{Additional Factors and Multi-Sensor Fusion}\label{sec:lamp:additionals}
\begin{figure}[!ht]
  \centering
  \includegraphics[width=0.45\textwidth]{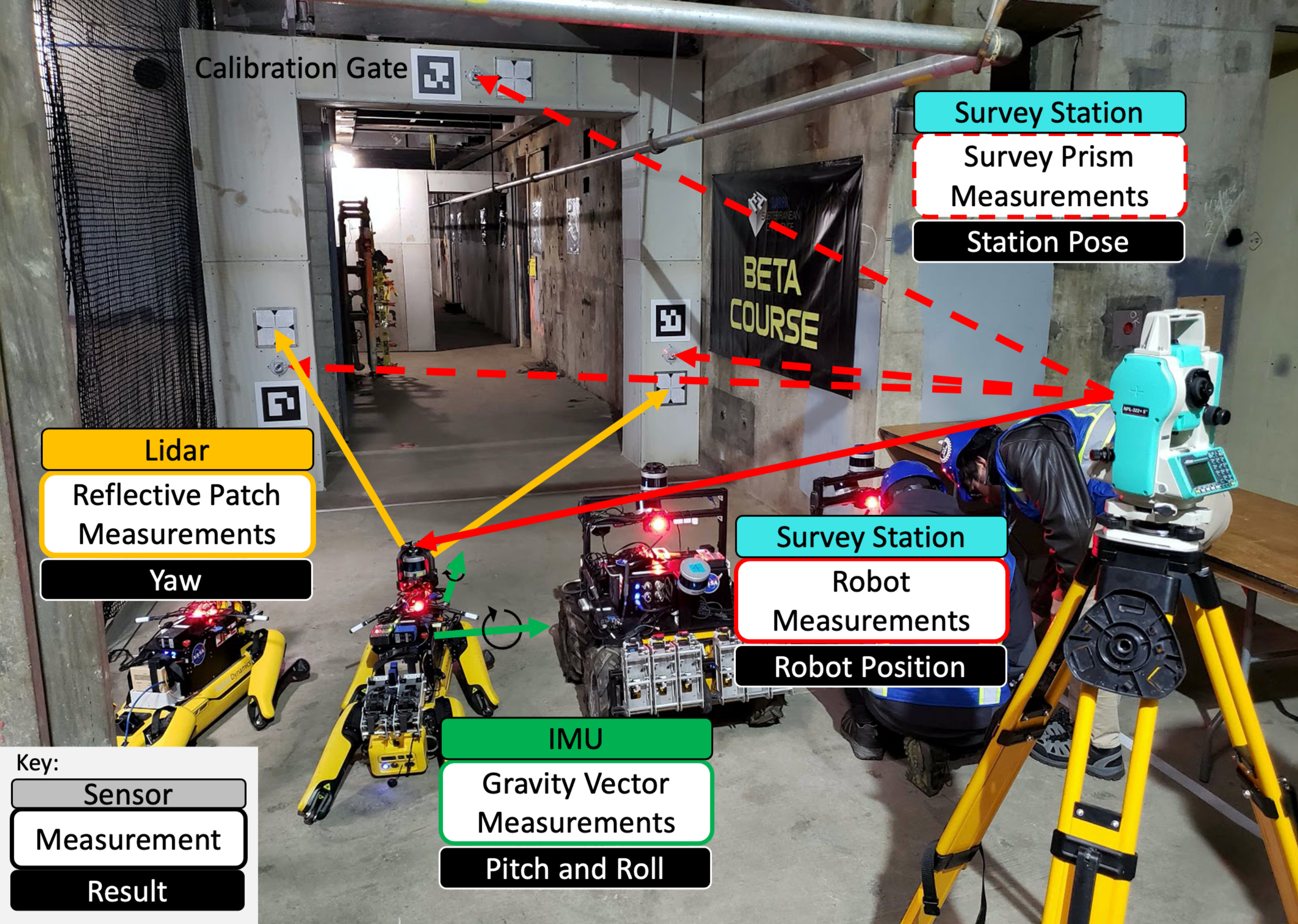}
  \caption{Example calibration to global frame from the DARPA Subterranean Challenge Urban Circuit. The coordinates of the survey prisms and reflective patches on the gate are provided and define the reference frame. A survey station measures the prisms to determine its pose in the global frame, after which it can measure the position of robots. LiDAR measurements of the reflective patches provide a yaw estimate and an IMU computes roll and pitch assuming gravity alignment of the reference frame.}
  \label{fig:fiducial_cal}
\end{figure}

LAMP fuses multiple sensing inputs into the factor graph (\autoref{fig:lamp_pose_graph}) to improve the robustness and accuracy of the SLAM solution. We present four examples here:

\begin{enumerate}
    \item \underline{IMU Gravity Factors}: When the robot is stationary, the accelerometers on the IMU can be used to obtain a robust estimate of the local gravity vector, which is added to the factor graph as a constraint on roll and pitch (yellow factors in \autoref{fig:lamp_pose_graph}). 
    \item \underline{Landmark Factors}: Measurements of distinct landmarks can either be used to detect loop closures or to directly provide constraints to the factor graph. These landmarks fall into two categories. 
    \begin{enumerate}
        \item \underline{Deployed Landmark Factors}: These landmarks include visual beacons, ranging beacons or retro-reflecting beacons and are deployed from a robot while exploring an unknown environment. For example, we have implemented deployable Ultra-Wide-Band (UWB) ranging beacons in our system (see \autoref{sec:hardware_static_asset} for hardware details, and~\cite{funabiki2020uwb} for algorithmic details). The signals from the beacons robustly and efficiently identify loop closures, to seed LiDAR- or vision-based alignment computations for single- and multi-robot teams (green node and factors in \autoref{fig:lamp_pose_graph}). 
        \item \underline{Environmental Landmark Factors}: Existing features in the environment such as signs, salient objects and the shape of junctions (e.g. \cite{DARE-SLAM}) can be used as landmarks. For example, we use observations of specific objects, such as backpacks and fire extinguishers (called artifacts in SubT), with sets of range-bearing observations (dashed black lines in \autoref{fig:lamp_pose_graph}) from the artifact relative-localization module (\autoref{sec:artifacts}). By fusing the object observations into the factor graph we also ensure the most up-to-date global location of those objects for situational awareness (and scoring in SubT). 
    \end{enumerate}
    \item \underline{Calibration Factors}: At the start of a mission, each robot is aligned with a global reference frame from a combination of LiDAR, IMU and survey station measurements of the robot and a calibration gate (e.g. \autoref{fig:fiducial_cal}). These initial calibration measurements, as well as any additional measurements generated during the mission, are added as constraints to the factor graph. 
\end{enumerate}

\subsection{Metric and Semantic Map Generation} 
LAMP builds both a geometric and semantic global map from sensor measurements attached to the nodes in the factor graph. Both maps are built by projecting sensor measurements into the global reference frame by using the latest, optimized state of the associated pose-nodes in the factor graph. For the geometric map, these sensor measurements (Keyed Sensor Data in \autoref{fig:lamp_architecture}) are either point clouds (from LiDAR, depth cameras or visual feature tracking) or local occupancy grids. In particular, structures like confidence-rich occupancy grids \cite{CRM}, allows for encoding the environmental uncertainty, which then can be used for perception-aware coverage planning and enabling SMAP-like behaviors \cite{CRM-planning}. For the semantic map, the sensor measurements are descriptive observations, such as detections of distinct objects (e.g. backpacks, survivors), semantic classifications of 3D spaces (e.g. doorways, stairs) or ambient measurements (e.g. temperature, pressure, gas concentration). The resulting 3D semantic map provides rich situational awareness to the operator of the robotic team, and can be the critical output data product of the overall system. The semantic map is especially important in the context of SubT, where the semantic map primarily consists of the globally localized artifact observations (both objects and ambient measurements), which is exactly the information needed for scoring (\autoref{sec:artifacts}).

\subsection{LAMP Performance}\label{sec:lamp_results}
\begin{figure}[!ht]
  \centering
  \includegraphics[width=1.0\textwidth]{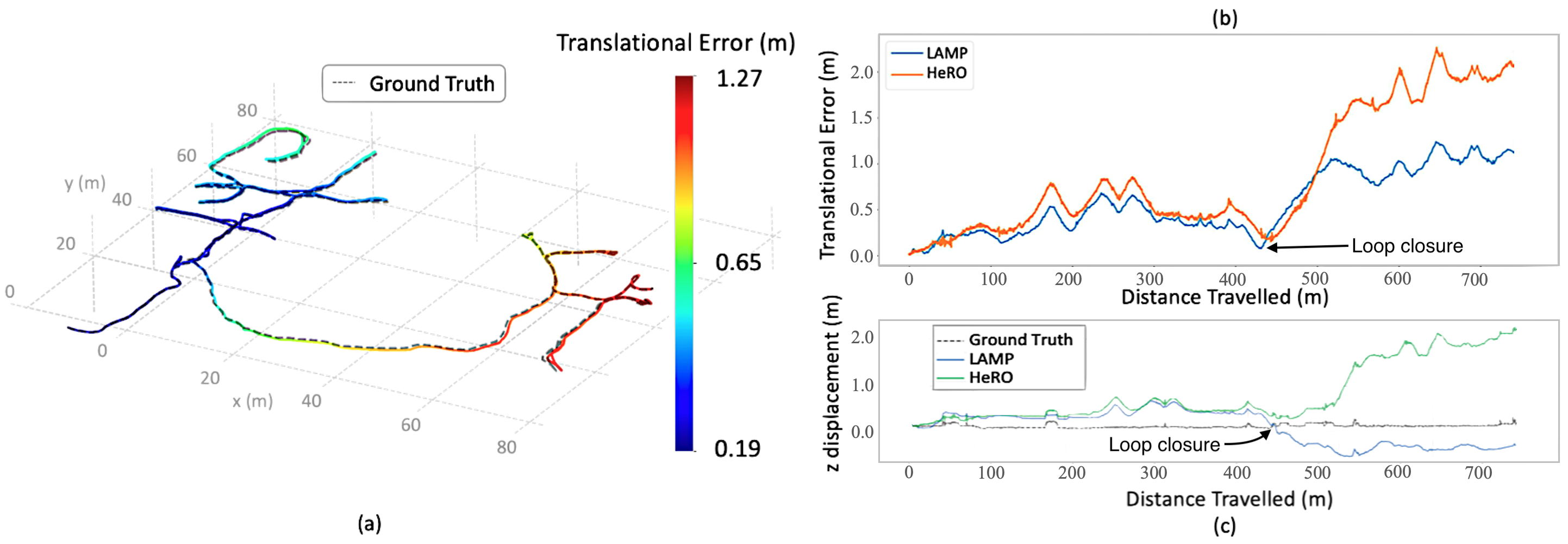}
  \caption{Single-robot LAMP performance with loop closures and IMU factors in the Urban Circuit of the DARPA Subterranean Challenge: (a) LAMP localization accuracy compared to DARPA-provided ground truth. (b) Translational error against distance travelled for LAMP, and the HeRO odometry input to LAMP. (c) $z$ trajectory against distance travelled for LAMP, HeRO and the ground truth. IMU factors included in LAMP help to constrain attitude drift and achieved improved performance in $z$, and overall, compared to the input odometry.}
  \label{fig:lamp_single_robot_results}
\end{figure}

\begin{figure}[!h]
  \centering
  \includegraphics[width=0.9\textwidth]{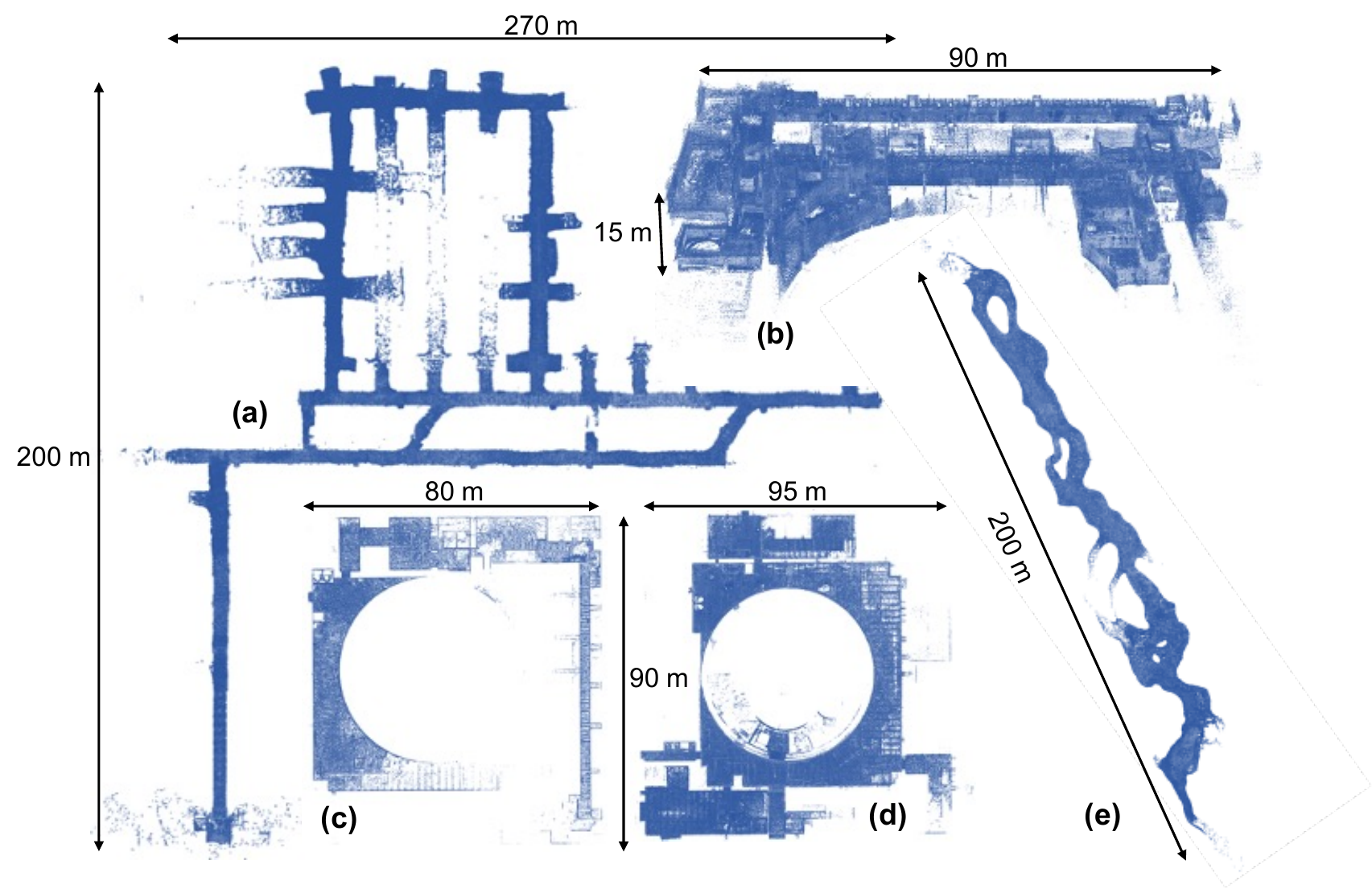}
  \caption{Maps from the five LAMP single-robot benchmark datasets. (a) Tunnel dataset from Bruceton mine on a Husky. (b) Multi-level urban dataset from Satsop Power Plant on a Spot robot. (c) and (d) Urban dataset from Satsop Power Plant on a Husky robot platform. (e) Cave dataset from Valentine Cave on a Spot robot. The ground truth is provided by DARPA for (a)-(d) and by survey scans for (e). 
  }
  \label{fig:lamp_benchmark}
\end{figure}

\begin{figure}[h!]
  \centering
  \includegraphics[width=0.85\textwidth]{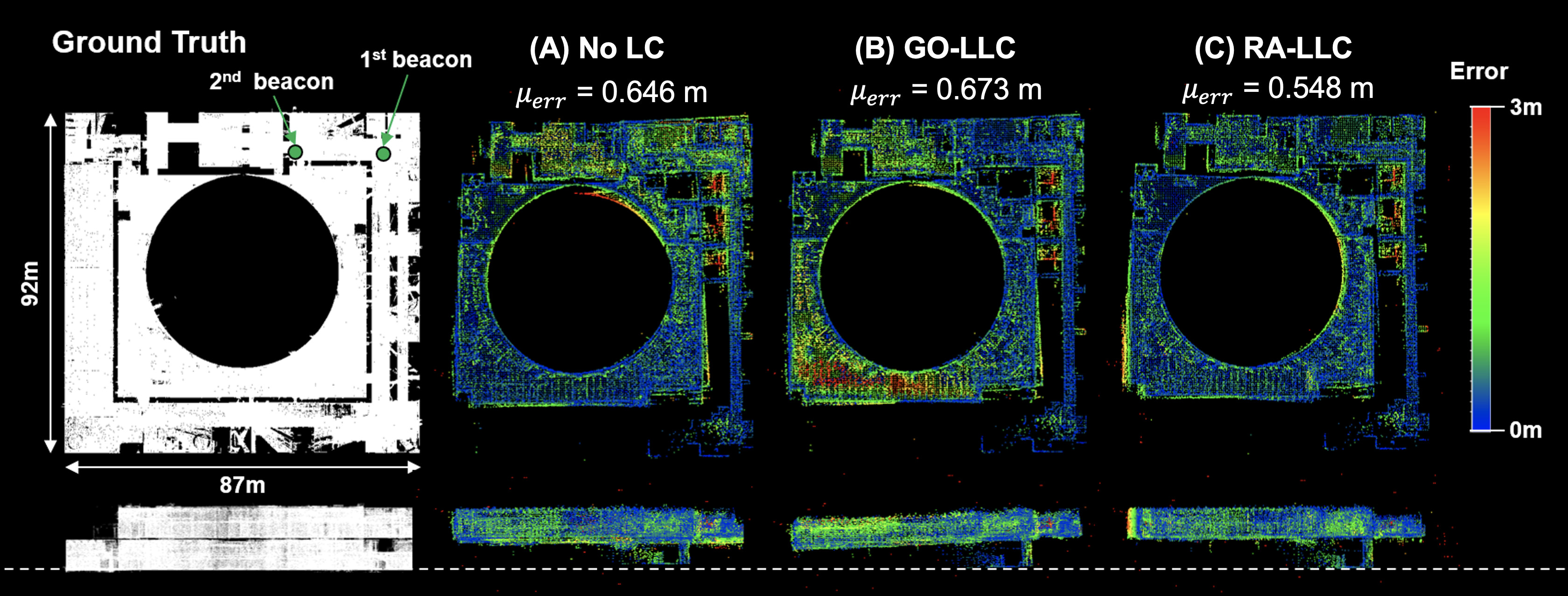}
  \caption{Multi-robot map accuracy results\rev{: (A) no loop closure (No LC); (B) geometric-only lidar loop closure (GO-LLC); and (C) range-aided lidar loop closure (RA-LLC).} In this test, data from two robots in the Urban Circuit of the DARPA Subterranean Challenge are used, with two deployed UWB beacons (as shown on the left overlaid on the DARPA provided ground-truth map). $\mu_{error}$ indicates the mean map error to the ground truth map.}
  \label{fig:lamp_uwb_multi-robot}
\end{figure}

The performance of LAMP on a single robot dataset from a husky robot equipped with 3 LiDARs is demonstrated in \autoref{fig:lamp_single_robot_results}. LAMP achieves error at less than $0.2\%$ of the distance travelled, with the IMU gravity factors assisting in reducing the $z$ error in the latter portion of the trajectory (\autoref{fig:lamp_single_robot_results}.c). Further single robot tests are summarized in \autoref{tab:lamp_benchmark_stats}, from the five benchmark datasets shown in \autoref{fig:lamp_benchmark}. These benchmarks show LAMP achieving the accuracy better than $5$m on all other than the tunnel and cave datasets, which have motion distorted LiDAR measurements. 
Multi-robot LAMP performance is demonstrated in \autoref{fig:lamp_uwb_multi-robot} with two huskies deployed in an urban environment. The results demonstrate the benefit of using UWB beacons (\autoref{fig:lamp_uwb_multi-robot}.c) compared to using \rev{pure LiDAR-based loop closures.} 
Further results are presented in \autoref{sec:experiments}. Please refer to~\cite{LAMP} for detailed results in tunnel environments, including the impact of loop closures, \cite{funabiki2020uwb} for results using UWB beacons for a single robot and \cite{morrell2020robotic} for single robot cave exploration results.

\begin{table}[!b]
\begin{tabular}{|l|l|l|r|r|r|r|r|}
\hline
        & \multicolumn{3}{l|}{Dataset Characteristics}             & \multicolumn{2}{l|}{Absolute Transl. Error}             & \multicolumn{2}{l|}{Absolute Rot. Error}                     \\ \hline
Dataset & Robot & Environment & \multicolumn{1}{l|}{Dist. (km)} & \multicolumn{1}{l|}{Max (m)} & \multicolumn{1}{l|}{Mean (\%)} & \multicolumn{1}{l|}{Max (deg)} & \multicolumn{1}{l|}{Mean (deg/m)} \\ \hline
(a)   & Husky & Tunnel      & 1.65                               & 9.7                          & 0.93\%                         & 5.3                            & 0.006                             \\ \hline
(b)    & Spot  & Urban       & 0.65                               & 2.2                          & 0.46\%                         & 5.0                            & 0.019                             \\ \hline
(c)    & Husky & Urban       & 0.62                               & 3.5                          & 0.42\%                         & 3.9                            & 0.011                             \\ \hline
(d)    & Husky & Urban       & 0.75                               & 0.9                          & 0.19\%                         & 1.8                            & 0.006                             \\ \hline
(e)    & Spot  & Cave        & 0.6                                & 10.6                         & 1.68\%                         & 6.0                            & 0.020                             \\ \hline
\end{tabular}
\caption{Performance statistics for LAMP operating on single-robot benchmark datasets from different robots and in different environments (\autoref{fig:lamp_benchmark})}
    \label{tab:lamp_benchmark_stats}
\end{table}

\section{Semantic Understanding and Artifact Detection} \label{sec:artifacts}
Semantic understanding of the environment and detecting objects of interest and artifacts are important capabilities to enable higher levels of robotic autonomy in unknown environments. Semantic mapping and artifact detection are among the main components of the NeBula autonomy framework. This section discusses NeBula’s solution for detecting, localizing, and visualizing objects of interest on heterogeneous robots with different sensor configurations. Here, we focus on both (\emph{i}) static objects, with clear visual, thermal, or depth signature, and (\emph{ii}) spatially-diffused phenomena such as gas propagation and WiFi signal. The pipeline is explained in detail in~\cite{artifact2020}, with a summary and recent extensions detailed here.

\pr{Requirements} 
The object detection system needs to (\emph{i}) make detections in real-time across multiple sensor modalities; (\emph{ii}) permit high-accuracy localization; \rev{(\emph{iii}) adjust the sensor configuration based on the detection and localization confidence; and (\emph{iv}) apply filtering to present the most likely detection candidates to the mission supervisor (when a communication link is established)}. While the method presented in this section is general,  in the context of the DARPA Subterranean challenge, we focus on a set of predefined object types including gas sources (e.g., CO\textsubscript{2} source) and man-made objects such as fire extinguisher, drill, rope, helmet, survivor manikin, backpack, vent, and cell phone\revv{~\cite{agrawalGroundUpDesign,darpaJFR}}. \rev{Object signatures exist in one or more modalities:} visual, thermal, depth, WiFi, audio, and chemical.

\pr{Architecture}
\begin{figure}[t]
    \centering
    \includegraphics[width=0.99\linewidth]{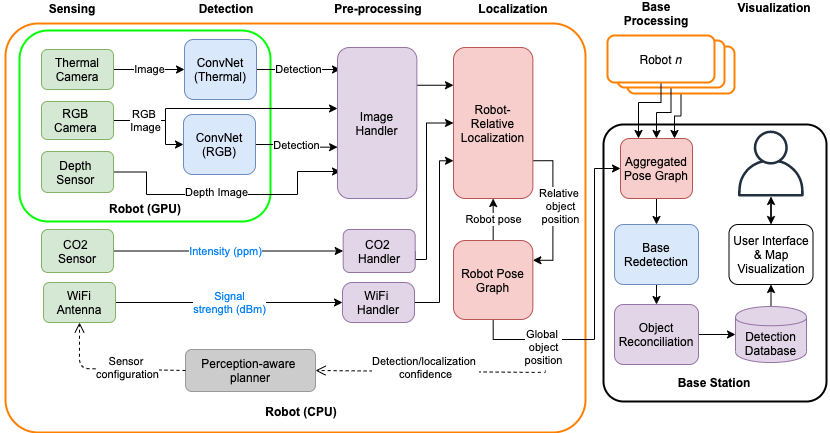}
    \caption{Multi-robot, multimodal object detection, localization and visualization pipeline.}
    \label{fig:artifactDetectionArchitecture}
\end{figure}
\autoref{fig:artifactDetectionArchitecture} shows the proposed object detection, localization, and visualization pipeline. We break down the underlying object detection problem into two stages: (1) an image-based object detection pipeline to first find the object; and (2) a relative localization filter applying projective geometry to the detection to estimate its position explicitly. By splitting the detection and localization tasks, we can utilize high quality and fast detection from existing algorithms and apply them to generic camera types in our relative localization approach. For temporally-static objects, NeBula relies on multi-modal detection and, when available on robots, leverages visual cameras, depth measurements, and thermal cameras. 

\ph{Spatially Diffuse Phenomena} For temporally-dynamic and spatially-diffused phenomena, we rely on source seeking methods based on gas sensors 
and WiFi sensing
. The detection confidence provides the uncertainty assessment to the perception-aware planner. The planner motivates the sensor to adjust the configuration to make new measurements with higher fidelity, leading to more accurate detections. We discuss the three stages of detection, localization, and base station processing in the rest of this section. 

\pr{Detection}
For visually-observable objects, detections are made in both the color and thermal spectra using state-of-the-art convolutional neural networks (CNNs). CNN produces a bounding box on the image to pass to the relative localization module. NeBula relies on different CNN implementations to adapt to various processing capabilities. On ground robots, a YOLO Tiny~\cite{redmon2018yolov3,bochkovskiy2020yolov4} variant is used, leveraging GPU hardware (e.g., Nvidia Jetson Xavier) to run in real-time on multiple cameras (see~\autoref{sec:hardware}). On drones, a MobileDet variant~\cite{xiong2020mobiledets} is used, modified to run on a Google EdgeTPU.
To achieve sufficient detection performance for our specific application, we fine-tuned the detection networks with the appropriate domain-specific dataset. In the context of SubT, we have produced more than 40,000 annotations of the following objects: fire extinguisher (12.26 \text{\%}), drill (9.98\text{\%}), rope (17.90\text{\%}), helmet (19.29\text{\%}), survivor (19.84\text{\%}), and backpack (20.72\text{\%}). Our training is focused on maximizing recall that increases true positives. This is followed by an outlier rejection method using range, color, and size filters, reducing the false positives. For more details on training methods on this data, see~\cite{artifact2020}. 

\pr{Localization} The detection networks produce 2D bounding boxes (within image) that are combined with depth measurements, used to compute the position of the artifact relative to the robot. The depth measurements can be obtained from multiple sources: depth cameras (such as from an RGB-D camera), LiDAR scans mapped into the camera frame, or a size-based projection, where the depth is computed such that the bounding box, when projected to 3D at that depth, matches the expected size of the object. These methods are detailed in~\cite{artifact2020}. For robustness to sensor failure, each method is run in parallel, with the highest priority method (LiDAR, then depth, then size projection) used if the corresponding sensor is available. All methods jointly filter multiple detections to produce a combined relative location reported to the LAMP module (\autoref{sec:lamp}) to compute the global location before sending the report to the base station. The relative location can also be computed without depth measurements, as an additional back-up, and on systems without depth cameras or LiDAR (such as drones). In this case, we use a monocular based tracking approach over an image sequence (detailed in~\cite{ramtoula2020mslraptor}).

\begin{figure}[t]
    \centering
    \includegraphics[width=0.95\linewidth]{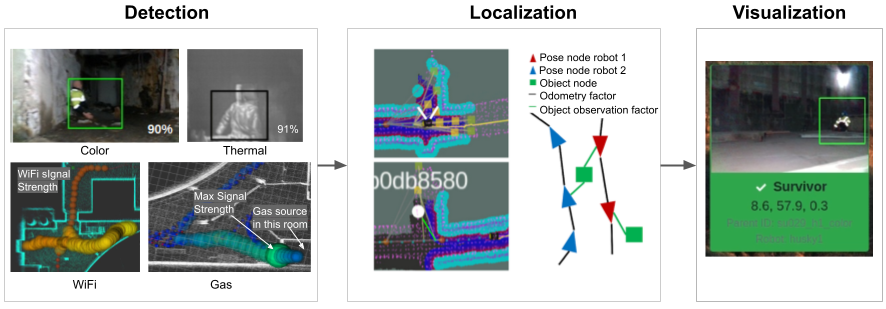}
    \caption{Visualization of representative parts of the object detection pipeline. In the left pane, a sample of detections for each of the four modalities (color, thermal, WiFi, and gas) is presented. In the middle pane, the addition of a natural landmark observation into the pose graph is depicted. The right pane shows the final operator view.}
    \label{fig:ArtifactPipeline}
\end{figure}

\pr{Multi-modal multi-robot artifact reconciliation} The artifact reports from each of the robots are further processed on robots with more powerful computational resources (or on base station). These reports include:(\emph{i}) detection class; (\emph{ii}) detection confidence; (\emph{iii}) reference RGB/thermal image; (\emph{iv}) bounding box\rev{; and (\emph{v}) location estimate}. The base station processes the reports, rejecting outliers and matching observations of the same artifact instances \rev{from other agents or previous visits}. To reduce the number of false positive reports, a larger and more performant detection network (YOLOv4~\cite{bochkovskiy2020yolov4}) is used to update the detection confidence of each report. Then, the report is compared to previous reports of the same class to identify repeat observations \rev{and observations of the same artifact instance reported by other agents}. This comparison uses both location proximity and a comparison of NetVLAD visual image descriptors~\cite{arandjelovic2016netvlad}. The highest confidence report of each object is then saved to a database and visualized for review by the human supervisor using the mission executive interface (see ~\autoref{fig:ArtifactPipeline}-Visualization block). When needed, to increase the confidence on the detected semantics, the perception-aware motion planner seeks new measurements, e.g., from a closer or better angle to the target, or by sending a different robot with complementary sensors to get multiple readings from the target.

\subsection{Spatially Diffuse Localization} 
To detect and locate spatially diffuse phenomena, such as gas leaks and WiFi sources, the robotic team is leveraged as a mobile sensor network, with distributed and moving ambient sensor measurements. Signal strength (e.g. CO\textsubscript{2} concentration or WiFi RSSI) is recorded at every robot position and is (\emph{i}) used to augment the global 3D semantic map (\autoref{fig:ArtifactPipeline}-Detection block); and (\emph{ii}) processed to produce an initial location estimate at the area of peak signal strength. The combination of the spatially informative semantic map with an initial location estimate seeds a local search for source locations, based on the 3D geometry. Automation of this local search is ongoing work. In tests presented here, information is sent to the base station for displaying to the operator, who performs the local search on inspection of the metric-semantic map. 
\begin{figure}
    \centering
    \includegraphics[width=0.95\linewidth]{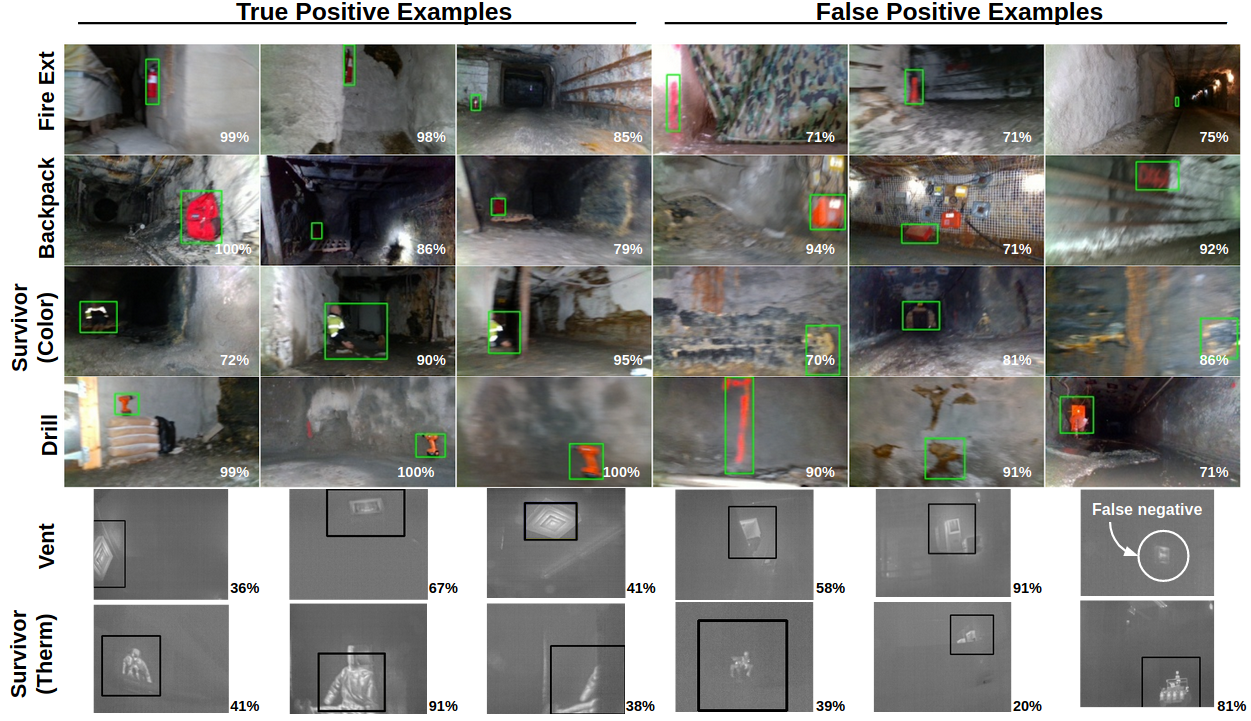}
    \caption{Examples of true and false positive detections of visual artifacts}
    \vspace{-5mm}
    \label{fig:tp_fp_urb2}
\end{figure}

\pr{Detection performance}
\autoref{fig:tp_fp_urb2} shows examples of a true and false positive detection for each visual artifact type. We observed that spray paint markings and existing equipment in the environments, which share the same gross features as the target objects, are incorrectly picked up. For spatially-diffuse detection (gas and WiFi), we extrapolate the source location by measuring the signal strength gradient and move the robot in directions that increase the detection confidence (see \autoref{fig:ArtifactPipeline}-Detection block).

\section{Risk-aware Traversability and Motion Planning} \label{sec:traversability}
A fundamental component of NeBula is its risk-aware traversability and motion planning (see~\autoref{fig:nebula_architecture}). This component, which we call STEP (Stochastic Traversability Evaluation and Planning), allows the robots to safely traverse extreme and challenging terrains by quantifying uncertainty and risk associated with various elements of the terrain. In this section, we briefly discuss NeBula’s traversability analysis, motion planning, and controls. For more details please see \cite{thakkur2020} and \cite{fan2021step}.

\begin{figure*}[ht]
    \centering
    \includegraphics[width=\linewidth]{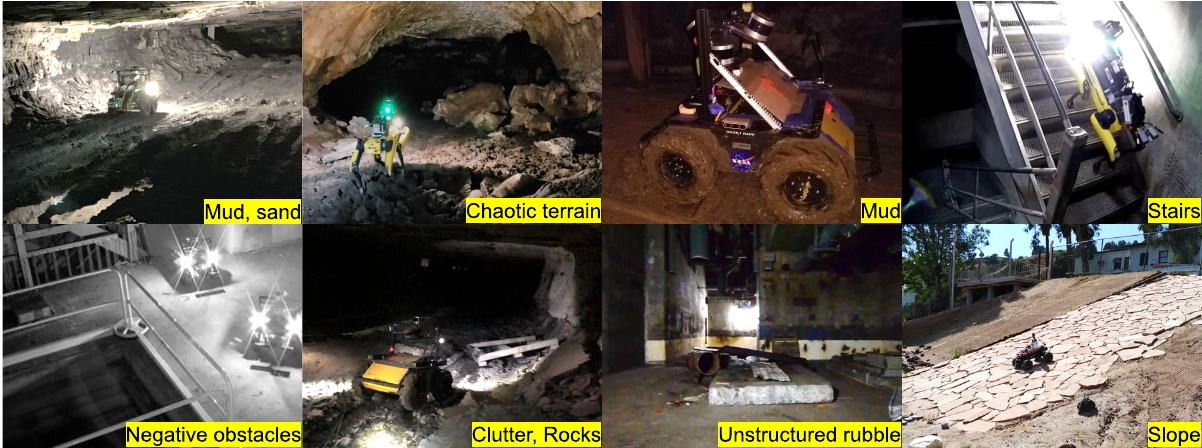}
    \caption{Mobility-stressing elements commonly found during testing in various tunnel, urban, and cave environments, including 800ft underground at Arch Mine in Beckley, WV, Valentine Cave at Lava Beds National Monument, CA, Satsop power plant in Elma, WA, and Mars Yard at JPL, Pasadena, CA.}
    \label{fig:TravTerrain}
\end{figure*}

\rev{\subsection{Design Philosophy}}
\pr{Challenges in extreme terrain motion planning}
Unstructured obstacle-laden environments pose large challenges for ground roving vehicles with a variety of mobility-stressing elements. Common assumptions of a benign world with flat ground and clearly identifiable obstacles do not hold; Environments introduce high risks to robot operations, containing difficult geometries (e.g. rubble, slopes) and non-forgiving hazards (e.g. large drops, sharp rocks) \cite{kalita2018path,leveille2010lava}. Additionally, subterranean environments pose unique challenges, such as overhangs, extremely narrow passages, etc. See \autoref{fig:TravTerrain} for representative terrain features. Determining where the robot may safely travel has several key challenges: (\textit{i}) Localization error severely affects how sensor measurements are accumulated to generate dense maps of the environment. (\textit{ii})
 Sensor noise, sparsity, and occlusion induces large biases and uncertainty in mapping and analysis of traversability. (\textit{iii}) The combination of various risk regions in the environment create highly complex constraints on the motion of the robot, which are compounded by the kinodynamic constraints of the robot itself.

\begin{figure*}[h!]
    \centering
    \includegraphics[width=\linewidth]{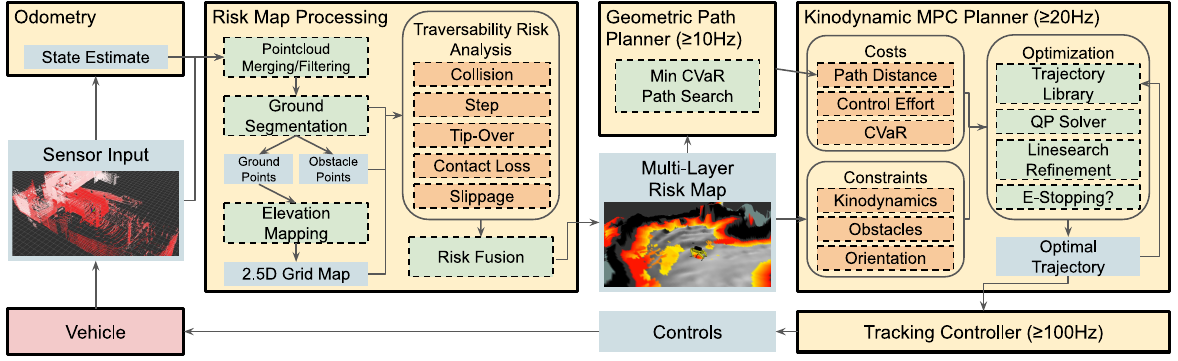}
    \caption{Traversability and Motion Planning Architecture Overview.
    }
    \label{fig:TravOverview}
\end{figure*}

\pr{System Architecture}
To address these issues, we develop a risk-aware traversability analysis and motion planning method, which 1) assesses the traversability of terrains at different fidelity levels based on the quality of perception, 2) encodes the confidence of traversability assessment in its map representation, and 3) plans kinodynamically feasible paths while considering mobility risks. \autoref{fig:TravOverview} shows an overview of the local motion planning approach. The sensor input (pointcloud) and odometry is sent to the risk analysis module, evaluating the traversability risk with its estimation confidence. The generated risk map is used by hierarchical planners consisting of a geometric path planner and a kinodynamic MPC (Model Predictive Control) planner. The planners replan at a higher rate to react to the sudden changes in the risk map. The planned trajectory is executed with a tracking controller, which sends a velocity command to the platform.

\pr{Robot agnosticism}
Our approach is highly extensible and general to our different ground robot types, requiring only a change in the dynamics model of the system. Moreover, using this approach, we are able to specify a wide array of constraints and costs, such as limiting pitch or roll of the vehicle on slopes, preferring one direction of motion, keeping some distance from obstacles, fitting through narrow passages, or slowing down / stopping around risky areas.  This flexibility has proven important in achieving robust navigation across the extreme traversability challenges encountered in highly unstructured environments (see \autoref{table:config_exp_platforms}).

\begin{table}[t]
\centering
\includegraphics[width=\textwidth]{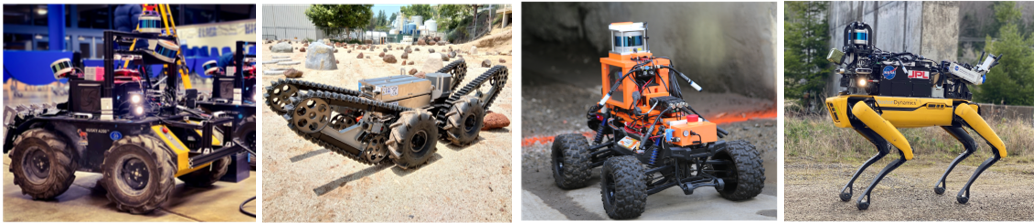}
\begin{tabular}{ | c || c | c | c | c | }
 \hline
 \textbf{Specification} & \textbf{Skid-steer$^{1,2}$} & \textbf{Tracked$^3$} & \textbf{Ackermann$^2$} & \textbf{Quadruped$^1$}\\
 \hline
 \textbf{Model} & Husky A200 & Telemax Pro & X-Maxx & Spot \\
 \textbf{Max. speed} & 1.0 m/s & 1.1 m/s & 22 m/s & 1.6 m/s \\
 \textbf{Steering} & Skid-steer & Skid-steer & Ackermann & Gait \\
 \hline
 \textbf{Distance traveled} & 5.98 km & 0.76 km & 0.39 km & 2.85 km \\
 \textbf{Avg. speed during traverse} & 0.93 m/s & 0.57 m/s & 0.83 m/s & 0.65 m/s \\
 \textbf{Autonomous recoveries / km} & 6.9 & 0.16 & 0.0 & 13.0\\
 \textbf{Critical failures / km} & 0.2 & 0.0 & 0.0 & 1.1 \\
 \hline
\end{tabular}

\caption{Specification of the different robots.  Superscript on the robot type indicates location/source of data used for computing statistics: $^1$DARPA SubT Urban Competition, $^2$DARPA SubT Tunnel Competition, $^3$Beckley Exhibition Coal Mine, WV.}
\label{table:config_exp_platforms}
\end{table}

\rev{\subsection{Uncertainty-aware Traversability}}
\rev{\pr{The importance of considering uncertainty}}
Precise traversability analysis and motion planning relies heavily on sensor measurements and localization.  However, the quality of state estimation can often degrade, especially in perceptually challenging environments such as tunnels, mines, and caves.  Additionally, sensors are subject to noise of various types, as well as occlusion, restricted field of view, etc.  Therefore a key idea is to incorporate uncertainty-awareness into our mapping for traversability and motion planning. We accomplish this by a multi-fidelity mapping approach in which we weigh more strongly high-confidence information from recent sensor measurements which are closer to the robot. Older and farther sensor measurements from the body of the robot are associated with higher uncertainties and decay more quickly. By aggregating these sensor measurements in an uncertainty-aware way, we create a robust and resilient belief-aware local map which is then used for traversability analysis. To reduce uncertainty and achieve higher levels of resiliency, we rely on multi-sensor high-FOV measurements \rev{(dense depth camera data and LiDAR data)} to efficiently update local maps and reduce sensitivity to localization uncertainties (similar, related work include  \cite{apfpf, hines_attitude}).

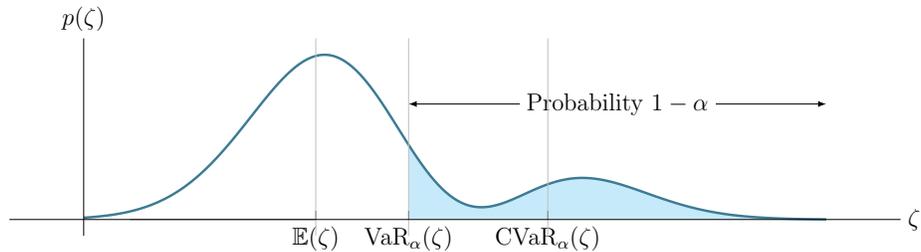
\begin{figure} 
\centering{
    \resizebox{0.75\textwidth}{!}{
\begin{tikzpicture}
\tikzstyle{every node}=[font=\large]
\pgfmathdeclarefunction{gauss}{2}{%
  \pgfmathparse{1000/(#2*sqrt(2*pi))*((x-.5-8)^2+.5)*exp(-((x-#1-6)^2)/(2*#2^2))}%
}

\pgfmathdeclarefunction{gauss2}{3}{%
\pgfmathparse{1000/(#2*sqrt(2*pi))*((#1-.5-8)^2+.5)*exp(-((#1-#1-6)^2)/(2*#2^2))}
}

\begin{axis}[
  no markers, domain=0:16, range=-2:8, samples=200,
  axis lines*=center, xlabel=$\zeta$, ylabel=$p(\zeta)$,
  every axis y label/.style={at=(current axis.above origin),anchor=south},
  every axis x label/.style={at=(current axis.right of origin),anchor=west},
  height=5cm, width=17cm,
  xtick={0,5,7,10}, ytick=\empty,
  xticklabels={$0$, , , ,},
  enlargelimits=true, clip=false, axis on top,
  grid = major
  ]
  \addplot [fill=cyan!20, draw=none, domain=7:15] {gauss(1.5,2)} \closedcycle;
  \addplot [very thick,cyan!50!black] {gauss(1.5,2)};
 
 \pgfmathsetmacro\valueA{gauss2(5,1.5,2)}
 \draw [gray] (axis cs:5,0) -- (axis cs:5,\valueA);
  \pgfmathsetmacro\valueB{gauss2(10,1.5,2)}
  \draw [gray] (axis cs:4.5,0) -- (axis cs:4.5,\valueB);
    \draw [gray] (axis cs:10,0) -- (axis cs:10,\valueB);

 \draw [gray] (axis cs:1,0)--(axis cs:5,0);
 \node[below] at (axis cs:7.0, -0.1)  {$\mathrm{VaR}_{\alpha}(\zeta)$}; 
\node[below] at (axis cs:5, -0.1)  {$\mathbb{E}(\zeta)$}; 
\node[below] at (axis cs:10, -0.1)  {$\mathrm{CVaR}_{\alpha}(\zeta)$};
\draw [yshift=2cm, latex-latex](axis cs:7,0) -- node [fill=white] {Probability~$1-\alpha$} (axis cs:16,0);
\end{axis}
\end{tikzpicture}
}
\caption{Comparison of the mean, VaR, and CVaR for a given risk level $\alpha \in (0,1]$. The axes denote the values of the stochastic variable $\zeta$, which in our work represents traversability cost. The shaded area denotes the $1-\alpha\%$ of the area under $p(\zeta)$.  $\mathrm{CVaR}_{\alpha}(\zeta)$ is the expected value of $\zeta$ under the shaded area.}
\label{fig:cvar}
}
\end{figure}

\begin{figure*}[h!]
    \centering
    \includegraphics[width=\linewidth]{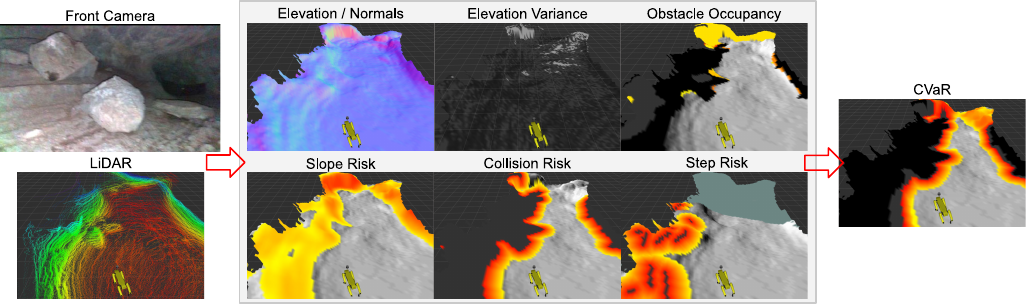}
    \caption{Traversability risk analyses which consider different sources of risk, as well as uncertainties, and fuses them into one CVaR metric costmap.  Left:  Raw sensor measurements are used to construct a pointcloud map.  Middle:  Statistical properties of the map are identified as various sources of traversability risk.  Right:  Risk sources are fused into one CVaR metric, which will be used for planning.}
    \label{fig:TravAnalysis}
\end{figure*}

\pr{Traversability risk analysis}
To assess traversability risks, we utilize the constructed multi-fidelity local map. A ground segmentation method \cite{himmelsbach2010fast} is applied to the merged pointcloud to filter obstacle and ceiling points. The ground pointcloud is used to build a 2.5D elevation map for efficient query of terrain geometry. The elevation map and segmented pointclouds are used to assess the risk of traverse from various perspectives including collision, tip-over, traction loss, and negative obstacles.  Individual risk analysis is fused into a single risk value estimate with a confidence value, and sent to the planning module (\autoref{fig:TravAnalysis}).  We define this risk value in terms of CVaR (Conditional Value-at-Risk) metric (\autoref{fig:cvar}), which quantifies the severity of the risk of a given path \rev{given all uncertainties $\zeta$ from the traversability risk analysis} according to a desired threshold of probability \rev{$\alpha$}. 
This threshold can be changed by mission-level decision making in order to vary the level of acceptable risk during the mission. \rev{This risk metric is particularly useful as it captures tail events with low probability of occurrence, which may have high consequences on the success of the mission and should be taken into account.  
By approximating all uncertainties with a Gaussian distribution, the CVaR is efficiently evaluated to account for different types of terrain \cite{fan2021step}.}

\pr{Semantic traversability factors}
In addition to geometric traversability analyses, we can also identify certain terrain features semantically and incorporate them into the risk map.  Features such as water are sometimes identifiable in LiDAR pointclouds due to differences in reflectivity. Other features such as stairs can be identified using computer vision methods, which locate stairs semantically and also identify stair slope, angle, and height from known geometric priors.  This information is useful for identifying non-geometric mobility risks, as well as notifying our planners to approach certain hazards differently (e.g. walking down stairs). 

\begin{figure}[t!]
    \centering
    \includegraphics[width=0.7\linewidth]{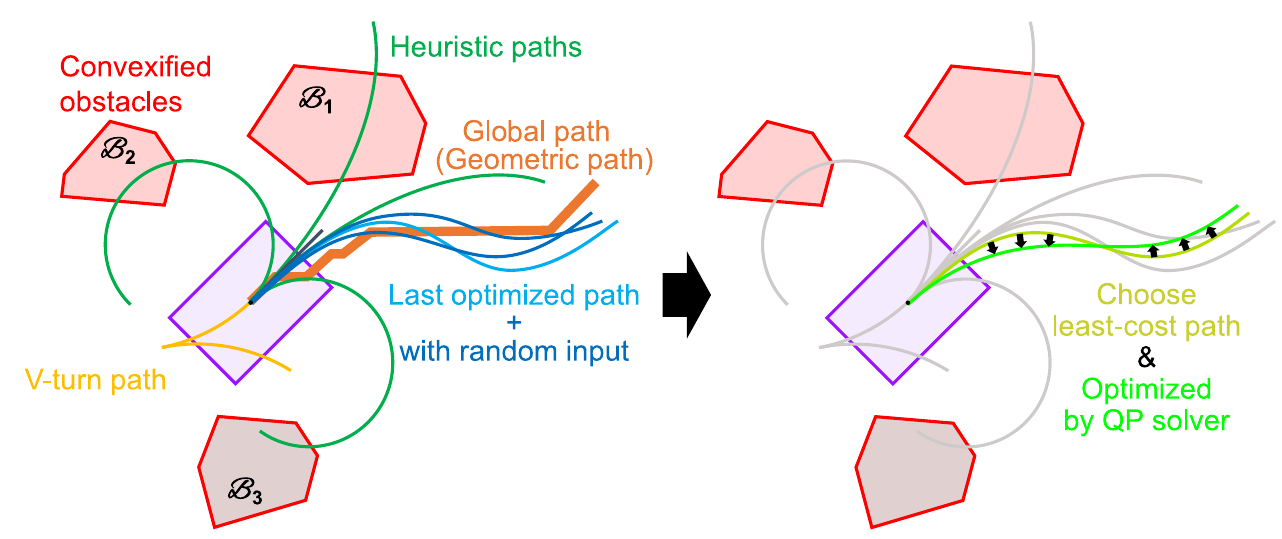}
    \caption{Schematic of our risk-aware kinodynamic trajectory optimization approach, which combines a trajectory library-based search with convex optimization.  \rev{Global path is determined by A* over a 2D grid.  Heuristic paths (including v-turns, random input, etc.) are not necessarily feasible but are used to initialize the local trajectory optimization.}  Obstacles, constraints, and costs are convexified \rev{about the best initial path} and used to \rev{iteratively} solve convex Quadratic Programming (QP) optimization problem\rev{s in real-time}.}
    \label{fig:traj_vis}
\end{figure}

\rev{\subsection{Uncertainty-aware traversability analysis and motion planning}}
\pr{Efficient Risk-aware Kinodynamic Planning}
Using the computed CVaR metric values on the map, we must search for a path which minimizes these values.  This is done in a two-stage hierarchical fashion.  The first stage operates on longer distances (40 m) and takes into account positional risk. \rev{Using an A* algorithm over a 2D grid, this first stage yields a global geometric plan that minimizes the risk of this path.} 
The second stage operates on shorter distances (8 m) and searches for a kinodynamically feasible trajectory that minimizes CVaR, while maintaining satisfaction of various constraints including obstacles, dynamics, orientation, and control effort. This kinodynamic planner operates in a model predictive control (MPC) fashion\rev{, and} is based on a combination of stochastic trajectory optimization and gradient-based convex optimization techniques \cite{fan2021step}\rev{.  It runs efficiently in real-time at 20--50 Hz, requiring roughly $\sim20\%$ of one CPU core, As shown in \autoref{fig:traj_vis}, this accounts for a rich set of heuristics, making it robust to local minima}. Once a trajectory is optimized, it is sent to an underlying tracking controller for execution on the platform.

\pr{Recovery Behaviors}
In the real world, failures are unavoidable. While our risk-aware traversability and planning architecture is meant to reduce the probability of failures, they do happen from time to time, as a consequence of failure in localization, undetected edge cases in traversability, hardware failures, or unknown unknowns.  As a last line-of-defense, we design behaviors to recover the system from non-fatal failures. \rev{Recovery behaviors are autonomously executed when we locally detect that a commanded motion is not being followed, or no valid and safe path is found to move away from the current position.}  These behaviors including clearing/resetting the local traversability map, increasing the allowable threshold of risk (to try to escape an untraversable area), and moving the robot in an open-loop fashion towards the direction of maximum known free space.  In most cases these recovery behaviors are sufficient to recover the robot from a stuck condition, as long as the robot has not suffered a catastrophic failure.

\pr{Learning and Adaptation}
Over the course of a mission we often see changes in vehicle dynamics or environmental factors.  For example, decaying battery life, mechanical wear, or mud/water can all affect the vehicle's intrinsic dynamics.  Additionally, changes in surfaces in the environment can strongly affect vehicle motion.  To adapt to these changes we employ learning-based methods \rev{using Gaussian processes} which adapt critical vehicle parameters and dynamics models based on the past history of performance \cite{fan2020bayesian} \cite{fan2020deep}. \rev{By accounting for both the epistemic and aleatoric uncertainties of these statistical models,} these methods ensure safety and robustness even in light of changing dynamics models.

\begin{figure}[h!]
    \centering
    \includegraphics[width=0.5\linewidth]{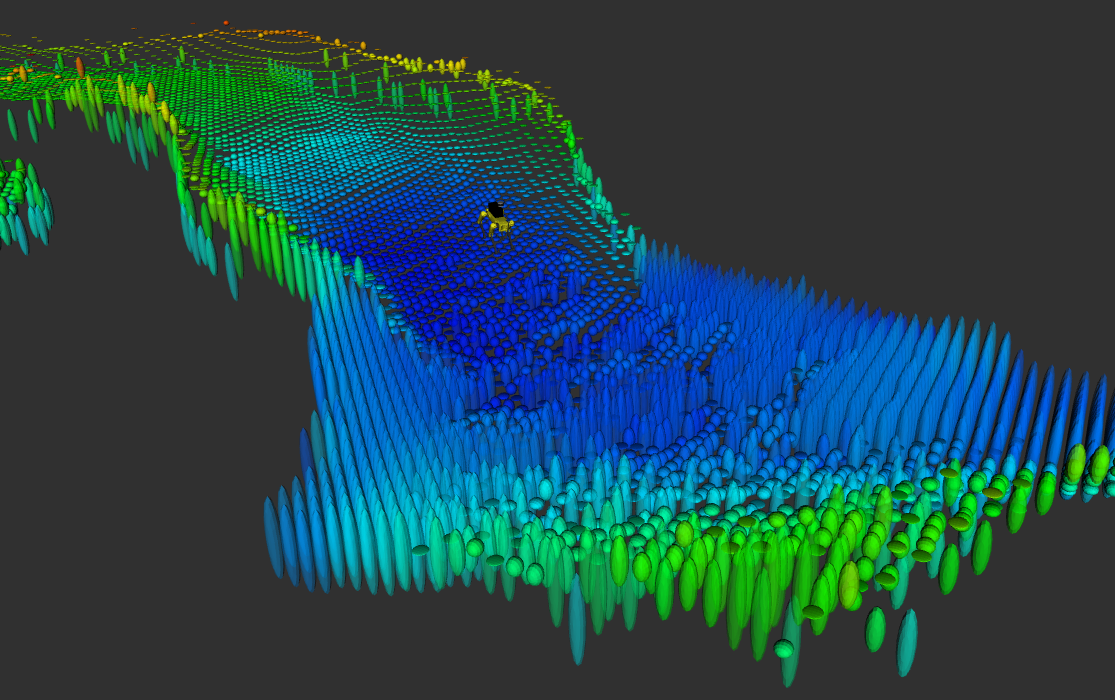}
    \caption{Belief-cloud:  We aggregate uncertainties in sensor noise and localization into a belief point cloud.  (Ellipses describe uncertainties associated with points at the center of each ellipse.)  This can be used for accurate and efficient uncertainty aware traversability analysis and planning.}
    \label{fig:beliefcloud}
\end{figure}

\pr{Ongoing Work}
Our ongoing work lies in increasing the ability of NeBula to account for and handle uncertainties in both perception and planning.  One major thrust of ongoing work involves perception-aware planning.  By incorporating sensor models into our traversability maps, we can move towards perception-aware behaviors which maximize sensor coverage, optimally reduce uncertainty, and automatically generate active learning behaviors.  This thrust is particularly important in perceptually degraded and complex environments filled with occlusion and traversability hazards.  A second direction of ongoing work lies in incorporating localization uncertainty into our traversability risk mapping in a more theoretically satisfying way, while remaining computationally tractable.  We call this approach "mapping using belief-clouds", and the idea is to propagate the uncertainty of the robot pose while taking sensor measurements into an aggregated point cloud.  The result is a belief-cloud, i.e a pointcloud which encodes uncertainty information directly and efficiently.  Our ongoing work involves extracting traversability metrics and risks from this uncertainty-aware mapping data (see \autoref{fig:beliefcloud}).

\section{Uncertainty-aware Global Planning} \label{sec:global_planning}
Autonomous global planning for environment exploration and coverage is a core part of the NeBula architecture (see \autoref{fig:nebula_architecture}).
NeBula formulates the autonomous exploration in unknown environments under motion and sensing uncertainty by a Partially Observable Markov Decision Process (POMDP), one of the most general models for sequential decision making.
This formulation allows NeBula to jointly consider sequential outcomes of perception and control at the planning phase in order to achieve higher levels of resiliency during the mission operation.
In this section, we discuss our POMDP-based global planning, referred to as PLGRIM (Probabilistic Local and Global Reasoning on Information roadMaps). For more details, please see \cite{kim_plgrim_icaps_2021,AutoSpot,youtubespotpaper}.

\subsection{Problem Formulation}
\ph{POMDP Formulation}
A POMDP is described as a tuple $\langle \mathbb{S}, \mathbb{A}, \mathbb{Z}, T, O, R \rangle$, where $\mathbb{S}$ is the set of states of the robot and world, $\mathbb{A}$ and $\mathbb{Z}$ are the set of robot actions and observations, respectively \cite{KLC98,Pineau03}.
At every time step, the agent performs an action $a \in \mathbb{A}$ and receives an observation $z \in \mathbb{Z}$ resulting from the robot's perceptual interaction with the environment.
The motion model $T(s, a, s') = P(s'\,|\,s, a)$ defines the probability of being at state $s'$ after taking an action $a$ at state $s$.
The observation model $O(s, a, z) = P(z\,|\,s, a)$ is the probability of receiving observation $z$ after taking action $a$ at state $s$.
The reward function $R(s, a)$ returns the expected utility for executing action $a$ at state $s$.
In addition, a belief state $b_t \in \mathbb{B}$ at time $t$ is introduced to denote a posterior distribution over states conditioned on the initial belief $b_0$ and past action-observation sequence, i.e., $b_{t} = P(s \,|\, b_0, a_{0:t-1}, z_{1:t})$.
The optimal policy $\pi^* \! : \mathbb{B} \to \mathbb{A}$ of a POMDP for a finite receding horizon is defined as follows:
\begin{align}
  \pi_{t:t+T}^*(b) &= \argmax_{\pi \in \Pi_{t:t+T}} \, \mathbb{E} \sum_{t'=t}^{t+T} \gamma^{t'-t} r(b_{t'}, \pi(b_{t'})),
  \label{eq:receding_objective_function}
\end{align}
where $\gamma \in (0, 1]$ is a discount factor for the future rewards, and $r(b,a)=\int_s R(s,a)b(s)\mathrm{d}s$ denotes a belief reward which is the expected reward of taking action $a$ at belief $b$.
$T$ is a finite planning horizon for a planning episode at time $t$.
Given the policy for last planning episode, only a part of the optimal policy, $\pi^*_{t:t+\Delta t}$ for $\Delta t \in (0, T]$, will be executed at run time. A new planning episode will start at time $t+\Delta t$ given the updated belief $b_{t+\Delta t}$. %
The computational complexity of a POMDP grows exponentially with the planning horizon \cite{Pineau03}, and we tackle this challenge with hierarchical belief space representation and planning as to be detailed in \autoref{sec:plgrim}.

\ph{Application to Simultaneous Mapping and Planning (SMAP)}
To formalize our SMAP problem as a POMDP, we define the state $s = (q, W)$ as a pair of robot $q$ and world state $W$.
We further decompose the world state as $W = (W_{occ}, W_{cov})$ where $W_{occ}$ and $W_{cov}$ describe the occupancy and the coverage states of the world, respectively associated with their uncertainties (e.g., \cite{CRM}).
A reward function for coverage can be defined as a function of information gain $I$ and action cost $C$ as follows:
\begin{align}
  R(s, a) = \mathrm{fn}(I(W_{cov}, z),\; C(W_{occ}, q, a)),
  \label{eq:coverage_reward}
\end{align}
where $I(W_{cov}, z) = H(W_{cov}) - H(W_{cov} \,|\, z)$ is quantified as reduction of the entropy $H$ in $W_{cov}$ after observation $z$,  %
and $C(W_{occ}, q, a)$ is evaluated from actuation efforts and risks to take action $a$ at robot state $q$ on $W_{occ}$. \rev{Minimizing this cost function in Eq.~(\ref{eq:receding_objective_function})) simultaneously solves for the mapping and planning (SMAP), maximizing the coverage for \textit{artifact detection} and minimizing the action risk (e.g., collision chance).}
This reward function can be generalized to Simultaneous Localization and Planning (SLAP) problems (e.g., \cite{agha2018slap} or \cite{BVL}) by incorporating information gain based on localization entropy reduction events, such as a loop closure or landmark detection.

\begin{figure}[t!]
  \centering
  \includegraphics[width=0.6\columnwidth]{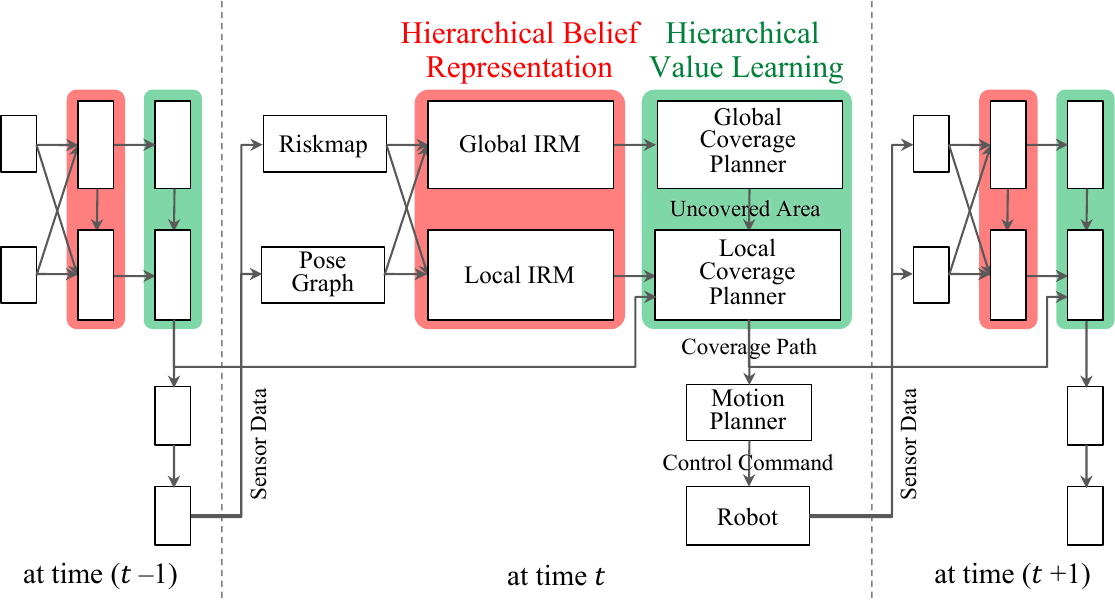}
  \caption{\rev{Illustration of PLGRIM framework for large-scale exploration in unknown environments.
 Over the receding-horizon planning episodes, PLGRIM (\textit{i}) maintains hierarchical beliefs about the traversal risks and coverage states, and
  (\textit{ii}) performs hierarchical value learning to construct an exploration policy.}} 
  \label{fig:gp_framework} 
\end{figure}

\subsection{Hierarchical Coverage Planning on Information Roadmaps} \label{sec:plgrim}
In this subsection, we introduce NeBula's solution for uncertainty-aware global coverage planning, PLGRIM (Probabilistic Local and Global Reasoning on Information roadMaps) \cite{kim_plgrim_icaps_2021}.
PLGRIM proposes a hierarchical belief representation and belief space planning structure to scale up to spatially large problems while pursuing locally near-optimal performance (see \autoref{fig:gp_framework}).
At each hierarchical level, it maintains a belief about the world and robot states in a compact form, called Information RoadMap (IRM), and solves for a POMDP policy to generate a coverage plan over a non-myopic temporal horizon, in a receding horizon fashion.

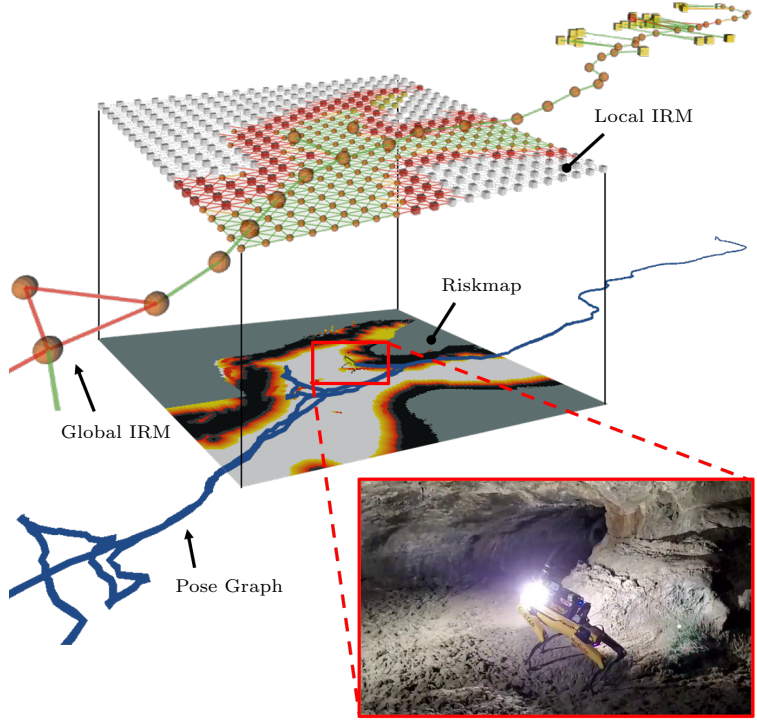
\begin{figure}[t!]
\centering
    \begin{tikzpicture}
    \node[anchor=south west,inner sep=0] (image) at (0,0) {\includegraphics[width=0.6\columnwidth]{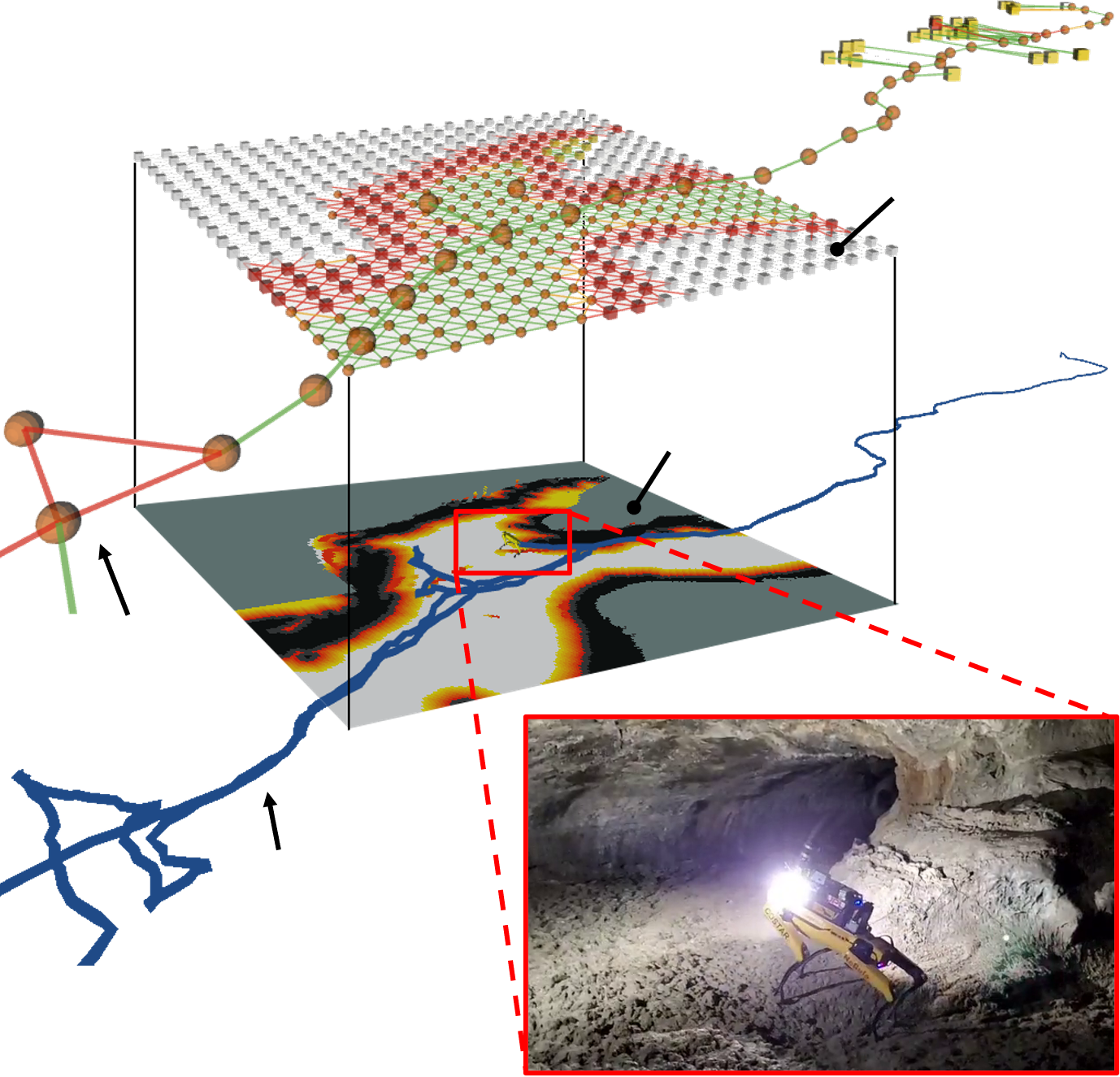}};
	    \begin{scope}[x={(image.south east)},y={(image.north west)}]

	    	\node [font=\scriptsize,above left,align=right,black] at (0.93,0.82) {Local IRM}; %
	    	\node [font=\scriptsize,above left,align=right,black] at (0.7,0.58) {Riskmap}; %
	    	\node [font=\scriptsize,above left,align=right,black] at (0.38,0.16) {Pose Graph};
	    	\node [font=\scriptsize,above left,align=right,black] at (0.23,0.38) {Global IRM};

	    \end{scope}
	\end{tikzpicture}	
  \caption{Hierarchical IRM generated during autonomous exploration of Valentine's cave at Lava Beds National Monument, Tulelake, CA, using a quadruped robot} 
  \label{fig:IRMs} 
\end{figure}

\ph{Hierarchical POMDP Formulation}
First, we formulate the receding-horizon SMAP in Eq.~(\ref{eq:receding_objective_function}) into a hierarchical POMDP problem \cite{kaelbling2011planning,kim2019pomhdp}.
Let us decompose a belief state $b$ into local and global belief states, $b^\ell = P(q, W^\ell)$ and $b^g = P(q, W^g)$, respectively.
$W^\ell$ is a local, rolling-window world representation with high-fidelity information, while $W^g$ is a global, unbounded world representation with approximate information (see \autoref{fig:IRMs}).
With $\pi^\ell$ and $\pi^g$ denoting the local and global policies, respectively, we approximate Eq.~(\ref{eq:receding_objective_function}) as cascaded hierarchical optimization problems as follows:
\begin{align}
  &\pi_{t:t+T}(b)
   \approx \argmax_{\pi^\ell \in \Pi^\ell_{t:t+T}} \, \mathbb{E} \sum_{t'=t}^{t+T} \gamma^{t'-t} r^\ell(b^\ell_{t'}, \pi^\ell(b^\ell_{t'}; \pi_{t:t+T}^g(b^g_t))),
  \label{eq:llp_optimization}
  \\
  &\,\,\,\text{where }
  \pi_{t:t+T}^g(b^g) = \argmax_{\pi^g \in \Pi^g_{t:t+T}} \, \mathbb{E} \sum_{t'=t}^{t+T} \gamma^{t'-t} r^g(b^g_{t'}, \pi^g(b^g_{t'})).
  \label{eq:glp_optimization}
\end{align}
$r^\ell(b^\ell, \pi^\ell(b^\ell))$ and $r^g(b^g, \pi^g(b^g))$ are approximate belief reward functions for the local and global belief spaces, respectively.
Note that the co-domain of the global policy $\pi^g(b^g)$ is a parameter space $\Theta^\ell$ of the local policy $\pi^\ell(b^\ell; \theta^\ell)$, $\theta^\ell \!\! \in \! \Theta^\ell\!$.\,

\ph{Hierarchical Belief Representation}
For a compact and versatile representation of the world, we rely on a graph structure, $G = (N, E)$ with nodes $N$ and edges $E$, as the data structure to represent the belief about the world state.
We refer to this representation as an IRM \cite{Ali14-IJRR}.
We construct and maintain IRMs at two hierarchical levels, namely, Local IRM and Global IRM (see \autoref{fig:IRMs}).
The Local IRM is a dense high-resolution graph that contains high-fidelity information about the occupancy, coverage, and traversal risks, but locally around the robot.
In contrast, the Global IRM sparsely captures the free-space connectivity. It encodes uncovered area by so-called frontier nodes, which allow for effective representation of large environments, spanning up to several kilometers.
In addition to the map uncertainty, IRM can be generalized to incorporate the robot localization uncertainty (e.g., \cite{BVL} or \cite{agha2018slap}) in the planning framework when traversing narrow passages and challenging environments where robot location uncertainty can hinder robot's ability to navigate the environment.


\begin{figure}[t!]
\centering
    \begin{tikzpicture}
    \node[anchor=south west,inner sep=0] (image) at (0,0) {\includegraphics[width=0.63\columnwidth]{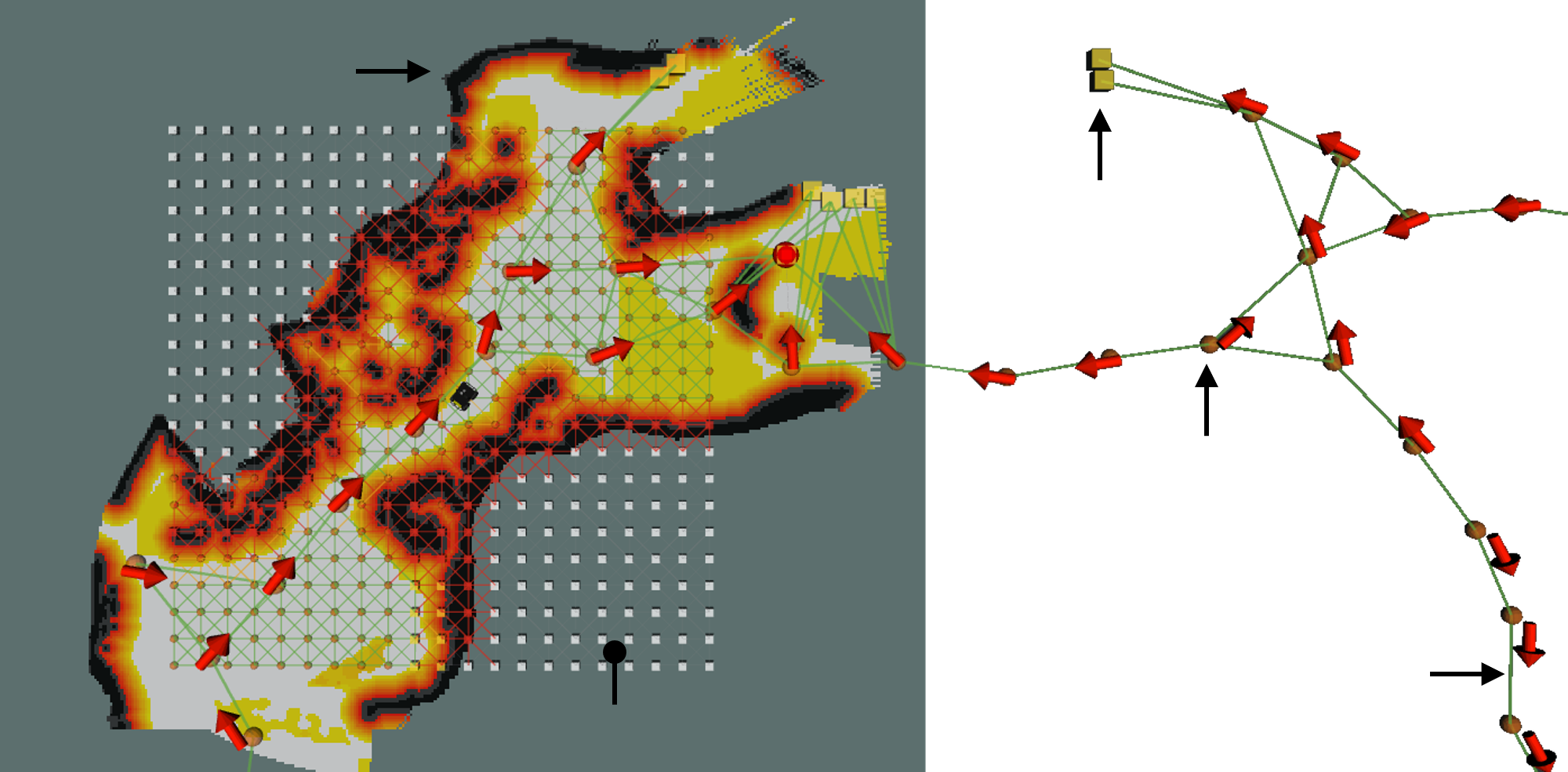}};
	    \begin{scope}[x={(image.south east)},y={(image.north west)}]
	    	
	    	\node [font=\scriptsize,above left,align=right,black] at (0.8,0.69) {Frontier Node}; 
	    	\node [font=\scriptsize,above left,align=right,black] at (0.86,0.36) {Breadcrumb}; 
	    	\node [font=\scriptsize,above left,align=right,black] at (0.82,0.2925) {Node}; 
	    	\node [font=\scriptsize,above left,align=right,black] at (0.235,0.86) {Riskmap}; 
	    	\node [font=\scriptsize,above left,align=right,black] at (0.475,0.018) {Local IRM}; 
	    	\node [font=\scriptsize,above left,align=right,black] at (0.92,0.091) {Global IRM}; 

	    \end{scope}
	\end{tikzpicture}	
  \caption{QMDP policy (red arrows displayed above \textit{breadcrumb} nodes) for Global Coverage Planning (GCP). A red sphere indicates the QMDP frontier goal.}
  \label{fig:graph-level-planner}
\end{figure}
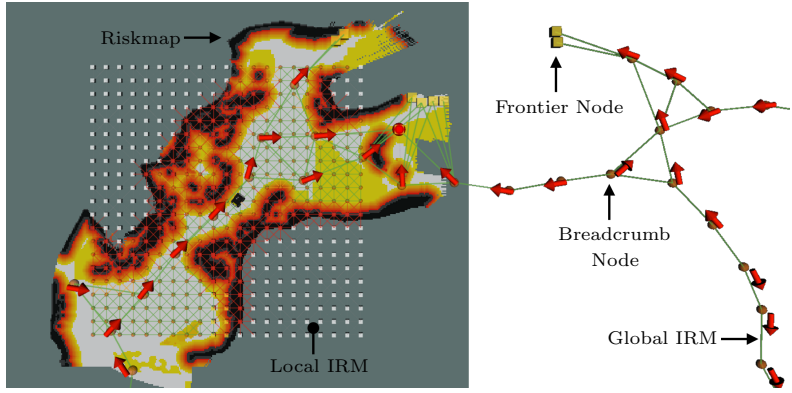

\begin{figure}[t!]
  \centering
  \includegraphics[width=1.0\columnwidth]{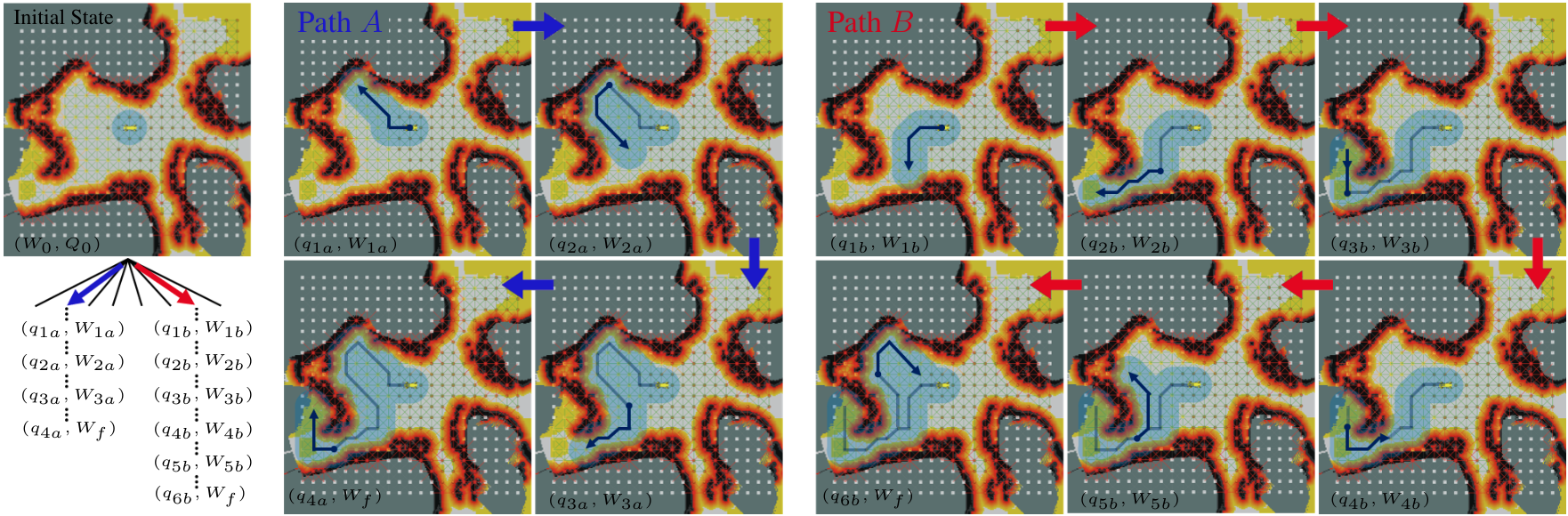}
  \caption{Illustrative example of coverage path planning on the Local IRM with Monte-Carlo Tree Search. The field-of-view of the robot's coverage sensor is represented by a blue circle. 
  Macro actions (6 steps on Local IRM in this example) associated with the two tree branches, paths \textit{A} and \textit{B}, are shown. Note that the final world coverage states in both branches are identical. Path \textit{A} is evaluated to be more rewarding than \textit{B} since fewer actions were required to cover the same area.}
  \label{fig:lattice-level-planner}
\end{figure}

\ph{Hierarchical Coverage Planning--GCP}
Given Local and Global IRMs as the hierarchical belief representation of $W^\ell$ and $W^g$, respectively, we solve the cascaded hierarchical POMDP problems. %
At first, Global Coverage Planner (GCP) solves for the global policy in Eq.~(\ref{eq:glp_optimization}), providing \textit{global guidance} to Local Coverage Planner (LCP) of Eq.~(\ref{eq:llp_optimization}).
The global guidance enhances the coverage performance and global completeness. It is especially helpful when LCP has fully covered the local area and needs global guidance to move to another area.
In order for GCP to scale up to very large problems, we adopt \rev{the QMDP} approach \cite{littman1995learning}.
The main idea is to assume the state becomes fully observable after one step of action \rev{under uncertainty}, so that the value function for further actions can be evaluated efficiently in an MDP (Markov Decision Process) setting.
This assumption is acceptable for GCP since its main role is to guide the robot to an uncovered area and let LCP lead the local coverage behavior. In other words, GCP's policy search can terminate at frontier nodes of the Global IRM, and thus, we can assume no changes in the coverage state during GCP's planning episode and adopt QMDP for efficient large-scale planning.
\rev{The complexity of GCP is $O(N_{iter}^g \, |N^g| \, |N_{nn}^g|^2)$, where $N_{iter}^g$ is the maximum number of iterations for Bellman update, $|N^g|$ is the number of nodes on Global IRM, and $|N_{nn}^g|$ is the number of nearest neighbor nodes for a node connected by an edge on Global IRM which is bounded to a small finite number.
Thus, the complexity of GCP grows only linearly with $|N^g|$.}

\ph{Hierarchical Coverage Planning--LCP}
In the hierarchical optimization framework, LCP solves Eq.~(\ref{eq:llp_optimization}), given a parameter input from GCP.
LCP constructs a high-fidelity policy by considering the information gathering (with visibility check given obstacles), traversal risk (based on proximity to obstacles, terrain roughness, and slopes), and robot's mobility constraints (such as acceleration limits and non-holonomic constraints of wheeled robots).
\rev{LCP has two phases: i) reach the target area based on GCP's guidance, and ii) construct a local coverage path after reaching the target area.
In the first case, when the target frontier is outside the Local IRM range, LCP instantiates high-fidelity motion commands to reach the target frontier.
In the second case, when the target frontier is within the Local IRM range, then LCP performs the information-gathering coverage optimization, as described in Eq.~(\ref{eq:llp_optimization}).} 
In order to solve \rev{the coverage optimization problem} we employ POMCP (Partially Observable Monte Carlo Planning) algorithm \cite{silver2010monte,kim2019bi}.
POMCP is a widely-adopted POMDP solver that leverages the Monte Carlo sampling technique to alleviate both the curses of dimensionality and history.
Given a generative model (or a black box simulator) for discrete action and observation spaces, POMCP can learn the near-optimal value function of the reachable belief space with adequate exploration-exploitation trade-off.
\rev{We limit the complexity of LCP process to $O(N_{iter}^\ell N_{depth}^\ell)$ of calling the generative model, where $N_{iter}^\ell$ is the number of iterations for episodic forward simulation and $N_{depth}^\ell$ is the depth of planning horizon for each iteration.
Since it is a local rolling-window planner, there is no increased complexity with the total size of the environment.}

\subsection{Experimental Evaluation}\label{sec:exp_results}
In order to evaluate our proposed framework, we perform high-fidelity simulation studies with a four-wheeled vehicle (Husky robot) and real-world experiments with a quadruped (Boston Dynamics Spot robot). Both robots are equipped with custom sensing and computing systems, enabling high levels of autonomy and communication capabilities~\cite{Otsu2020}. The entire autonomy stack runs in real-time on an Intel Core i7 processor with 32 GB of RAM. The stack relies on a multi-sensor fusion framework. The core of this framework is 3D point cloud data provided by LiDAR range sensors mounted on the robots~\cite{LAMP}. We refer to our autonomy stack-integrated Spot as Au-Spot~\cite{AutoSpot}.

\ph{Baseline Algorithms}
We compare our PLGRIM framework against a local coverage planner \rev{baseline} (next-best-view \rev{method}) and a global coverage planner \rev{baseline} (frontier-based \rev{method}). 
\vspace{-16pt}
\begin{enumerate}[label={\arabic*)}]
  \itemsep2em 
  \setlength{\itemsep}{2pt}
  \setlength{\parskip}{0pt}
  \item \textbf{Next-Best-View (NBV)}: 
  NBV is a widely-adopted local coverage planner that returns a path to the best next view point to move to.
  It uses an information gain-based reward function as ours but limits the policy search space to a set of shortest paths to sampled view points around the robot.
\rev{While NBV is able to leverage local high-fidelity information, it suffers from spatially limited world belief and sparse policy space.}
  \item \textbf{Hierarchical Frontier-based Exploration (HFE)}:
  Frontier-based exploration is a prevalent global coverage planning approach that interleaves moving to a frontier node and creating new frontiers until there are no more frontiers left (e.g., \cite{umari2017autonomous}).
  While it optimizes for the global completeness of environment exploration but often suffers from the local suboptimality due to its large scale of the policy space and myopic one-step look-ahead decision making.
	\rev{The performance of frontier-based methods can be enhanced by modulating the spatial scope of frontier selection, but it still suffers from downsampling artifacts and a sparse policy space composed of large action steps.}
\end{enumerate}
\vspace{-16pt}

\rev{
\subsubsection{Simulation Evaluation}
We demonstrate PLGRIM's performance, as well as that of the baseline algorithms, in a simulated subway, maze, and cave environment. \autoref{fig:maps_of_cave} visualizes these environments. In our comparisons, in order to achieve reasonable performance with the baseline methods in complex simulated environments, we allow baseline methods to leverage our Local and Global IRM structures as the underlying search space.

\ph{Simulated Subway Station}
The subway station consists of large interconnected, polygonal rooms with smooth floors, devoid of obstacles. There are three varying sized subway environments, whose scales are denoted by 1x, 2x, and 3x. 
\autoref{fig:all_together_plot}(a)-(c) shows the scalable performance of PLGRIM against the baselines. 
In a relatively small environment without complex features (Subway 1x), NBV performance is competitive as it evaluates high-resolution paths based on information gain.
However, as the environment scale grows, its myopic planning easily gets \textit{stuck} and the robot's coverage rate drops significantly. 
HFE shows inconsistent performance in the subway environments. The accumulation of locally suboptimal decisions, due to its sparse environment representation, leads to the construction of a globally inefficient IRM structure. As a result, the robot must perform time-consuming detours in order to \textit{pick up} leftover frontiers.  

\begin{figure}[t!]
\centering
	\subfloat[][\rev{Simulated Subway 1x}]{
        \begin{tikzpicture}
    	    \node[anchor=south west,inner sep=0] (image) at (0,0) {\includegraphics[width=0.28\columnwidth]{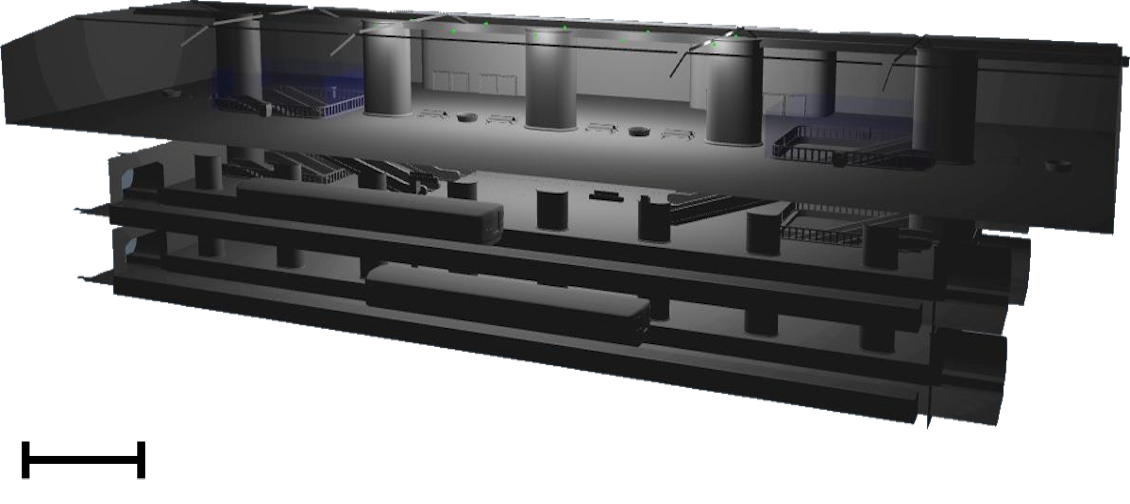}};
    	    \begin{scope}[x={(image.south east)},y={(image.north west)}]
    	    	\node [font=\scriptsize,above left,align=right,black] at (0.17,0.1) {15 m};
    	    \end{scope}
    	\end{tikzpicture}	
	}
	\subfloat[][Simulated Maze]{
        \begin{tikzpicture}
    	    \node[anchor=south west,inner sep=0] (image) at (0,0) {\includegraphics[width=0.25\columnwidth]{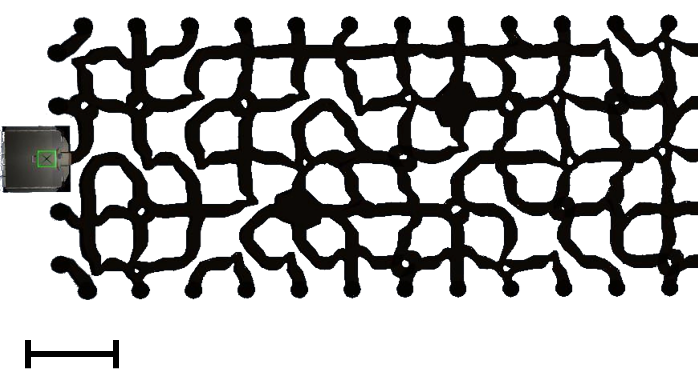}};
    	    \begin{scope}[x={(image.south east)},y={(image.north west)}]
    	    	\node [font=\scriptsize,above left,align=right,black] at (0.2,0.09) {10 m};
    	    \end{scope}
    	\end{tikzpicture}	
	}
	\subfloat[][Simulated Cave]{
        \begin{tikzpicture}
    	    \node[anchor=south west,inner sep=0] (image) at (0,0) {\includegraphics[width=0.38\columnwidth]{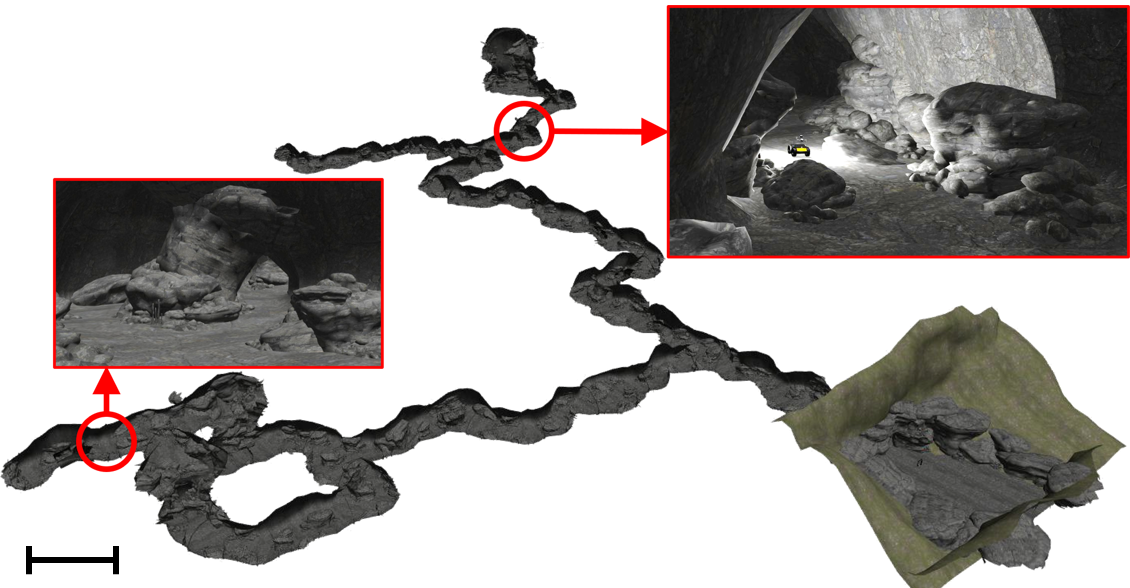}};
    	    \begin{scope}[x={(image.south east)},y={(image.north west)}]
    	    	\node [font=\scriptsize,above left,align=right,black] at (0.13,0.055) {10 m};
    	    \end{scope}
    	\end{tikzpicture}	
	}
\caption{\rev{Simulated environments for performance validation: (a) subway station,} (b) maze (top-down view), and (c) cave.}
\label{fig:maps_of_cave}
\end{figure}
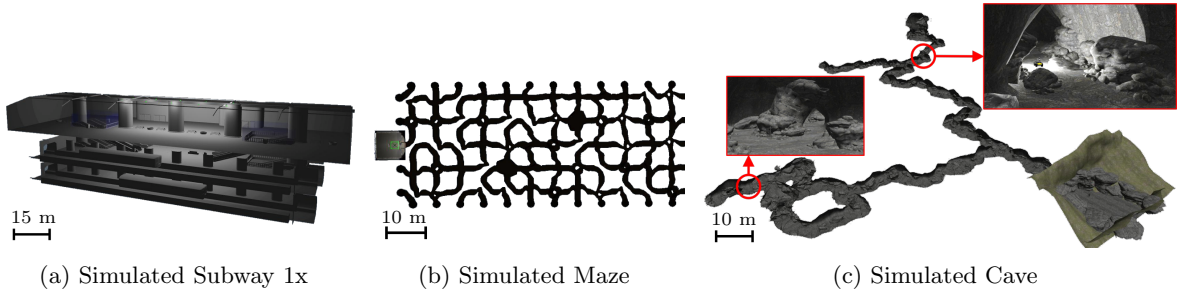

\begin{figure}[t!]
    \centering
		\subfloat[][Subway 1x]{\includegraphics[height=0.2735525\columnwidth]{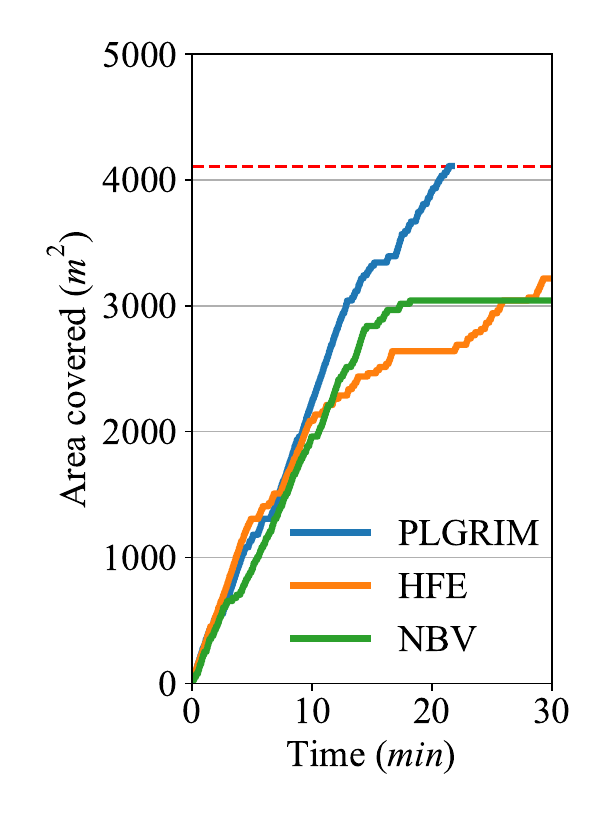}}
		\subfloat[][Subway 2x]{\includegraphics[height=0.27082125\columnwidth]{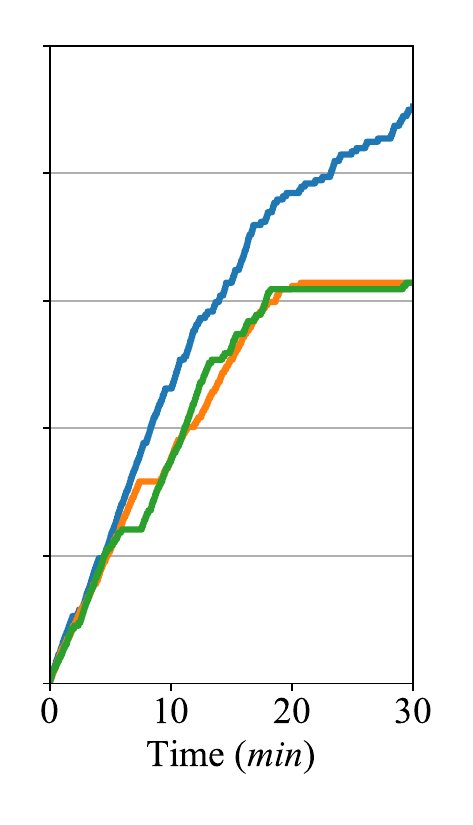}}	
		\subfloat[][Subway 3x]{\includegraphics[height=0.27082125\columnwidth]{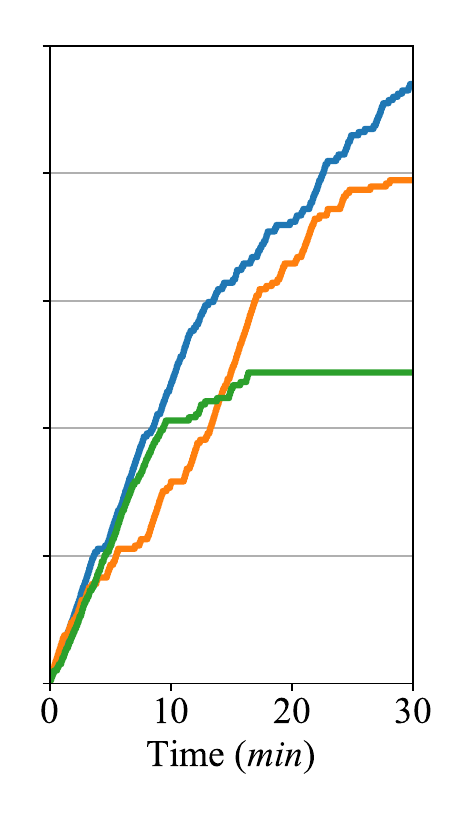}}	
        \subfloat[][Maze]{\includegraphics[height=0.27082125\columnwidth]{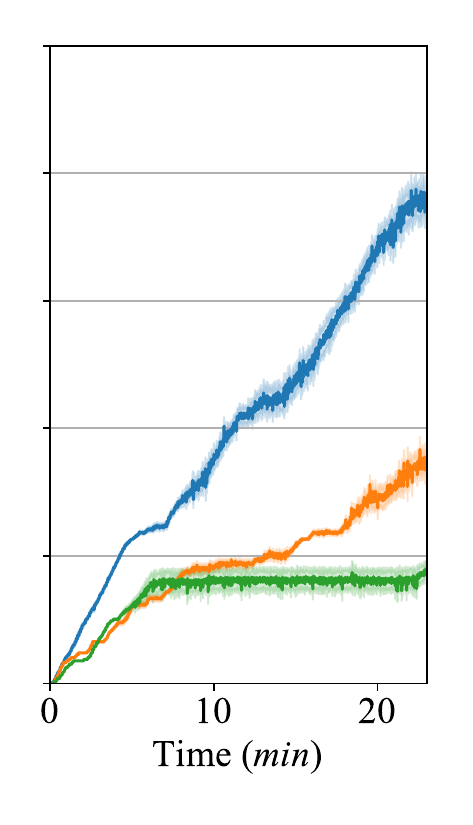}}
        \subfloat[][Cave]{\includegraphics[height=0.27082125\columnwidth]{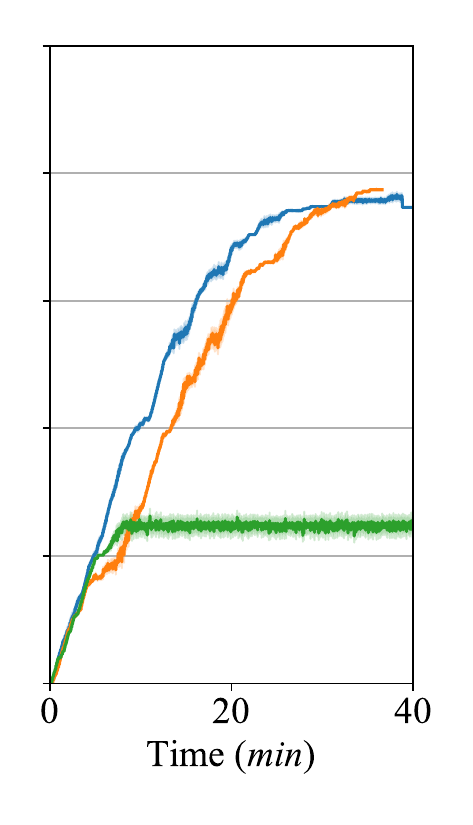}}
		\subfloat[][Lava Tube]{\includegraphics[height=0.2735525\columnwidth]{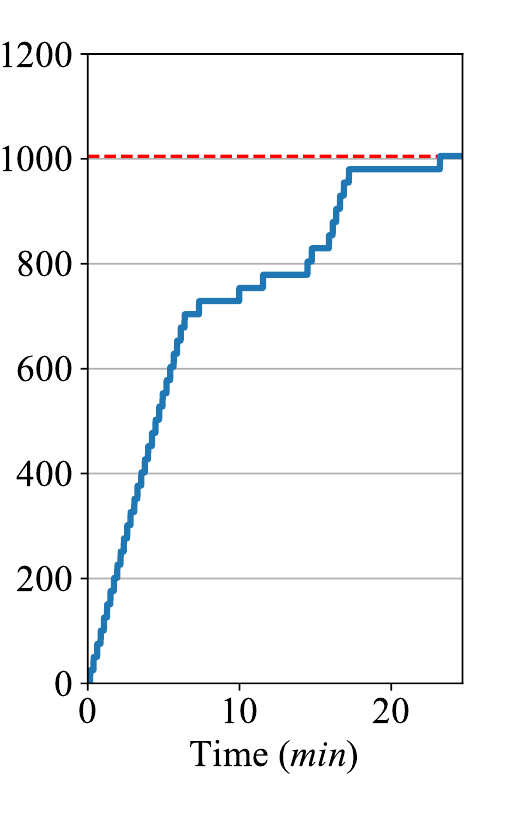}}
		\caption{Exploration by PLGRIM and baseline methods in simulated subways of increasing size (a)-(c), simulated maze and cave (d)-(e), and real-world lava tube (f). For (d) and (e), the covered area is the average of two runs. Red dashed lines indicate 100\% coverage of the environments, wherever applicable. 
		}
    \label{fig:all_together_plot}
\end{figure}

\ph{Simulated Maze and Cave}
The maze and cave are both unstructured environments with complex terrain (rocks, steep slopes, etc.) and topology (narrow passages, sharp bends, dead-ends, open-spaces, etc.).
The coverage rates for each algorithm are displayed in \autoref{fig:all_together_plot}(d)-(e). 
PLGRIM outperforms the baseline methods in these environments. By constructing long-horizon coverage paths over a high-resolution world belief representation,
PLGRIM enables the robot to safely explore through hazardous terrain. 
Simultaneously, it maintains an understanding of the global world, which is leveraged when deciding where to explore next after exhausting all local information.
In the cave, NBV's reliance on a deterministic path, without consideration of probabilistic risk, causes the robot to drive into a pile of rocks and become inoperable. NBV exhibits similarly poor performance in the maze. However, in this case, NBV's myopic planning is particularly ineffectual when faced with navigating a topologically-complex space, and the robot ultimately gets \textit{stuck}.   
As was the case in the subway, HFE suffers in the topologically-complex maze due to the accumulation of suboptimal local decisions. In particular, frontiers are sometimes not detected in the sharp bends of the maze, leaving the robot with an empty local policy space. As a result, the robot cannot progress and spends considerable time backtracking along the IRM to distant frontiers.  
}

\begin{figure}[t!]
  \centering
  \includegraphics[width=1.0\columnwidth]{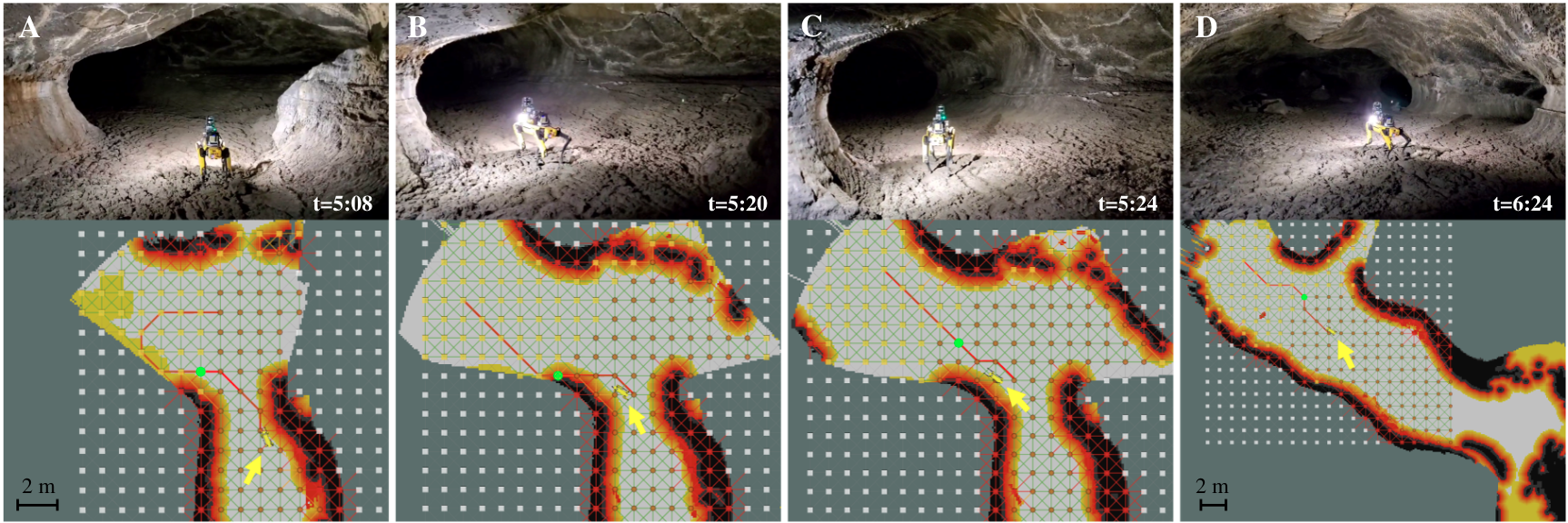}
  \caption{The Local IRM (yellow, brown, and white nodes represent uncovered, covered and unknown areas, respectively) is shown overlaid on the Riskmap. A yellow arrow indicates the robot's location. LCP plans a path (red) that fully covers the local area (snapshot A). When $p(W^\ell)$ updates, the path is adjusted to extend towards the large uncovered swath while maintaining smoothness with the previous path. Another $p(W^\ell)$ update reveals that the path has entered a hazardous area---wall of lava tube (snapshot B). As a demonstration of LCP's resiliency, the path shifts away from the hazardous area, and the robot is re-directed towards the center of the tube (snapshot C). One minute later, the robot encounters a fork in the cave. The LCP path curves slightly toward fork apex (for maximal information gain) before entering the wider, less-risky channel (snapshot D).}
  \label{fig:mlp_hardware_tests}
\end{figure}

\begin{figure}[t!]
  \centering
  \includegraphics[width=1.0\columnwidth]{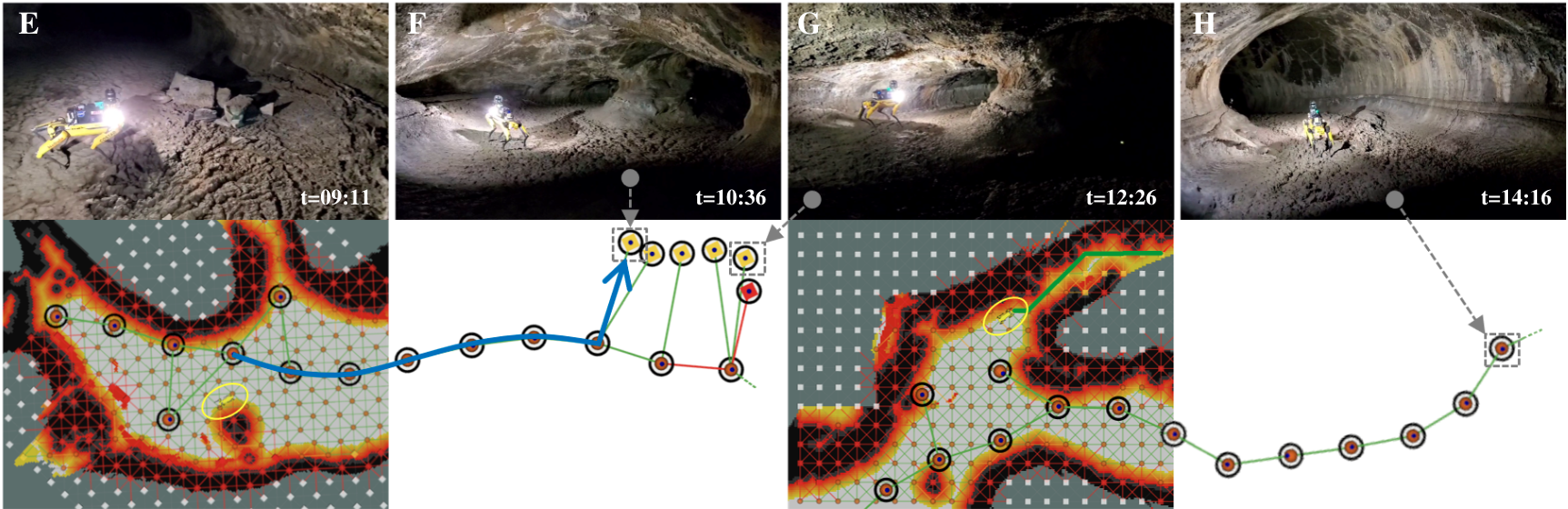}
  \caption{Portions of the Global IRM constructed in the lava tube are visualized--yellow nodes represent frontiers, brown nodes represent breadcrumbs. Gray arrows associate a frontier with a snapshot of the robot exploring that frontier. GCP plans a path (blue) along the Global IRM to a target frontier after the local area is fully covered (snapshot E). The robot explores the area around the frontier (snapshot F), and then explores a neighboring frontier at the opening of a narrow channel to its right. LCP plans a path (green) into the channel (snapshot G). Later, after all local areas have been explored, the robot is guided back towards the mouth of cave along the breadcrumb nodes (snapshot H).}
  \label{fig:glp_hardware_tests}
\end{figure}

\subsubsection{\rev{Real-World Evaluation}}
We extensively validated PLGRIM on physical robots in real-world environments. In particular, PLGRIM was run on a quadruped robot Valentine lava tube, located in Lava Beds National Monument, Tulelake, CA. The cave consists of a main tube, which branches into smaller, auxiliary tubes. The floor is characterized by ropy masses of cooled lava. Large boulders, from ceiling breakdown, are scattered throughout the tube.
Fig.~\ref{fig:mlp_hardware_tests} and \ref{fig:glp_hardware_tests} discuss how PLGRIM is able to overcome the challenges posed by large-scale environments with complex terrain and efficiently guide the robot's exploration. Fig.~\ref{fig:all_together_plot}(f) shows the area covered over time.
\section{Multi-Robot Networking} \label{sec:multirobot_networking}
Multi-robot systems offer advanced capabilities to enable complex and time-constrained missions in large-scale complex environments. Resilient wireless mesh networking solutions are a prerequisite for reliable and efficient multi-robot missions. NeBula is inherently a “networked” solution (\autoref{fig:nebula_architecture}). While it can be implemented on a single autonomous robot, it also allows for faster and more efficient missions with multiple potentially heterogeneous robots (see \autoref{fig:nebula_robots}). NeBula's goal in the SubT challenge is to map an unknown subterranean environment, locate artifacts, and communicate that information to the base station via a wireless mesh network for submission to the DARPA Command Post.  Inter-robot wireless communication in subterranean environments is particularly challenging and uncertain in the reliability, capacity, and availability of communication links because of: i) limited line-of-sight opportunities, ii) the complicated nature of the interaction of radio signals with the environment (e.g. reflecting, scattering, multipath), and iii) the unknown nature of the environment.  In this section, we will go over NeBula’s CHORD (Collaborative High-bandwidth Operations with Radio Droppables) communication system for comm-degraded subterranean environments. The objective of CHORD is to maintain high-bandwidth links to multiple robots for efficient commanding, autonomous operation, and data gathering in complex unknown environments.
 For more details on the development of NeBula’s networking solutions, see \cite{Otsu2020,CHORD}.

\begin{figure*}[t]
  \centering
  \includegraphics[width=0.9\textwidth]{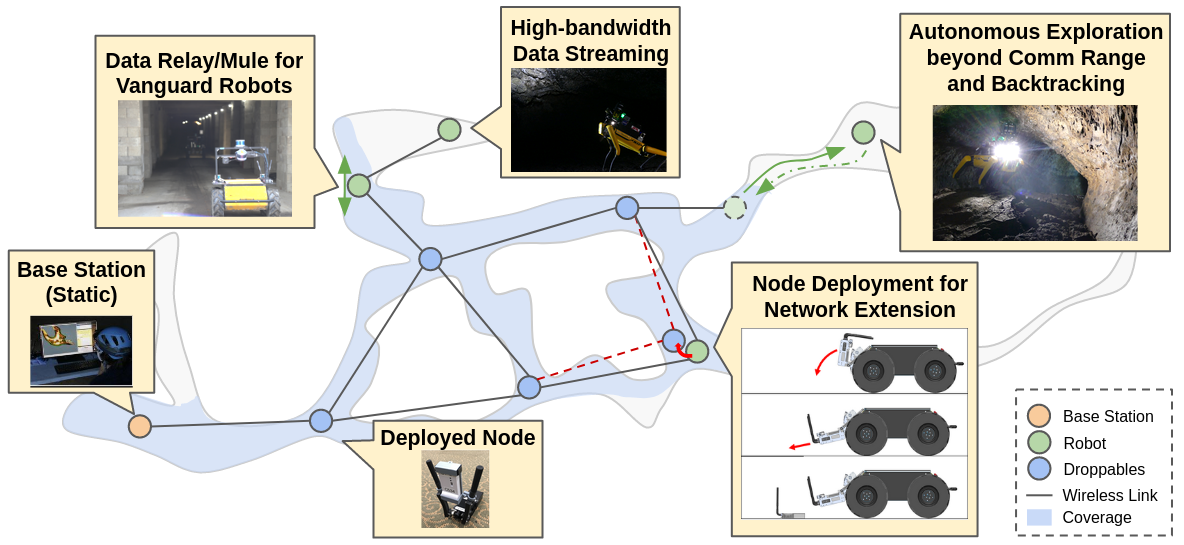}
  \caption{This figure shows how the mesh network is extended into the subterranean environment by deployed communication nodes which form a backbone network. Robots inside the coverage area of the backbone network can extend the network to robots in their communication range. Robots without a route through the network to the base station must be fully autonomous and return to an area with communication coverage to return data. Alternatively, (not pictured) multiple robots without a wireless route to the base station may share data and have one act as a data-mule to carry the data back to communication range.
  }
  \label{fig:chord_deployment}
\end{figure*}

\subsection{CHORD System Design}
\ph{Architecture and ConOps}
NeBula's operations in the SubT Challenge consists of three general types of agents:  1)  static  agents  (e.g.\ base  station),  2)  mobile  agents  (e.g.\ robots),  and  3)  deployable static agents  (e.g.\ communication relay nodes).
Each agent communicates by means of a wireless mesh network using commercial off-the-shelf radios. 
We use a hybrid of ROS 1 and ROS  2  for the communication middleware \cite{CHORD,fadhil-agu}. We use ROS 1 for intra-robot communications, and ROS 2 for inter-robot communications.
The mesh network can be extended, as shown in \autoref{fig:chord_deployment}, into the subterranean environment by deploying communication nodes (\autoref{sec:hardware_static_asset}) from robots to build a backbone wireless mesh network. The decision of where to drop communication nodes is based on the 3D map, data route, signal to noise ratio, and the estimated available bandwidth between each radio \cite{vaquero2020traversability}.  The exact coverage area of each node is dependent on many factors including surface materials, roughness, and environment shape.  Dropping communication nodes in range of another node with a route to the base station reduces the uncertainty of getting data back from robots near that node.
Robots may also be used to extend the mesh network when they are in communication range of another asset (robot or communication node) with a route to the base station. 
When a robot needs to send data to the base station, but has no communication route to the base station, it can either return to the communication range of an asset that does have a route to the base station (usually be backtracking to a node on the backbone network) or communicate that data to another robot (referred to as a data mule) which is returning.

\ph{Intra-robot Communication} Each robot consists of a combination of computers and sensors (see~\autoref{sec:hardware}) connected by Gigabit Ethernet. Where possible (current generation Husky hardware) this network is separate from the radio network and connects only through a single computer which is connected to a radio. That computer runs the ROS 1 core, the ROS 1-2 bridge, and is responsible for handling inter-agent communication. This further isolates intra-robot communication from inter-agent communication and prevents inadvertent radio traffic. Intra-robot communications are monitored using ROS 1 topic statistics.

\ph{Inter-agent Communication} %
Even with relatively high power/bandwidth radios, bandwidth is still a shared, limited, and temporary resource. Efforts must be taken to manage bandwidth usage and be robust to communication loss when robots operate outside the range of the radios.
CHORD uses ROS 2 over the wireless mesh network for inter-agent communication. The advanced Quality of Service (QoS) features of ROS 2 are used to guarantee delivery of important priority data   while  maintaining  network  stability  over  low-bandwidth  links. This configuration enables traffic prioritization and resource control. We configured two categories of QoS for inter-robot topics with different mission requirements. Topics that require full message history transfer for post-processing or that deliver mission-critical information have higher priority and are configured so that the messages are reliably delivered even though the network may be down for periods of time. Estimated link capacities, data routes, and data traffic are also monitored to ensure stability.

\subsection{Evolution of CHORD}

\ph{Tunnel Circuit} During the Tunnel competition we used ROS 1 in combination with a custom cross-master messaging mechanism (multimaster-JPL) in combination with radios from Silvus Technologies and Persistent Systems \cite{Otsu2020}.
While our communication system and radios supported our six robot operation well during the tunnel competition, we observed some communication issues. We found that careful attention was needed to avoid ROS 1 attempting to share global topics (like TFs and diagnostics) across all robots. In addition, without better quality of service (QoS) controls, robots outside of the communication range of the backbone network would buffer all data and flood the network on their return. Some of our data products were also larger than expected. See \cite{Otsu2020} for more details. 

\ph{Urban Circuit} 
Before the Urban competition we switched to using ROS 2 for inter-agent communication in combination with radios from Silvus Technologies. We observed better performance than our previous ROS 1-only data sharing system. First, by using a different protocol for inter-robot  communication,  we  isolated  the  ROS  1  networks completely and avoided unintended data flows between agents. The  network  isolation  also  helped  to  diagnose  network issues  easily  as  every  inter-robot  ROS  topic  passes  through the  bridge  node. Second, we  were  able  to keep network traffic inside  our  bandwidth  budget,  which  contributed  to  the stability  of  the  dynamic  network. For more details on the results, see \cite{CHORD}.

\section{Mission Planning and Autonomy} \label{sec:mission_planning}

Having the capability to autonomously plan, reconfigure, and perform tasks for a multi-robot system is a crucial component of the NeBula autonomy framework (see \autoref{fig:nebula_architecture}), enabling exploration of large, complex, and unknown environments. Especially in the context of the SubT challenge, when there is none or unreliable intermittent communication between robots and a single human supervisor, autonomy is crucial to achieve mission objectives, under time and resource constraints. In this section, we present NeBula’s mission autonomy components, while integrating and allowing a single human supervisor to oversee and interact with a team of more than five heterogeneous robots at the same time, under range-limited and unreliable communication in challenging environments. For technical details, please see \cite{Otsu2020,vaquero2020traversability,kaufmann2021copilotIEEE}.

\subsection{Architecture}

\begin{figure}[b!]
    \centering
    \includegraphics[width=\linewidth]{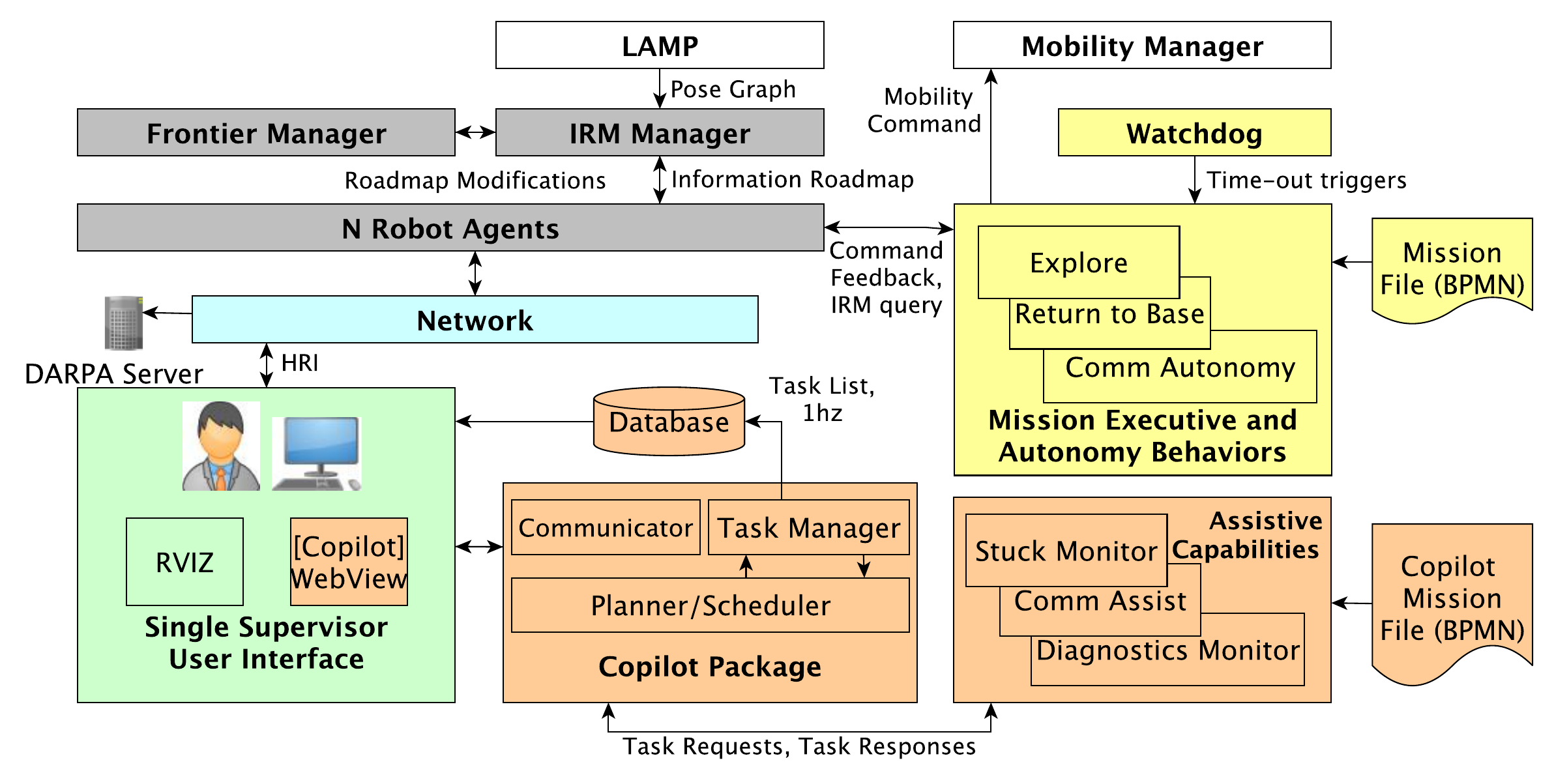}
    \caption{Mission Autonomy architecture and components (robot-side: yellow, base-side: orange).}
    \label{fig:ma-arch}
\end{figure}

\autoref{fig:ma-arch} illustrates the components of the Mission Autonomy architecture and their interface to components. In the following paragraphs, we describe key components of the system's mission planning and autonomy.

\begin{itemize}[leftmargin=20pt]
\item \textbf{Mission Executive}: The executive is responsible for stepping through the mission flow as defined and specified in the Mission File (see next point). The executive triggers the robot’s autonomy behaviors based on its mission state. 

\item \textbf{Mission File}: The Mission File defines a set of mission autonomy behaviors and their execution flow during a mission. 

\item \textbf{Autonomy Behaviors}: Mission Autonomy Behaviors implement the logic to interact with world beliefs and send commands to other modules or platforms. See \autoref{subsec:ma-onboard} for specific behavior implementations.
 
\item \textbf{IRM Manager}: 
The IRM is a key data structure to exchange information between humans and machines.
Certain robot autonomy behaviors and assistive capabilities utilize the latest belief states in the IRM within autonomous decision-making and planning processes.

\item \textbf{Mission Watchdog}: The Mission Watchdog monitors mission progress and communication links. It is an external mechanism that ensures that the robots can transmit their world belief back to the base station regardless of the Mission Executive states. %

\item \textbf{Copilot and Assistive Behaviors}: The mission autonomy assistant, Copilot for short, encompasses a series of monitoring and assistive capabilities (e.g. system health monitor, comm node drop assistance) that perform autonomous tasks and keep the human in the loop if possible and needed.
The details of Copilot, including its {Assistive Capabilities}, {Planning and Scheduling}, and human interaction via {User Interfaces}, are explained in~\autoref{subsec:ma-hri}.

\item \textbf{User Interface}: The User Interface consists of carefully designed web-based components and RViz displays for increased situational awareness. It provides an efficient interface to send various levels of commands with minimal control.
\end{itemize}

\subsection{Autonomy}
\label{subsec:ma-onboard}


Exploring unknown and complex environments with a team of multiple robots comes with several challenges. The difficulty of operating a single robot under limited available communications bandwidth, let alone a team, motivates the need for full autonomy during different phases of an exploration mission when communications are not available. In situations where robots are outside of each other's communication range, robots must remain fully autonomous and independently reason about their environment to determine their next task. A subset of this problem arises when robots are within each other's communication range and the robots must devise a coordinated plan. This section describes how we utilize various levels of mission autonomy.

\ph{Planning and Scheduling} Mission planning and task scheduling constitute the highest layer of the planning modules and are integral components in achieving full autonomy outside and within a communication range (\autoref{fig:conops_diagram}). The mission planner maintains a world belief, as described in \autoref{sec:conops}, which comprises the state of the robot team (e.g., robot health, robot location, detected artifacts on each robot), the state of the world (e.g., geometric and semantic maps), the state of the mission (e.g., remaining mission time, margin to desired mission output), and the state of communication (e.g., network connectivity, location of comm nodes). As the world belief increases and improves, the mission planner dynamically re-tasks robots to new goals or deploys new robots in order to achieve the mission goals defined in the mission specification. Ongoing work focuses on using semantic information about the world (e.g., stair wells, door frames, intersections, room volume) in the mission planner to provide additional belief of where critical information may lie to help improve mission success. The task scheduler works in conjunction with the mission planner and robot autonomy behaviors to actively schedule and assign agents for each task given various constraints. This is one of the mission autonomy features that allows the system to actively deal with temporal uncertainties, dependencies, dynamic resource constraints, and varying risks and mission strategies. The scheduling component is modular and can be interchanged with a variety of existing planning and scheduling frameworks. Currently, a planning problem is formulated using the Planning Domain Definition Language~\cite{Fox_pddl:modelling} and is solved using OPTIC ~\cite{benton2012temporal}. The solver updates the task schedule at a fixed cadence.

\ph{Executive} The Executive is a task manager that ensures each scheduled task is dispatched to each respective agent at the correct time. It tracks the state of all tasks and requests the task scheduler to reschedule when tasks fail, or relaxes temporal constraints when the current schedule is infeasible.

\begin{figure}[t!]
    \centering
    \includegraphics[trim={0 5.5cm .5 0.1cm},clip,width=0.95\linewidth]{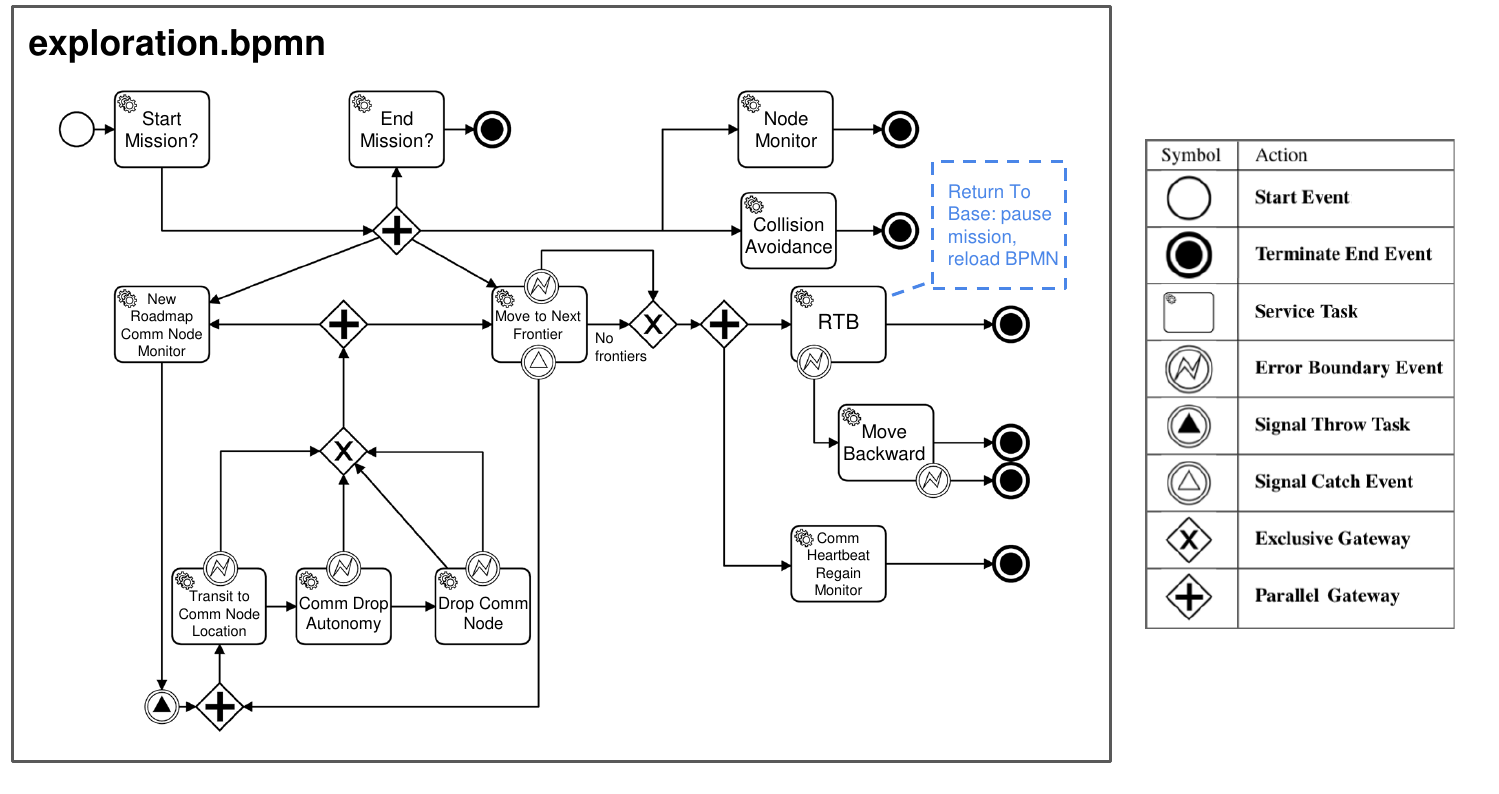}
    \caption{Annotated Exploration Mission File used during the self-organized cave circuit.}
    \label{fig:ma-full-mission}
\end{figure}

\ph{Mission Specification} A Mission File is used to describe a mission and combines several high-level Robot Autonomy Behaviors to create the flow for more complex scenarios. For this, we use the Traceable Robotic Activity Composer and Executive (TRACE) proposed in~\cite{de2017mission} and the Business Modelling Process Notation (BPMN). \autoref{fig:ma-full-mission} shows one example ``Exploration'' Mission File that was used in one of our Cave exploration scenarios. The mission starts off in the upper-left corner with a prompt for the user to begin the mission which ensures the robots can start moving autonomously in a safe manner. Each rectangular box depicts a Service Task\---they represent robot behaviors in our architecture. Once the mission is started, the mission flow moves towards the Parallel Gateway. This gateway allows multiple flows to branch off of a single flow. In the case of the Exploration BPMN, this allows five behaviors to run in parallel.

\ph{Robot Autonomy Behaviors} Behaviors used in the example Mission File (\autoref{fig:ma-full-mission}) represent a subset of NeBula's robot autonomy behaviors; some are described here: 
\vspace{-15pt}
\begin{itemize}[leftmargin=20pt]
    \item \texttt{Move to Next Frontier} - Receives the current frontier as a goal from the global planning module and commands the agent to move to it.
     \item \texttt{New Roadmap Comm Node Monitor} - Monitors the IRM for new mesh network extension requests and interrupts the \texttt{Move to Next Frontier} behavior temporarily to initiate \texttt{Comm Drop Autonomy}.
     \item \texttt{Comm Drop Autonomy} - Autonomous selection of optimal target location to drop a communication node to maximize communication coverage while minimizing the risk of violating safety and operational constraints for the robots traversing the local environment~\cite{vaquero2020traversability}.
    \item \texttt{Transit to Comm Node Location} -  Commands the agent to move towards the dropping location through the IRM.
    \item \texttt{Drop Comm Node} - Instructs the dropper firmware to drop a comm node immediately.
    \item \texttt{Collision Avoidance} - Prevents inter-robot collision by monitoring inter-robot distances. When robots are too close, this behavior performs a prioritized motion planning to resolve the situation.%
    \item \texttt{Comms Heartbeat Regain Monitor} - Continuously monitors for heartbeat messages from the base station and terminates successfully when a consistent stream of messages is detected.
    \item \texttt{RTB (Return to Base)} - Locates the agent within the current IRM and finds the shortest path to the Base IRM node. Then sends this path to the Mobility Manager and terminates successfully when the agent has reached the Base.
    \item \texttt{Stairs Helper} - Detects stairs and assists during a stair climbing procedure.
   
\end{itemize}

\subsection{Assistive Autonomy and Human-Robot Interaction}
\label{subsec:ma-hri}


In situations where there is sufficient communication bandwidth to interface with a team of multiple robots, having the capability to provide situational awareness of all robot activities and the mission progress to the operator becomes very useful. Assistive autonomy at the base station facilitates this human-robot interaction especially under limitations like the single supervisor requirement of the SubT challenge, or available cognitive workload.

\begin{figure}[t]
    \centering
    \includegraphics[width=\linewidth]{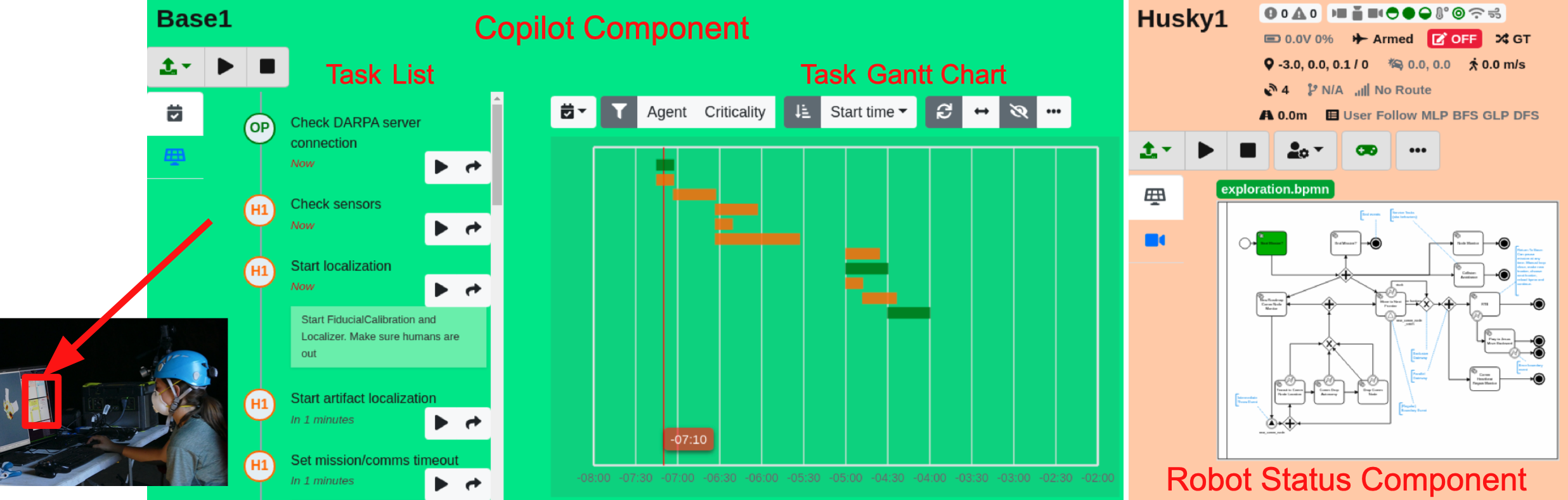}
    \caption{Base station single screen setup and human supervisor interacting with the user interface (left). Autonomy assistant Copilot MIKE's interface component with task list and Gantt chart mission overview (middle). Robot status component displaying the robot’s state in the exploration mission (compare \autoref{fig:ma-full-mission}), real-time information regarding sensor hardware, current pose and other housekeeping data that represents current beliefs (right).}
    \label{fig:ma-copilot-interface}
\end{figure}


\ph{Autonomy Copilot MIKE} The \textbf{M}ulti-robot \textbf{I}nteraction assistant for un\textbf{K}nown cave \textbf{E}nvironments (Copilot) is introduced in~\cite{kaufmann2021copilotIEEE} and supports the single human supervisor during the setup and mission phases of complex multi-robot operations. Copilot treats the human supervisor as one node (or a member) of the multi-robot system and it actively schedules and re-schedules the tasks for all members while considering world beliefs, resources and other constraints (e.g. human cognitive workload, available comm nodes, etc.). Some tasks can be automatically executed or resolved depending on the allowed autonomy level. Currently, Copilot comprises assistive capabilites, robot behaviors, assistive task scheduling, and user interfaces (\autoref{fig:ma-arch}) used to autonomously control the robots and guide the human supervisor's decision making while keeping them in the loop if a communication link is established.

\ph{Assistive Capabilities} Assistive capabilities are distributed across the robots and the base station to reduce the supervisor's workload. These capabilities assist to (1) detect system anomalies, (2) create tasks for the Copilot core module, (3) monitor the resolution of these tasks, and (4) resolve tasks autonomously or inform the supervisor that actions are needed. The assistive capabilities run independently without knowing the complete system status, which allows a modular extension of the autonomy functionality. Assistive capabilities address human limitations and autonomously intervene during mission execution as needed and when possible. Tasks created by these capabilities can inform the supervisor of the status of communication links between the robots and the base (e.g. communication loss, communication node drop suggestions), a robot's sensor suite status or mobility status (e.g. stuck, tilted, undesirable oscillations), and the system's autonomy software status (e.g. process health).


\ph{Assistive Task Scheduling} The assistive task scheduler shares the implementation of the task scheduler mentioned prior and considers the operator(s) as an agent (or resource) to which tasks can be assigned. As more tasks become automated and can be assigned to the Copilot, the operator becomes more readily available to assist with mission-critical tasks.

\ph{User Interface} The user interface consists of several web-based components and a configurable RViz view showing a 3D or 2.5D map representation of the explored environment \cite{Otsu2020}. \autoref{fig:ma-copilot-interface} illustrates the single screen interface which includes the Copilot and robot status components. The \textit{Copilot component} is designed to increase situational awareness, inform the supervisor of mission tasks, and dispatch urgent system notifications. The limited interaction between the system and the single human supervisor increases the supervisor’s trust in the system’s performance, as they are kept in the loop and may still intervene if and when a communication link is established and if the mission strategy requires a change (e.g, taking a more risky, more conservative posture). The user interface provides an additional \textit{robot status component} for each robot to directly interact with agents or conduct troubleshooting. \autoref{fig:ma-copilot-interface} gives an example of this component which is used when communication links exist; the human supervisor needs to shift their attention to a single robot to directly interact with it. This component comprises elements to reflect real-time housekeeping data, such as current position, sensor health status, estimated battery remaining and remaining comm nodes. It also integrates controls for changing a robot’s role to either act as a vanguard explorer or provide support to the leading agents in the mission. Lastly, the component integrates the current BPMN mission diagram and highlights an agents’ active autonomy behavior to reflect the robot internal state of the Mission Executive.

\ph{Human Trust in Autonomy} Copilot has been deployed in multiple field tests and mission simulations in diverse environments providing us with useful feedback regarding the interface design and multi-robot operations. Previous to the implementation of Copilot, we learned that our human supervisors were overwhelmed with remembering tasks instead of strategically overseeing the robots’ activities and overall mission progress. Now our human supervisors let the Copilot handle several decision making processes and must trust its autonomy capabilities~\cite{kaufmann2021copilotIEEE}, even if a supervisor could make their own decisions.

\section{Mobility Systems and Hardware Integration} \label{sec:hardware} 

\begin{figure}[h!]
    \centering
    \includegraphics[width=\linewidth]{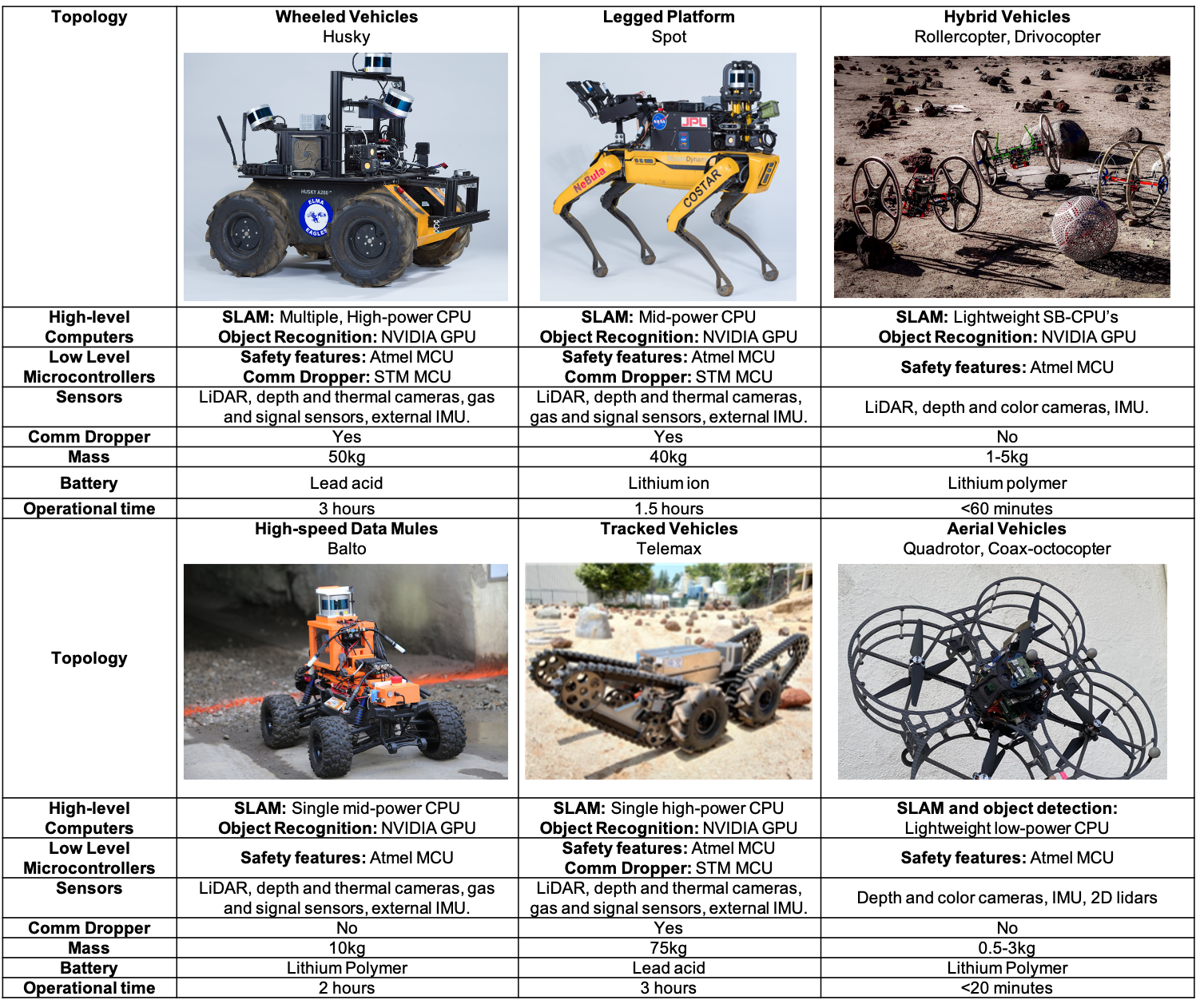}
    \caption{Hardware specifications for some of CoSTAR robots}
    \label{fig:HW1}
\end{figure}

As described in~\autoref{sec:conops}, to simultaneously address various challenges associated with exploring unknown challenging terrains, we rely on a team of heterogeneous robots with complementary capabilities in mobility, sensing, computing, and endurance. These assets are deployed in the mission as the robots learn about the terrain specifications. \autoref{fig:HW1} shows four classes of our robots: (i) Wheeled rovers: to cover general and relatively smooth surfaces and mild obstacles. (ii) Legged robots: to cover more challenging and uneven terrains where surmounting obstacles or staircases are required~\revv{\cite{AutoSpot,youtubespotpaper,miller2020,jenelten2020}}. (iii) Tracked robots: To complement legged platforms in handling different surface material, and (iv) Aerial and Hybrid locomotion: to enable traversing vertical shafts, and areas that are not accessible by ground robots.

\subsection{Ground robots} Our ground robots are able to carry heavy payloads. Hence they are equipped with high levels of sensory and processing capabilities enabling complex autonomous behaviors and artifact detection. Their battery capacity allows them to have longer operational time than flying vehicles. Below, we discuss the NeBula Autonomy Payload on these ground robots.

\begin{figure}[h!]
    \centering
    \includegraphics[width=0.5\textwidth]{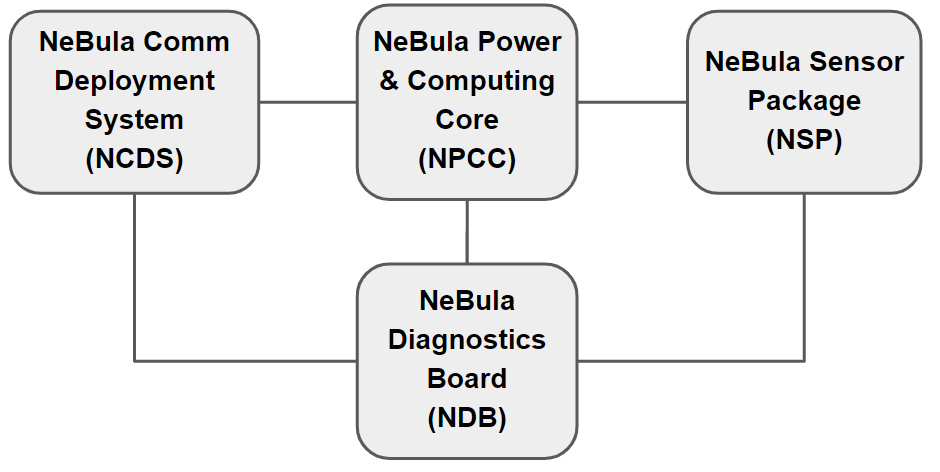}
    \caption{High-level overview of the ‘NeBula Autonomy Payload’ architecture}
    \label{fig:HW2}
\end{figure}

\ph{Architecture} The NeBula payload (\autoref{fig:HW2}) consists of the NeBula Sensor Package (NSP), NeBula Power and Computing Core (NPCC), NeBula Diagnostics Board (NDB), and NeBula Communications Deployment System (NCDS). Its electronics and software architecture is modular, to facilitate adaptation to varying mechanical and power constraints of each platform in our heterogeneous robotic fleet.

\ph{Design Principles} The key design principles are:
\vspace{-15pt}
\begin{itemize}[leftmargin=20pt]
    \item \underline{Durability:} Shock proofing of the system increases longevity and self-recovery chances when exploring challenging terrains.
    \item \underline{Lightweight materials:} Payload reduction maximizes robot agility and battery life.  
    \item \underline{Modularity:} A critical feature when developing a heterogeneous fleet of robots with various capabilities. For example, processors and sensors are adaptable across different mobility systems with differing mass/size constraints.  
    \item \underline{Portability:} Ease of transportation, minimize damage to valuable hardware.
    \item \underline{HW-autonomy co-design:} Adaptable design process to enable reconfiguring sensors to adapt to autonomy evolution.
\end{itemize}

\begin{figure}[h!]
    \centering
    \includegraphics[width=\linewidth]{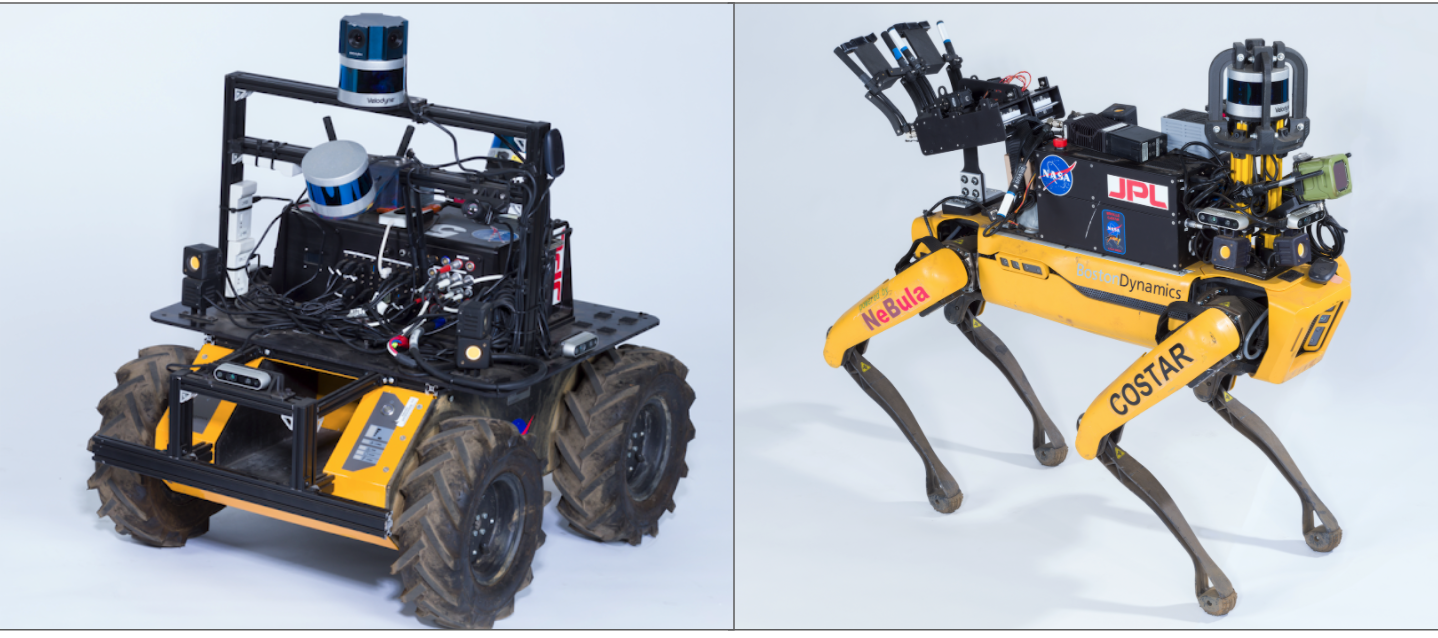}
    \caption{NSP equipped on Husky (left) and Spot (right).}
    \label{fig:HW3}
\end{figure}

\ph{NeBula Sensor Package (NSP)} The NSP empowers the NeBula autonomy solution on the robots by gathering sensory information in real-time from the environment. NSP is heterogeneous: on each robot NSP consists of a subset of the following sensors: LiDARs, monocular, stereo, and thermal cameras, external IMU’s, encoders, contact sensors, ultra high lumen LEDs, radars, gas sensors, UWB and wireless signal detectors. NSP is protected by custom superstructures with impact protection. A combination of hard resin urethane, semi rigid carbon-infused nylon, and aluminum were used. The NSP interfaces with the NPCC via high-bandwidth USB and Ethernet for data and custom serial messages and push-pull connectors for high-amp power. 
\rev{One version is depicted in \autoref{fig:HW3}, which shows NSP payloads utilizing an array of Velodyne Puck VLP-16, Intel Realsense d435i, and FLIR Boson 640, among several others; the sensors are interchangeable and reconfigurable to accommodate different sensor arrangements, manufacturers and sensing modalities.}


\begin{figure}[h!]
    \centering
    \includegraphics[width=\linewidth]{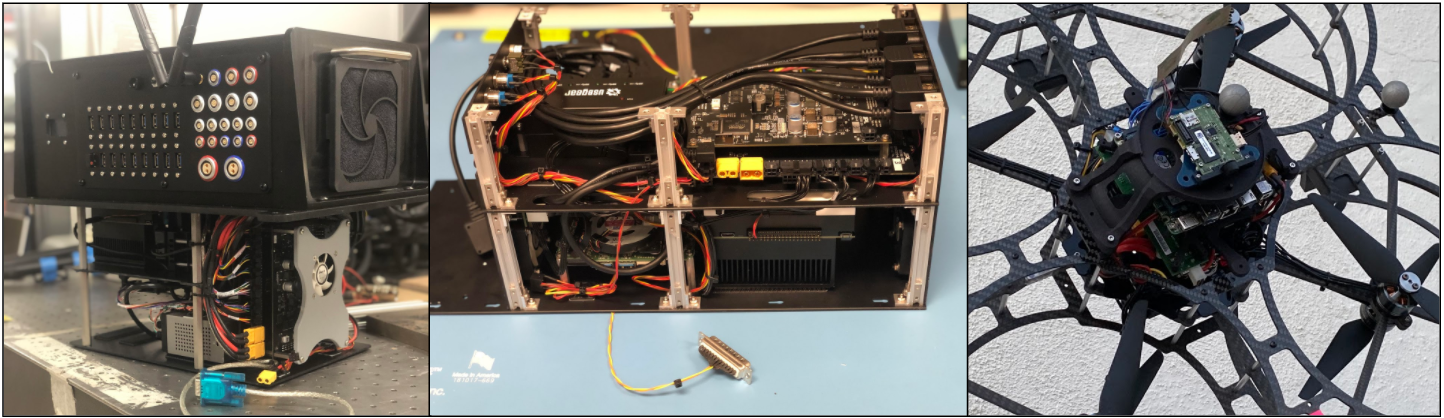}
    \caption{NeBula Power and Computing Core. Wheeled/tracked vehicles (left), Legged (middle), Aerial (right).}
    \label{fig:HW4}
\end{figure}

\ph{NeBula Power and Computing Core (NPCC)} The NPCC is an auxiliary payload which provides power to all NeBula sensors and computers used for autonomy. Aluminum enclosures provide protection to the internal electronics in the event of atypical loads and impacts such as falls and collisions. It is designed with considerations for thermal cooling and haptics due to extensive cycling of the connector interface panel. The modular, auxiliary-mount design was tweaked to accommodate for the reduced flight-weight of the drones. NPCC is powered from an external lithium high capacity battery to provide isolation and extended battery life for the internal battery across the heterogeneous fleet. The payload uses two high-power computers for sensing, autonomy, and semantic scene understanding and also hosts the low-level microprocessor. On some robots, NPCC is equipped with a GPU-based system-on-module with a custom interface to the power modules, cameras and sensors to accommodate machine learning and semantic understanding functionalities. The various configurations of the NPCC can be seen in \autoref{fig:HW4}. 

\ph{NeBula Diagnostics Board (NDB)} The NDB implements the system diagnostics, which monitors the vital power elements of the robot such as battery voltage, input current and individual regulator voltages. When the robot boots initially, all voltage regulators are powered up in a staggered sequence to limit the inrush loads to the NDB. After each voltage regulator is enabled, the processor checks that the voltage is within the expected range and reports errors if any are found. In addition, the current monitoring checks for high current draw when each of the regulators are enabled to detect possible short circuits in the various robot sensors and mechanisms. The power module has an input protection circuit to protect against voltage transients, reverse polarity and under voltage. A custom ROS message combines all these measurements and is constantly publishing the hardware power status to a specific robot diagnostic topic using the rosserial interface.

\begin{figure}[h!]
    \centering
    \includegraphics[width=0.5\linewidth]{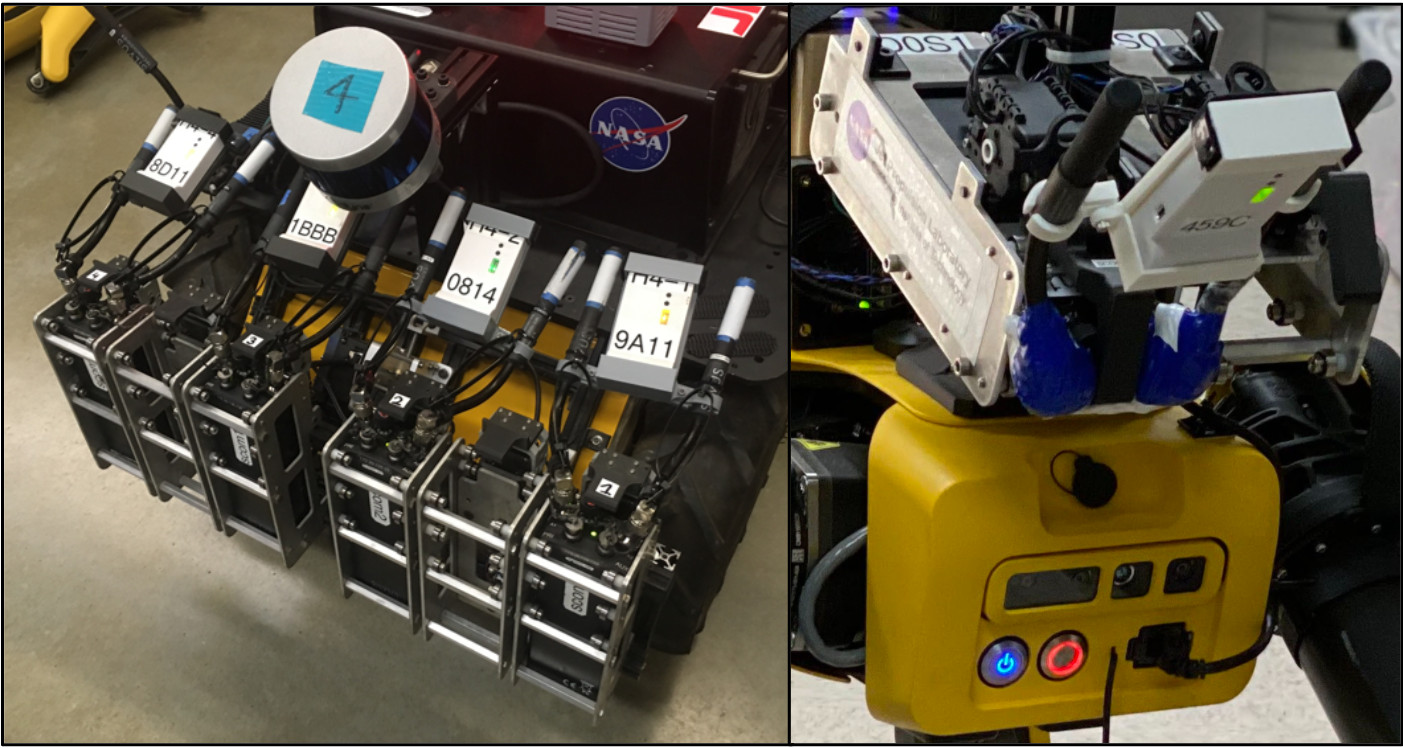}
    \caption{NCDS equipped on Husky (left) and Spot (right).}
    \label{fig:HW5}
\end{figure}

\begin{figure}[h!]
    \centering
    \includegraphics[width=0.5\linewidth]{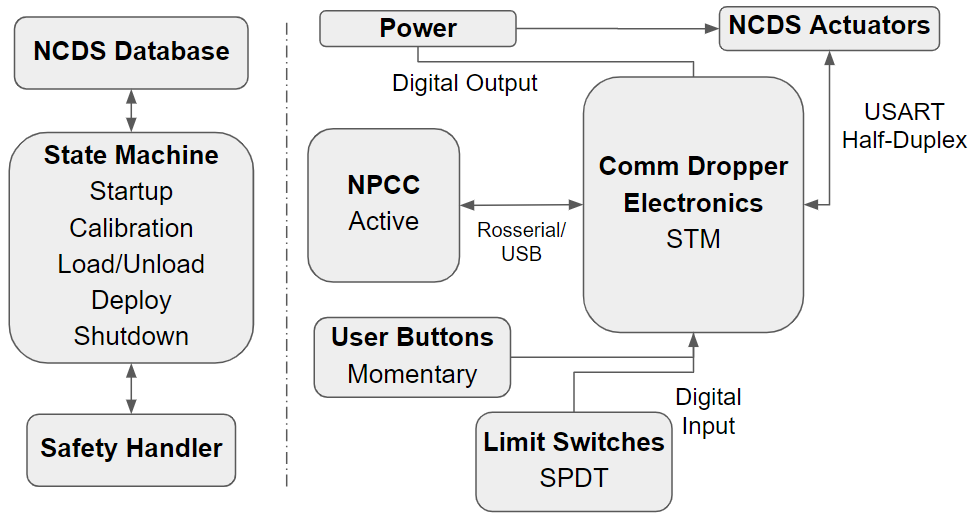}
    \caption{NCDS Software Architecture (left) and NCDS Hardware Architecture (right)}
    \label{fig:HW6}
\end{figure}

\ph{NeBula Comm Deployment System (NCDS)} As discussed in our ConOps in~\autoref{sec:conops}, we construct and expand a wireless mesh network near the environment entrance, to extend the reach of the base station. \rev {In order to do so, ground robots are equipped with a Comm Deployment System (NCDS), which allow them to carry and deploy communication radios (comm nodes) and static assets during the mission.} The radios are autonomously deployed using the NCDS which mounts at the rear of the robots seen in \autoref{fig:HW5}. The comm nodes are encased in aluminum until release, and the NCDS electronics are mounted locally on the mechanism and sealed for ingress protection. The core protection and brake-lock/release mechanism was modularized across the fleet though the deployment capacity was reduced for the legged and tracked vehicles due to available mounting points, payload constraints and sensor occlusion. It is driven by a geared brushless motor and activated via a custom embedded system communicating over rosserial. A ROS message provides the status of each radio (loaded, dropped).
The NCDS circuit board interfaces with the NPCC via serial communication. It monitors the hard stop switches used for calibration and the switches responsible for detecting if radios are loaded/released. A high level representation is depicted in \autoref{fig:HW6}. 


The NCDS software architecture relies on a finite state machine (FSM), with the following representative activities: 
\vspace{-15pt}
\begin{itemize}[leftmargin=20pt]
\item \underline{Start up:} Checks for battery power. Enables motor power and establishes links with motors.
\item \underline{Calibration:} Servos is actuated until limit switches are contacted. Encoder positional values are then stored in local memory. 
\item \underline{Load/Unload:} Manual switches allows the user to load and swap radios.
\item \underline{Deploy Radio:} Perform radio deployment. Use sensors to detect if deployment was successful.
\item \underline{Shutdown:} Disables motor power.
\end{itemize}

\subsection{Static assets \label{sec:hardware_static_asset}}
\begin{figure}[h!]
    \centering
    \includegraphics[width=0.25\linewidth]{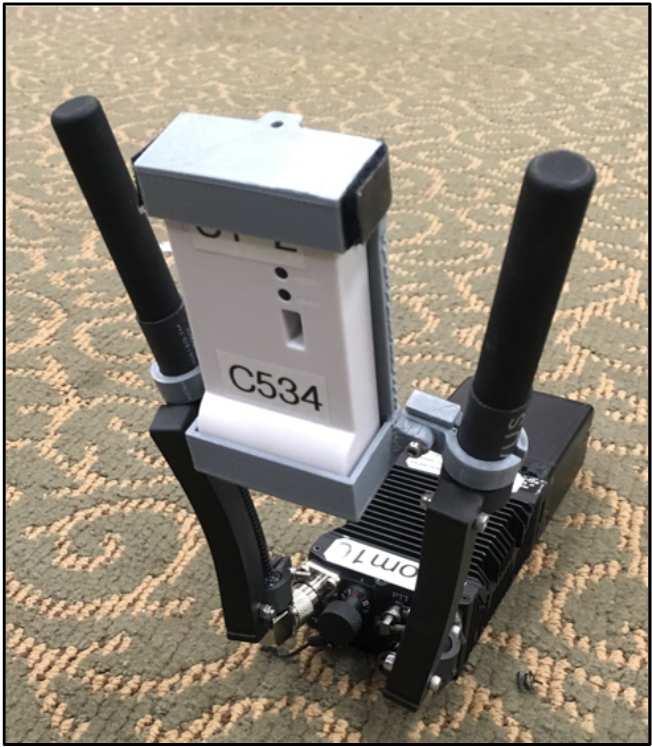}
    \caption{Static asset (comm node and UWB)}
    \label{fig:HW7}
\end{figure}

\rev{Static assets refer to hardware solutions which are not capable of changing their position after deployment. Our static assets are composed of: comm nodes and UWB modules as pictured in \autoref{fig:HW7}. The comm nodes construct NeBula’s mesh network to enable more efficient inter-robot information exchange, while the UWB provide auxiliary landmarks and provides ranging measurement to assist the SLAM and global localization modules, as explained in~\autoref{sec:lamp}. Static nodes can be powered prior to the start of a mission or prior deployment. They are carried by mobile robots and deployed by the NeBula Comm Deployment System (NCDS).}


\subsection{Flying vehicles}
NeBula has been implemented on several flying robots of varying sizes. Flying robots are responsible for exploring vertical shafts, areas not accessible with ground robots, or relaying data from the ground robots to the base station by quickly flying to regions with strong communication connectivity, e.g., for data muling (see \autoref{sec:multirobot_networking}). Their processors and sensing capability is much more constrained than our ground robots. The concept behind the drone development is to keep a balance between the payload, the size of the drone, and the endurance. Thus, an iterative design has been performed in order to conclude in an optimal hardware configuration \cite{jung2021robust}. \autoref{fig:X1} shows several NeBula-powered aerial vehicles: 
\begin{figure}[h!]
    \centering
    \includegraphics[width=0.5\linewidth]{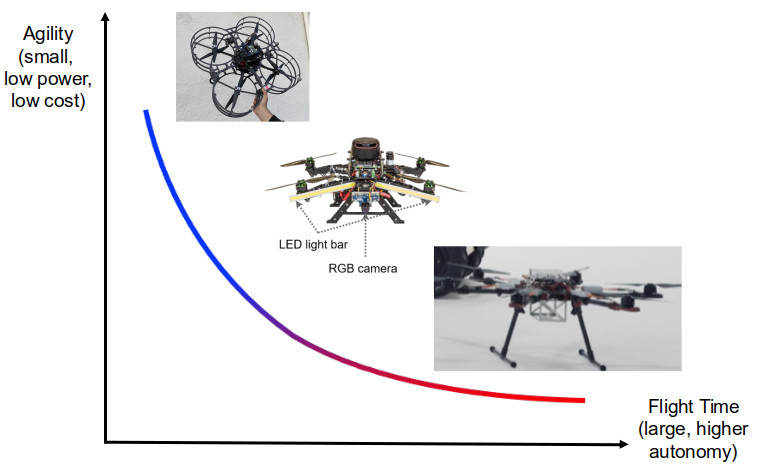}
    \caption{Flight time of similarly-sized vehicles with different autonomy capabilities. The general trend for aerial robots is long flight times are possible at the expense of large vehicles with less access and agility.}
    \label{fig:X1}
\end{figure}

\begin{figure}[h!]
    \centering
    \includegraphics[width=0.7\linewidth]{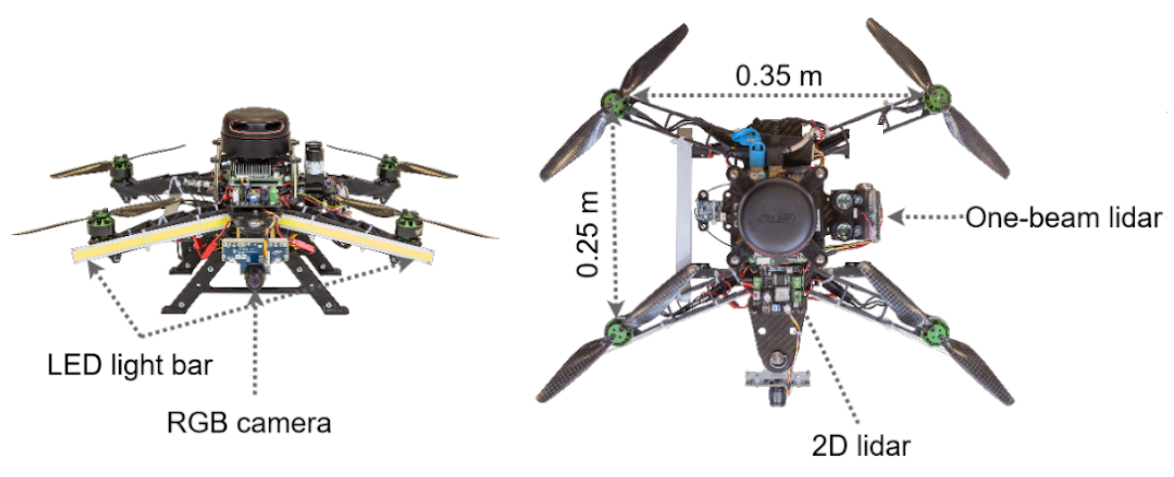}
    \caption{Drone equipped with a forward looking camera, a LED light bars, optical flow, 2D \rev{LiDAR} and single beam \rev{LiDAR} looking upward and downward.}
    \label{fig:X2}
\end{figure}

A specific example is shown in \autoref{fig:X2}, a custom drone extensively utilized in this project that carefully balances speed, weight, autonomy capability, and flight time. The vehicle’s weight is 1.5 kg and provides 12 minutes of flight time. A 2D rotating RP\rev{LiDAR} A2 is mounted on top of the vehicle, providing range measurements at 10 Hz and a monocular visual sensor at 30 FPS. The velocity estimation is based on the PX4Flow optical flow sensor at 20 Hz, while the height measurements are provided by the single beam \rev{LiDAR}-lite v3 at 100 Hz, both installed on the bottom of the vehicle pointing down as indicated in \autoref{fig:X2}. Furthermore, the aerial platform is equipped with two 10 W LED light bars in both front arms for providing additional illumination for the forward looking camera and four low-power LEDs pointing down for providing additional illumination for the optical flow sensor.

\subsection{Hybrid}
To extend the range of the flying drones, our team has been investigating and designing new hybrid ground/aerial vehicles, referred to as ``rollocopters". Rollocopters are designed to mainly roll on the ground; and when rolling is not possible, they can fly, negotiate a non-rollable terrain element, and land on the other side, and continue rolling. 
The combined rolling/flying behavior can extend the operational lifetime by several folds. 
Furthermore, the structure used for rolling the robot is designed to provide impact resilience while flying which provides further robustness. 
Our hybrid platforms consist of an adapted version of the above-mentioned NeBula payload along with the pixhawk firmware \cite{meier2015px4}. 
\cite{fan2020autonomousRollocopter} describes the details of integration of the NeBula autonomy stack on the  rollocopter platform. 

\begin{figure}[h!]
    \centering
    \includegraphics[width=\textwidth]{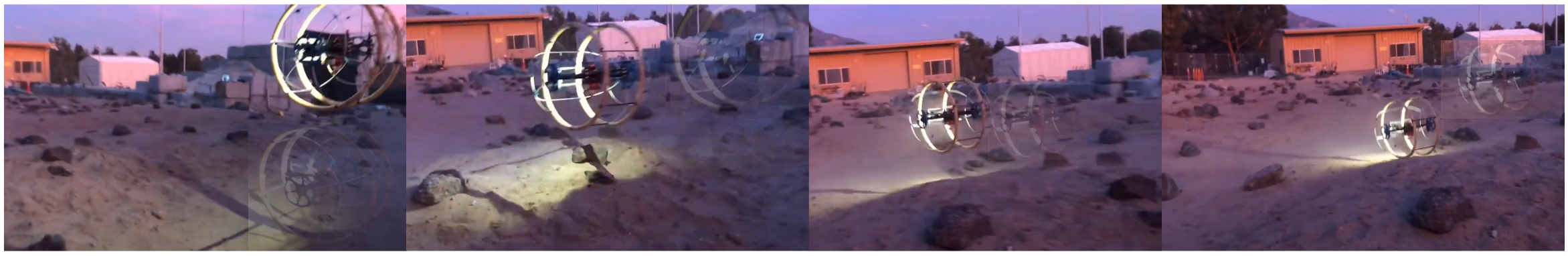}
    \caption{Hybrid platform flying over an obstacle at the JPL Mars Yard \cite{rolloearly}.}
    \label{fig:hybrid_test}
\end{figure}

\begin{figure}[h!]
    \centering
    \includegraphics[width=\textwidth]{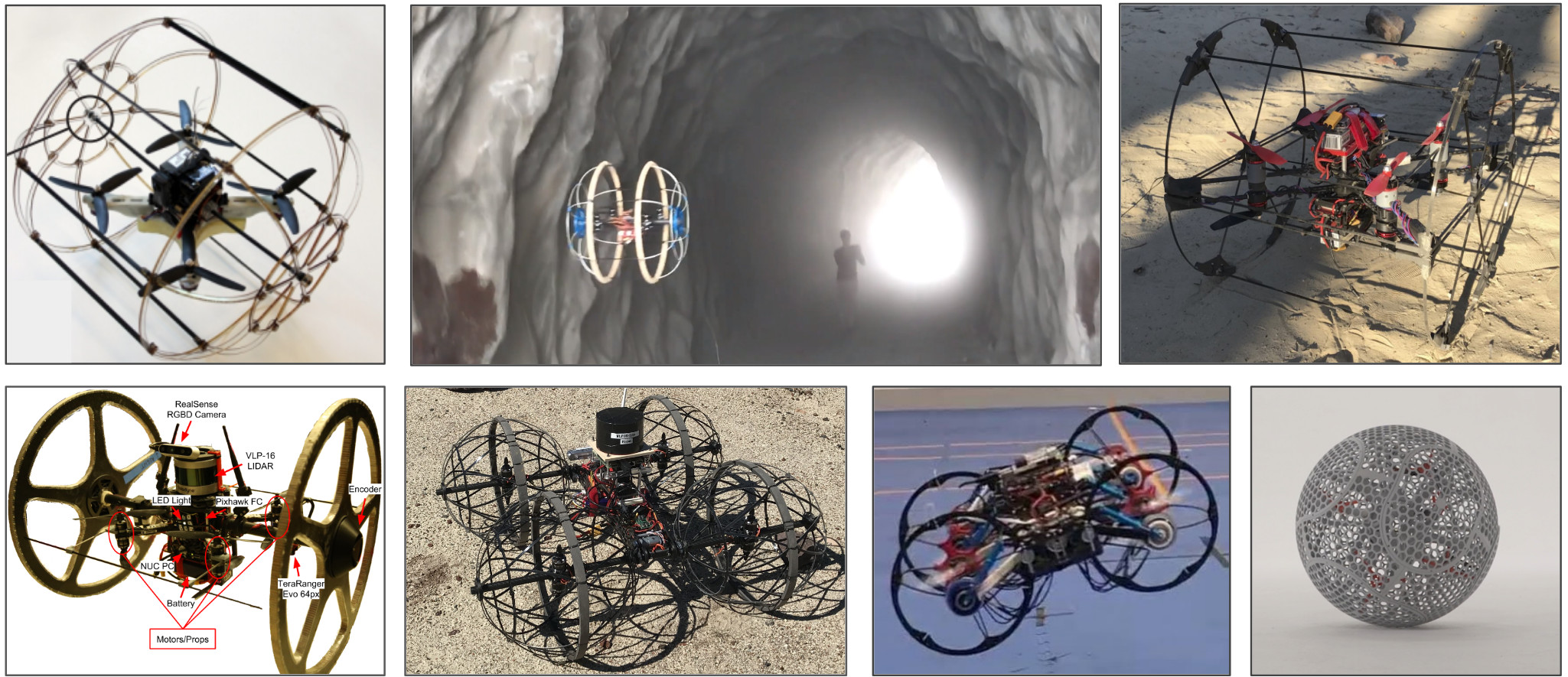}
    \caption{Hybrid platform hardware evolution (top row: left to right) pre-challenge Hytaq design \cite{kalantari2014modeling},  Modular hybrid vehicle \cite{agha2020shapeshifter},  (bottom row: left to right) Autonomous Rollocopter \cite{fan2020autonomousRollocopter,lew2019contact}, DrivoCopter \cite{kalantari2020drivocopter}, \rev{BAXTER} \cite{baxter}, Spherical Rollocopter \cite{sabet2020dynamic,sabet2019rollocopter}.
}
    \label{fig:hybrid_platform}
\end{figure}

\pr{Design Evolution} \autoref{fig:hybrid_platform} shows different variations of the rollocopter platform. 
The Hytaq version \cite{kalantari2014modeling} consists of a rotating cage, which encloses the drone to enable rolling.
Figure \ref{fig:hybrid_test} shows a time-lapse for early hybrid mobility tests on reinforced versions of this platform at JPL's Mars Yard \cite{rolloearly}.
Its modular, multi-agent version, called Shapeshifter ~\cite{agha2020shapeshifter,shapeshifter2}, self-assembles this shell using permanent electromagnets. Passive-wheeled-rollocopter (PW-rollocopter) \cite{fan2020autonomousRollocopter,lew2019contact} uses two independent passive wheels to enable large sensory FOV and autonomy. To reduce propeller-generated dust, Drivocopter \cite{kalantari2020drivocopter} and BAXTER \cite{baxter} implement dedicated wheeled actuators, at the expense of a slight reduction in flight time.
The latest work is focused on omnidirectional spherical rollocopters that provide maximum agility and maneuverability while having the minimum friction  \cite{sabet2020dynamic,sabet2019rollocopter}. 

\section{Experiments} \label{sec:experiments}
In this section, we present Team CoSTAR's multi-year effort to validate NeBula technologies on physical and virtual systems. We first briefly discuss our simulation-based validation approach, and then discuss the performance of NeBula on heterogeneous physical robot teams in various challenging real-world environments. The videos in ~\cite{youtubeurbanandtunnelvideo,youtubespotpaper,youtubeAutonomousCaveExploration,youtubeBDLavaTube, youtubeMarsDog} depicts some highlights of these runs.

\subsection{Simulation Results}
\ph{Simulator Configuration} We use simulations both for component development and integration testing. NeBula relies on multiple simulator configurations at different fidelity levels to enable faster and more focused development. Examples include a high-fidelity Gazebo simulator for testing ground vehicle behaviors, flight stack software-in-the-loop simulations for analyzing flight performance, a docker-based multi-robot networking simulator, and a low-fidelity dynamics simulator for Monte-Carlo analysis. Our simulator setup is portable, and used in local and cloud environments depending on the computational resources it requires and the number of agents that are deployed during the test.

\begin{figure}
    \centering
    \includegraphics[width=0.9\textwidth]{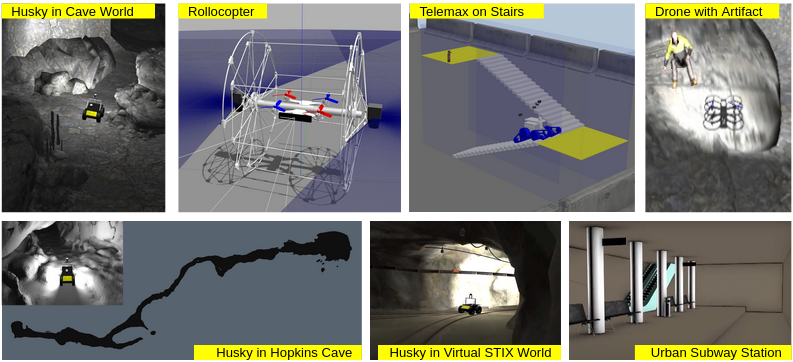}
    \caption{Robot and environment simulation models}
    \label{fig:simulation_model}
\end{figure}

\begin{figure}
    \centering
    \includegraphics[width=0.9\textwidth]{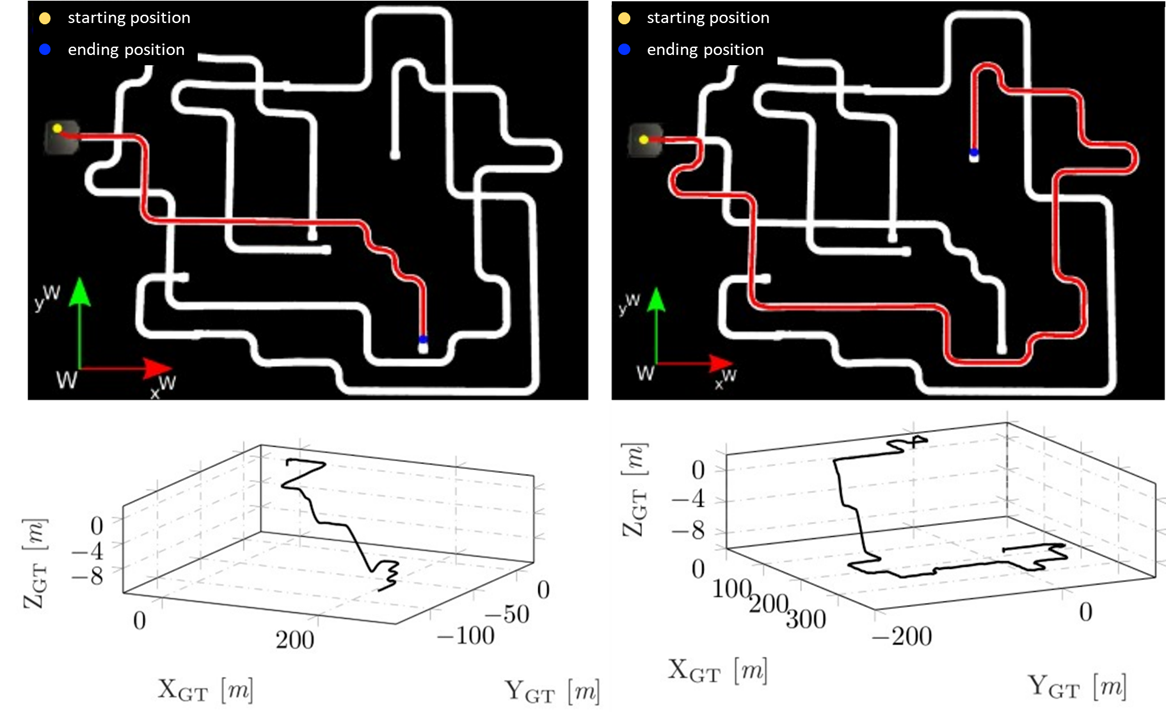}
    \caption{Drone navigation performance comparison in low-light tunnel environment with perception-aware nonlinear MPC, and heuristics-based junction sequence following.}
    \label{fig:drone_sim}
\end{figure}

\ph{Robot and Environment Models} \autoref{fig:simulation_model} shows selected examples of our simulated robot and environment models. In addition to many existing open-source models including the ones available at SubT Tech Repo \cite{subt_tech_repo}, we build our custom robot and environment models, some of which are submitted to and available on the same repository. For example, \autoref{fig:drone_sim} shows a comparative study of drone navigation algorithms using the robot and environment models that are publicly available on the repository.

\begin{figure}
    \centering
    \begin{subfigure}{0.5\textwidth}
        \includegraphics[width=0.98\textwidth]{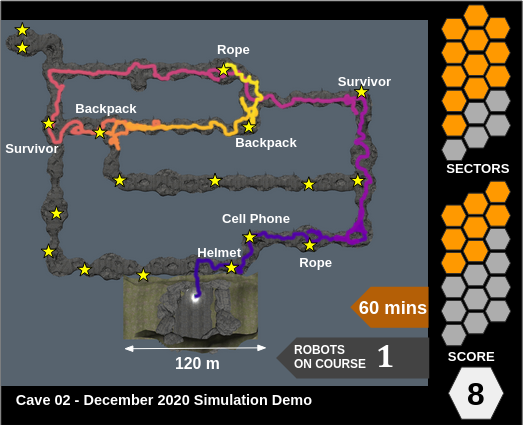}
        \caption{Run Summary}
    \end{subfigure}
    \begin{subfigure}{0.45\textwidth}
        \begin{tabular}{|l|l|}
            \hline
            \textbf{Exploration statistics} &  \\
            \hline
            Distance traveled & 2056 m \\
            Sector covered & 12 / 17 \\
            Rate of area covered & 557 m$^2$/min \\
            Cumulative stop time & 165 secs \\
            Operator intervention & 3 times \\
            \hline
            \textbf{Localization statistics} &  \\
            \hline
            Mean translational error & 0.001\% \\
            Mean rotational error & 0.00003 rad/m \\
            Max memory & 9.1 GB \\
            \hline
            \textbf{Artifact statistics} &  \\
            \hline
            Artifact scored & 8 / 18 \\
            Missed artifacts & \\
            ~~Out of FOV & 1 \\
            ~~No robot nearby & 9 \\
            \hline
        \end{tabular}
        \caption{Run Statistics}
    \end{subfigure}
    \begin{subfigure}{0.55\textwidth}
        \includegraphics[width=0.98\textwidth]{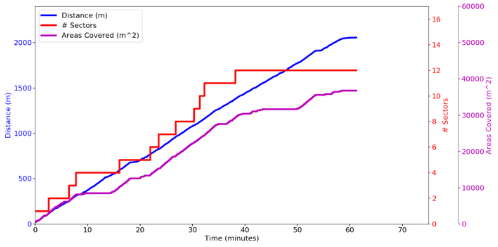}
        \caption{Exploration Efficiency}
    \end{subfigure}
    \begin{subfigure}{0.4\textwidth}
        \includegraphics[width=0.99\textwidth]{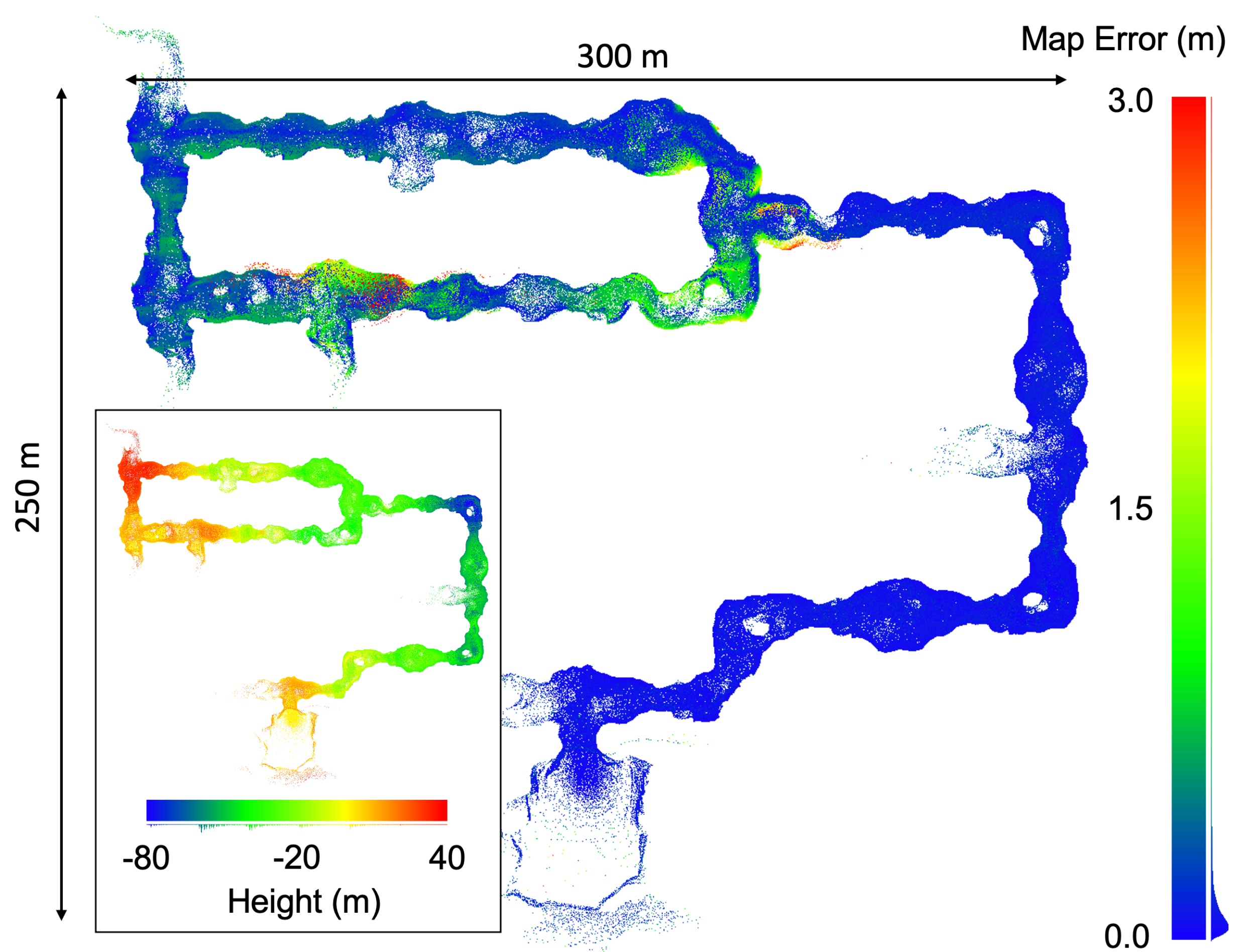}
        \caption{Localization Accuracy}
    \end{subfigure}
    \caption{Sample result from Team CoSTAR’s regular virtual demos}
    \label{fig:simulation_stats}
\end{figure}

\ph{Self-organized Simulation Events} We regularly organize simulation-based events (``virtual'' demos \rev{emulating the virtual DARPA SubT Challenge, see \cite{BARCS2020}}) to track performance statistics over time. In each virtual demo, we evaluate the latest system performance based on several evaluation metrics. We developed a set of automated analysis tools to evaluate statistics on exploration, localization, mapping, artifact scoring, and human intervention. Simulation enables us to measure the statistics from large-scale tests which cannot be obtained easily with hardware experiments. \autoref{fig:simulation_stats} shows exploration and operation statistics of a single robot simulation in the Gazebo simulator. The competition statistics (\autoref{fig:simulation_stats}-a) provide high-level evaluation with the number of sectors covered, number of artifacts scored, and robot trajectories in the course. The exploration statistics (\autoref{fig:simulation_stats}-b,c) evaluate how quickly and efficiently the autonomous exploration behavior covered the large-scale environment. The human supervisor intervention is monitored to measure the reliability of the autonomous system (\autoref{fig:simulation_stats}-b). The localization and mapping performance (\autoref{fig:simulation_stats}-b,d) is evaluated against the ground-truth dataset generated from the simulator. Finally, the artifact scoring performance is quantified with detection and localization evaluation (\autoref{fig:simulation_stats}-b).

\subsection{Field Tests and Demonstrations}
\begin{figure}[!hbt]
    \centering
    \includegraphics[width=0.8\textwidth]{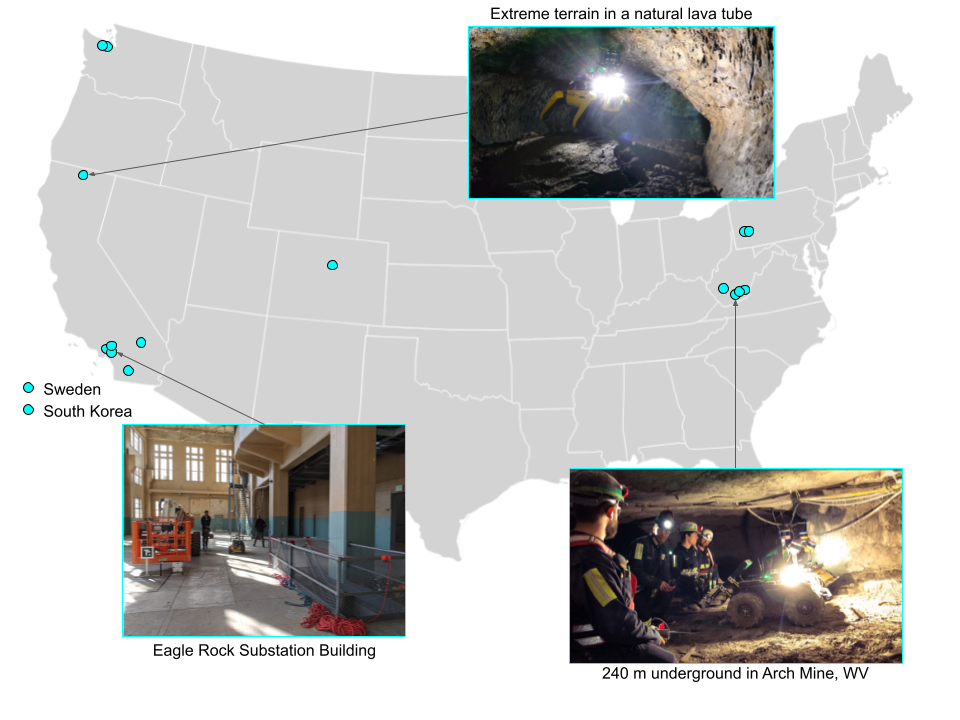}
    \caption{Locations and snapshots from field test activities. Each blue circle is a different field-test location.}
    \label{fig:field_test_locations}
\end{figure}

Our system has been rigorously field-tested in over 100 field tests in 17 different off-site locations, from lava tubes to mines 240m underground as well as numerous on-site locations including JPL's Mars Yard. This testing regime is a fundamental part of field-hardening our system to perform robustly in real-world challenging environments. A snapshot of some of our field test locations is shown in \autoref{fig:field_test_locations}. In this section, we discuss highlights of our system's performance at these self-organized field tests and demonstrations.

\begin{figure}[h!]
    \centering
    \includegraphics[width=0.8\textwidth]{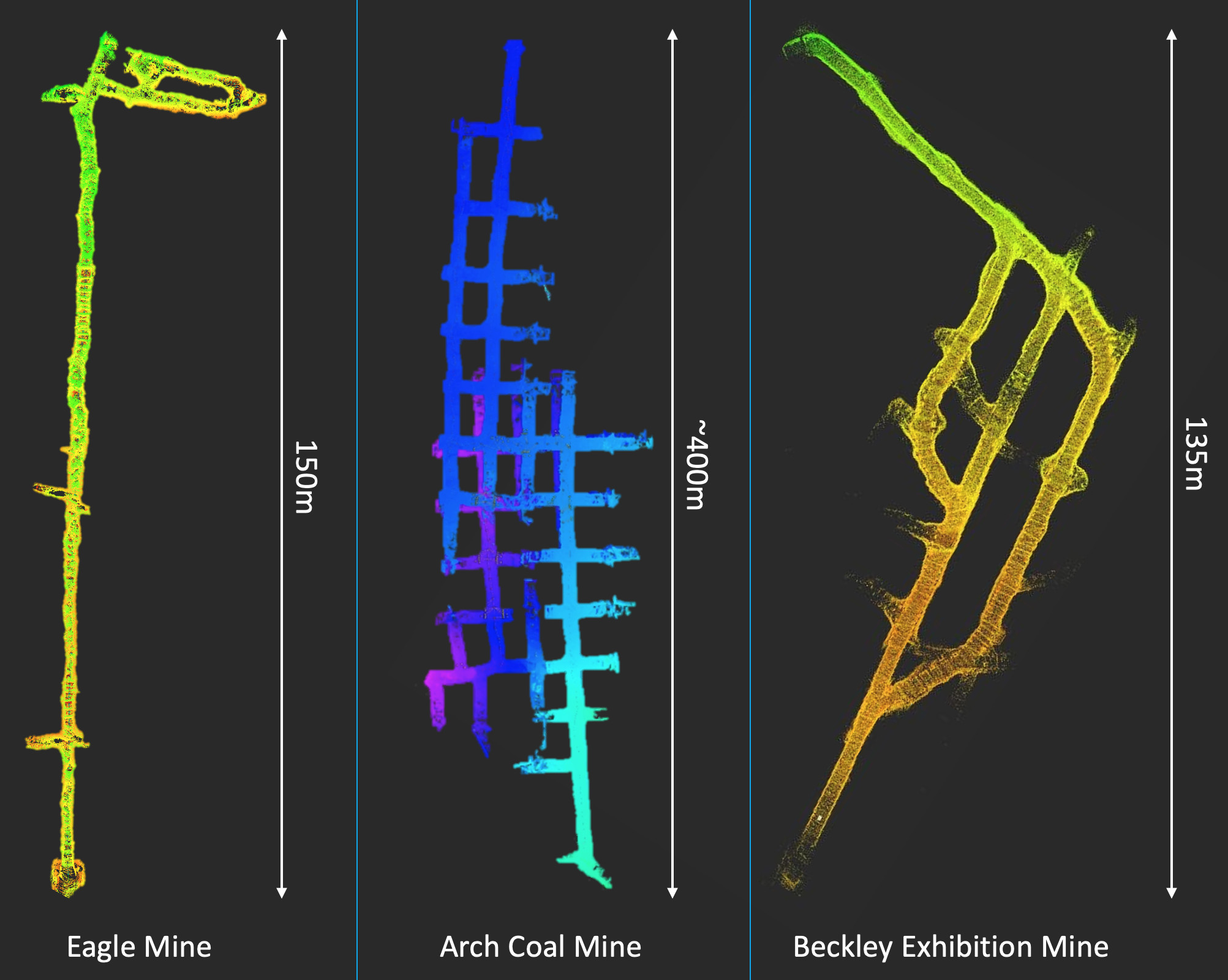}
    \caption{SLAM point cloud maps from field tests in different mine locations. Left: a silver mine in southern California. Center: a portion of a large coal mine in West Virginia, 800ft underground. Right: A near-surface historic coal mine in West Virginia.}
    \label{fig:tunnel_self_org}
\end{figure}

\ph{Mine and Tunnel Field Tests--Ground Vehicles} Field tests in tunnel environments as well as various underground mines stressed a variety of system capabilities, from dusty silver mines with narrow passages, to muddy coal mines with multiple decision points and massive scale. \autoref{fig:tunnel_self_org} presents three maps from these tests. The Arch Mine tests (\autoref{fig:tunnel_self_org}-center) were in a portion of an active coal mine, and included extreme traversability hazards (see \autoref{fig:TravTerrain}), complex topology, and severely degraded communications. The multi-robot map is the result of a 30 minute operation, with a total of $400 \times 100$ m$^2$ of area covered by the two-robot team. Eagle Mine and Beckley Exhibition mine are maps from a single robot. 

\begin{figure}[h!]
    \centering
    \includegraphics[width=0.8\textwidth]{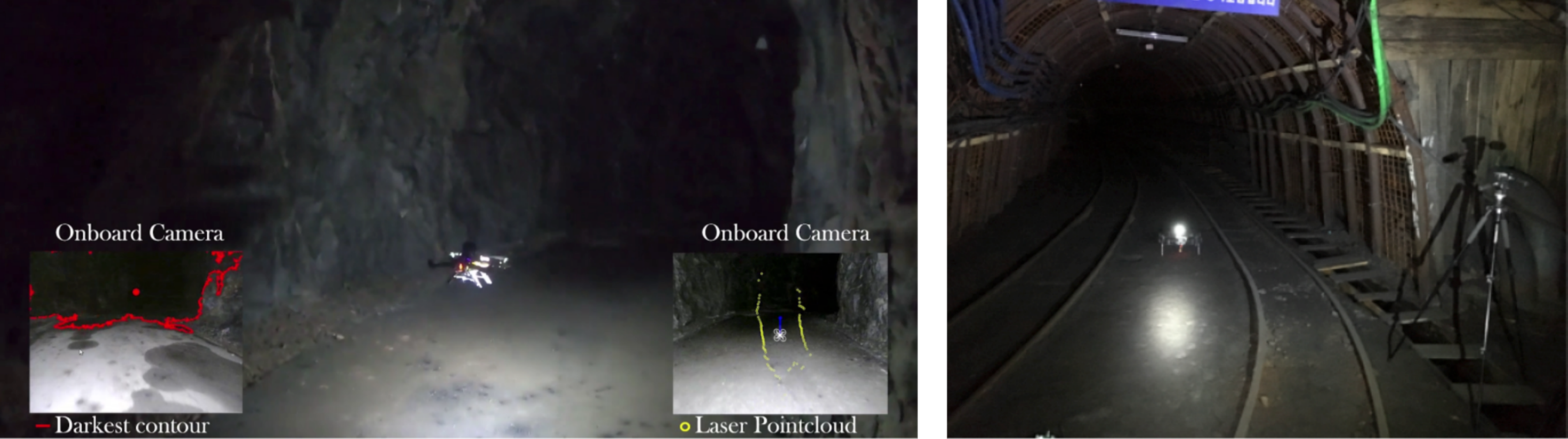}
    \caption{Autonomous flights in Mj\"{o}lkuddsberget Mine, Sweden (left) and Hwasoon Mine, Korea (right).}
    \label{fig:drone_tests}
\end{figure}

\begin{figure}[h!]
    \centering
    \includegraphics[width=0.8\textwidth]{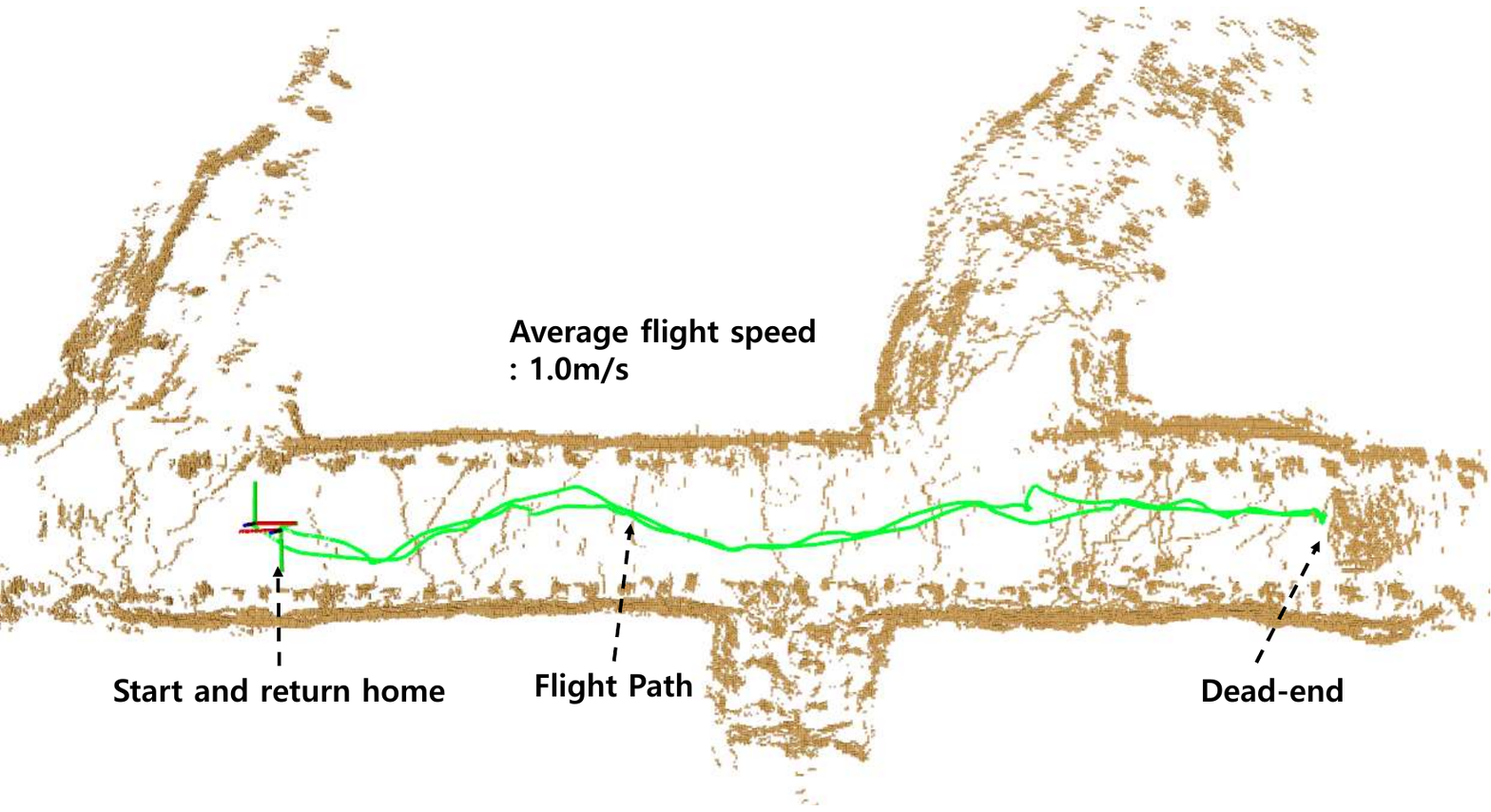}
    \vspace{-27pt}
    \caption{Autonomous flight test at Beckley Mine, Morgan Town, WV. In this example, the drone flies approximately 50m over 45s. While flying, the drone meets the dead-end and return to the communication range to deliver the map and artifact data before explore other branches of the mine.}
    \label{fig:drone_tests_beckley}
\end{figure}

\ph{Tunnel Field Tests--Aerial Vehicles} Aerial vehicles are deployed to perform fast navigation and exploration, while providing a rough topological map and collecting data from areas inaccessible to ground robots. \autoref{fig:drone_tests} shows two of the several hundred successful flight experiments conducted in tunnels located in the United States, South Korea, and Sweden. One specific example of drone flight is shown in \autoref{fig:drone_tests_beckley} \cite{youtubeScout}. In this example the drone shows successful flight autonomy capability with local planning method in confined and cluttered environment. The reliability of the developed system is best exemplified with the well over a thousand logged minutes of safe autonomous flight in complex and perceptually challenging tunnel environments.

\begin{figure}[h!]
    \centering
    \includegraphics[width=0.8\textwidth]{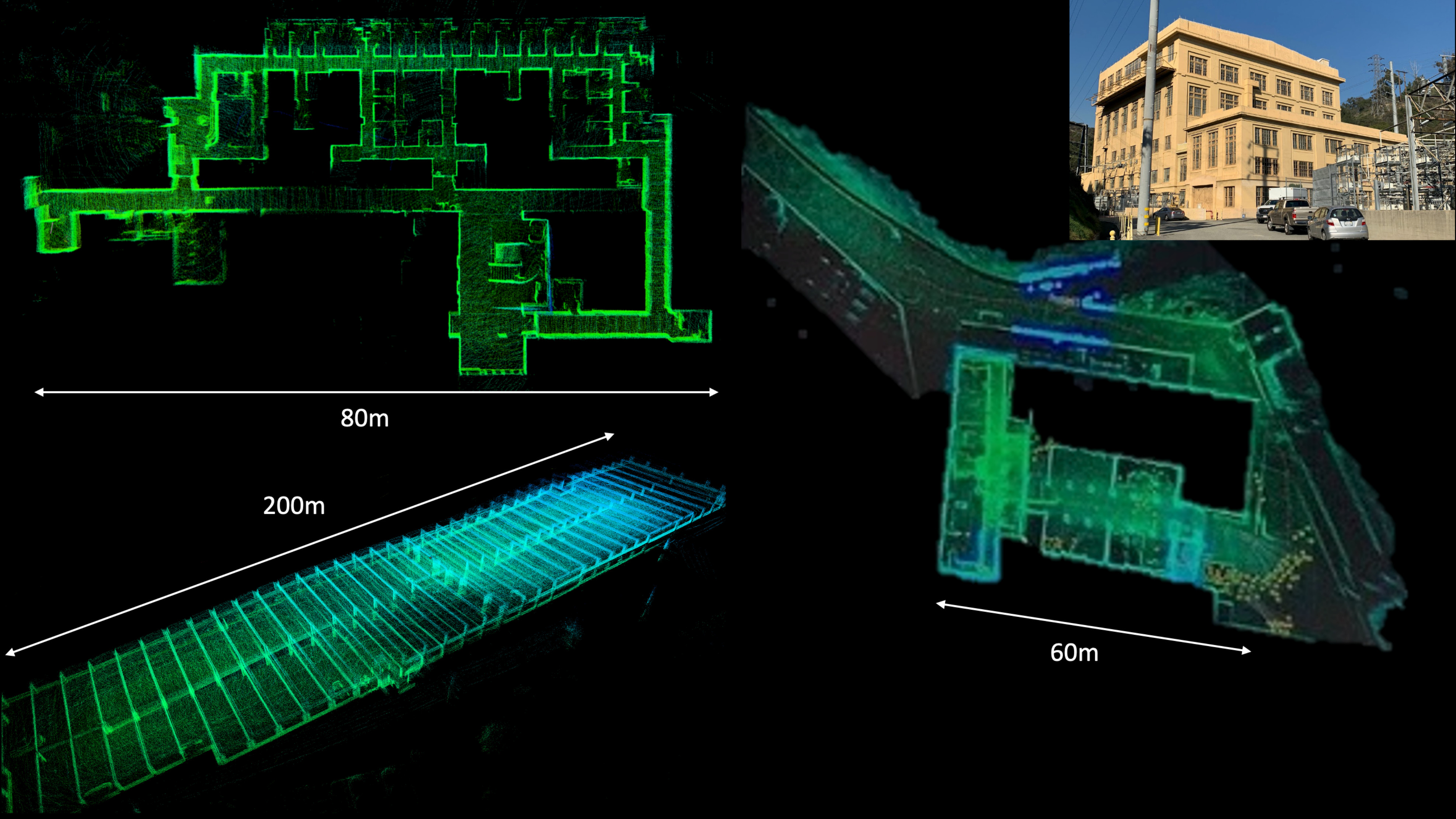}
    \caption{SLAM point cloud maps from field tests in different locations. Top Left: an office building at JPL. Bottom Left: A multi-level parking structure at JPL. Right: Eagle Rock substation and surrounds (pictured at top right)}
    \label{fig:urban_self_test}
\end{figure}

\ph{Urban Field Tests--Ground Vehicles} Urban field testing provides more abundance of possible locations, but also a large variety of challenging conditions. We tested our system, in narrow cubicle farms of multi-level office buildings, with regions of self-similar corridors (\autoref{fig:urban_self_test}-top-left); in multi-level parking structures with wide open spaces (\autoref{fig:urban_self_test}-bottom-left), and in large, open industrial buildings with high ceilings, narrow stairways and narrow doorways (\autoref{fig:urban_self_test}-right). 

\begin{figure}[h!]
    \centering
    \begin{subfigure}{\textwidth}
      \centering
        \includegraphics[width=0.8\textwidth]{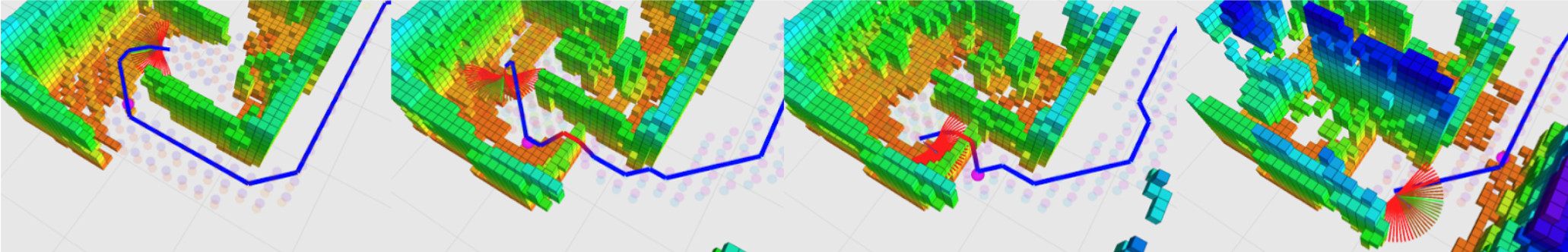}
        \caption{Time sequence of planned hybrid transition from rolling to flying and back to rolling.  Colored blocks are occupied voxels colored by z height.  Blue/Red path indicates hybrid planned path where blue is rolling and red is flying.  Pink sphere is goal waypoint.  Motion primitives from local planner are shown.  Note that as the vehicle moves forward, an obstacle is revealed and a small hop over it is planned and executed.}      
        \label{fig:rollo_map}
    \end{subfigure}
    \begin{subfigure}{\textwidth}
        \centering
        \includegraphics[width=0.8\textwidth]{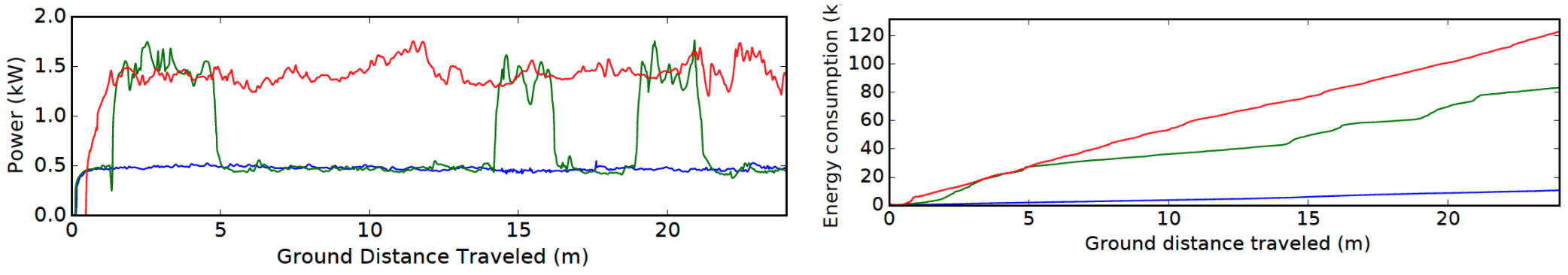}
        \caption{Power-energy consumption comparison with hybrid (green), rolling-only (blue), and flying-only (red) mobility modes. The hybrid mode flies over three obstacles, while rolling-only mode was tested without any obstacles.}
        \label{fig:rollo_power}       
    \end{subfigure}
    \caption{Rollocopter mobility mode comparison}
    \label{fig:rollo_iros}
\end{figure}

\ph{Urban Field Tests--Hybrid Vehicles} Hybrid ground/aerial mobility shows its benefit in the urban environment where there are plenty of easy-to-roll flat surfaces as well as vertical openings to navigate in multi-level environments. Autonomous hybrid mobility testing results \cite{fan2020autonomousRollocopter} show the benefit of hybrid mobility compared to pure ground or flying vehicles. The hybrid mode allows the robot to fly over obstacles that block the way for ground robots (\autoref{fig:rollo_map}). In comparison to the pure aerial vehicles, the hybrid mode shows better energy consumption profile by rolling on easy flat terrain (\autoref{fig:rollo_power}). 

\subsection{DARPA Organized Events} 
\begin{figure}[!h]
    \centering
    \includegraphics[width=0.8\textwidth]{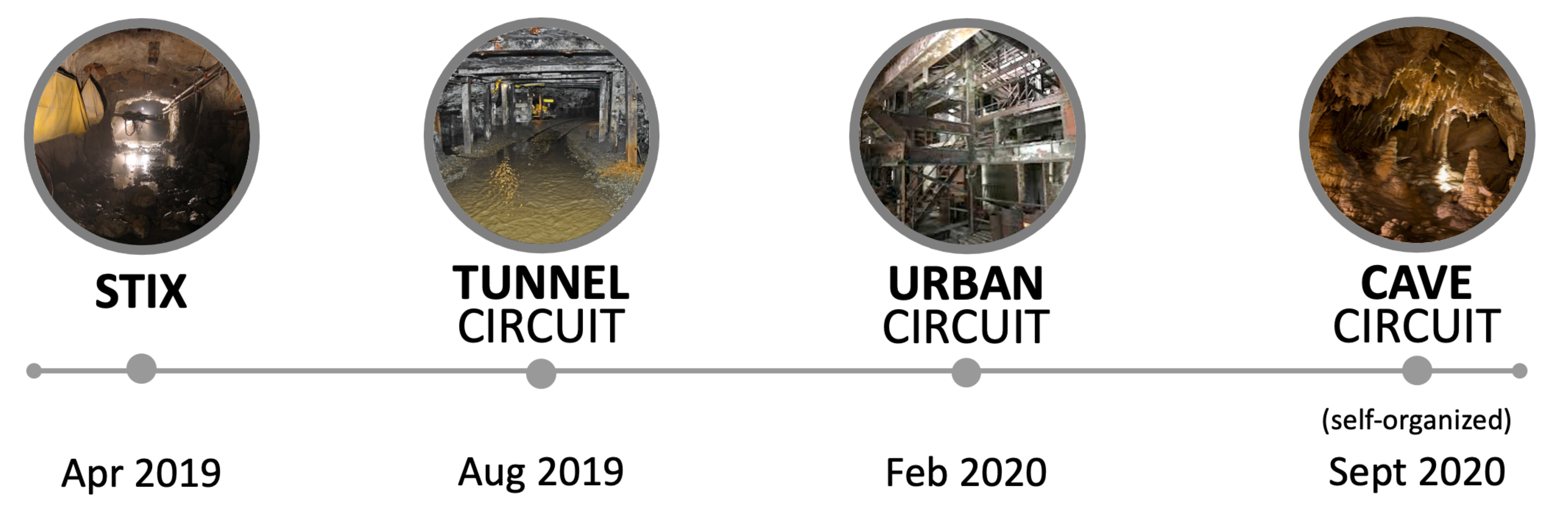}
    \caption{Overview of the timeline of DARPA organized test activities, with the expectation of increasing capability from one event to the next.}
    \label{fig:darpa_timeline}
\end{figure}

This section outlines results from testing NeBula in the series of DARPA facilitated test events as part of the Subterranean Challenge~\cite{darpaJFR}. The timeline of these events is summarized in \autoref{fig:darpa_timeline}, with \revv{substantial developments over each of the 6 months intervals between tests by all teams, including (but not limited to) \cite{cerberusJFR, coordRobJFR, csiroJFR,ctuJFR,explorerJFR, marbleJFR,nctuJFR}.
} 

\subsubsection{SubT Integration Exercise (STIX)}
\begin{figure}[!t]
    \centering
    \includegraphics[width=0.8\textwidth]{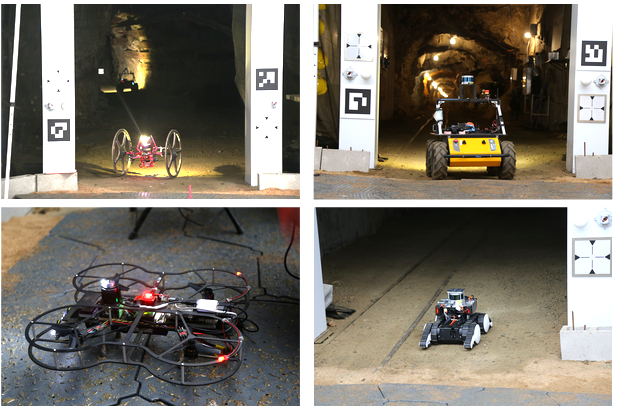}
    \caption{Four heterogeneous mobility platforms deployed at the STIX event. Clockwise from top left:: Rollocoper (with Husky in the course), Husky, Flipper and Scout.}
    \label{fig:stix_robots}
\end{figure}

\ph{Environment and Robots} STIX was held in April 2019 at the Edgar Experimental Mine, Idaho Springs, CO. The participating teams were offered two practice sessions and one simulated scored run in an environment representative of the Tunnel Circuit. For Team CoSTAR, one of the primary objectives was to quantify the capability and limitation of various mobility systems through the evaluation in a real mine environment. To that end, we deployed four different types of mobility platforms shown in \autoref{fig:stix_robots}, including Clearpath Husky wheeled vehicle, Flipper tracked vehicle, Scout quadcopter, and Rollocopter aerial/ground hybrid vehicle. 

\ph{Performance} Each platform demonstrated autonomous exploration capabilities in the mine. Husky autonomously traversed 200 m until it reached the first fork. During the drive, Husky detected 5 artifacts and reported back to the base station: two of which were correct in types and locations. Soon after Husky turned into a side passage, the communication to the base station was lost, which triggered the return to base behavior as designed. The robot successfully returned to the base by following the breadcrumb nodes in IRM.  Rollocopter performed four 3-minute runs during the practice sessions, each of which consisted of autonomous take-off, exploration, and landing operations. The robot exhibited robust navigation capabilities, traveling over 100m in cumulative distance.  Dust was a major issue for drone flight in the narrow passageways, causing vision-based state estimation failures.  Our approach of using heterogenous odometry sources enabled us to be resilient to these failures to a large degree \cite{hero2019isrr}.  Dust largely obscured all cameras after a few minutes into each run.  With no measurement updates from odometry sources, we relied on IMU-only inertial odometry to safely land the vehicle.  These results led to future work on improving state estimation resilience in the presence of dust and variable lighting. It also helped us to improve the camera placement design for the various scales of environments. \rev{The Flipper was able to navigate over the train tracks in the tunnel because of its tracked design but it was much slower than the Husky. This is because of greater contact area of the tracks that causes slower turning behavior compared to differential drive (wheeled) robots for the same commanded track/wheel speeds~\cite{dixit2020kinematics}. This is why we decided to incorporate a hybrid vehicle that had both wheels and tracks (Telemax) in the next circuit. We extensively tested Telemax in the Arch mine~\cite{thakkur2020} and were able to navigate over different types of terrain (in tracked mode) while maintaining speed when the robot was on flat terrain (in wheel mode).
}

\subsubsection{Tunnel Circuit}
\ph{Environment} The Tunnel Circuit took place in August 2019 at the NIOSH mine complexes in Pittsburgh, PA. There were two courses, Experimental (EX) and Safety Research (SR), focusing on different aspects of the tunnel environmental challenges. The EX course is composed of long straight corridors with featureless flat walls. The SR course has a grid-like structure that provides many decision points and loop closures.

\begin{figure}[!t]
    \centering
    \includegraphics[width=0.8\textwidth]{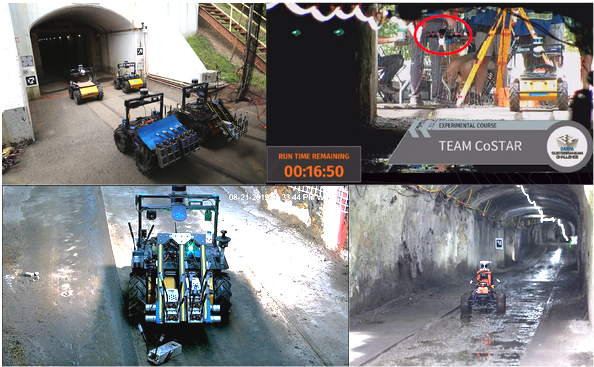}
    \caption{Robot team deployed at the Tunnel Circuit event. Top Left: Four Huskies at the Staging Area, Top Right: Drone deployed at the Experimental Mine, Bottom Left: Husky deploying a communication node, Bottom Right: RC car deployed at the Experimental Mine.}
    \label{fig:tunnel_robots}
\end{figure}

\ph{Robots} Team CoSTAR staged seven robots at the starting gate (see \autoref{fig:tunnel_robots}). The robot roster includes four Clearpath Husky robots (Husky 1-4), one Telerob Telemax track/wheel hybrid robot (Telemax 1), one high-speed RC car (Xmaxx 1), and one aerial drone (Scout 1). Our strategy was to adaptively deploy the heterogeneous robot team based on the complexity and challenges of the unseen course elements. The information from vanguard robots told that the environment is benign and mostly accessible by the ground vehicles, leading the decision to rely more on wheeled platforms that show higher endurance.

\begin{figure}[!bt]
    \centering
    \includegraphics[width=0.6\textwidth]{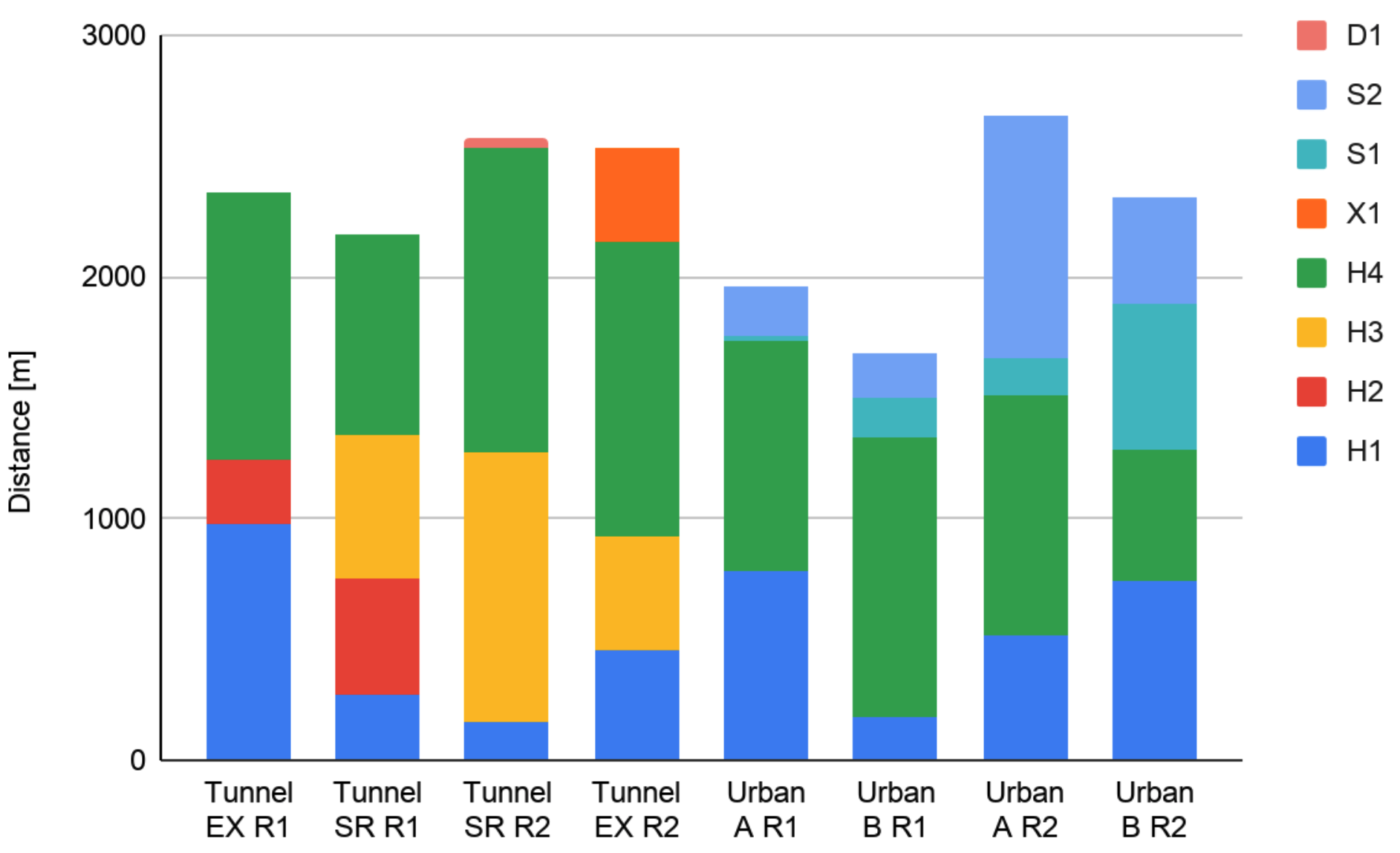}
    \caption{Driving distance statistics during the Tunnel and Urban Circuits. Prefixes refer to different robot types: D - Drone, H - Husky (wheeled robot), S - Spot (legged robot), X - XMax (1/5 scale RC car).}
    \label{fig:tunnel_urban_distances}
\end{figure}

\begin{figure}[h!]
    \centering
    \includegraphics[width=0.6\textwidth]{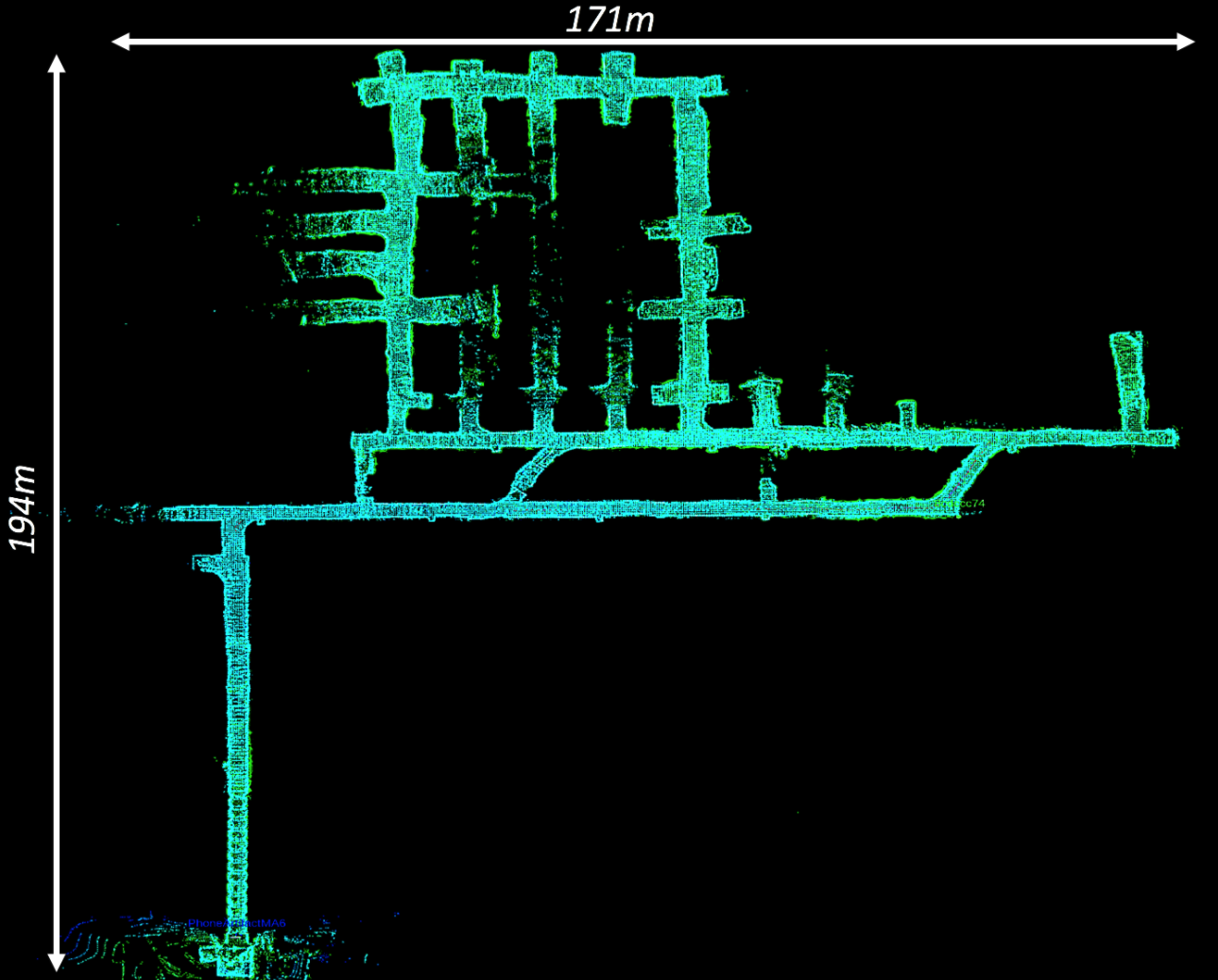}
    \caption{SLAM point cloud map from a single Husky robot run in the Safety Research run of the Tunnel Circuit. The robot starts from the bottom left, and completes the top loop in a clockwise direction.}
    \label{fig:sr2_map}
\end{figure}

\ph{Performance} In each run, we sent 3 to 4 robots to the course and achieved more than 2 km combined traversal (\autoref{fig:tunnel_urban_distances}). The video in ~\cite{youtubeurbanandtunnelvideo} depicts some highlights of these runs. The longest single-robot drive was 1.26 km by Husky 4 at the SR mine on Day 3, including long periods of no-communication autonomous exploration and successful returning to the communication range at the end of the mission. \autoref{fig:sr2_map} shows the map created by Husky 4 in this run, with an error under 1\% of the distance travelled. Four communication nodes are deployed from the robots to build a backbone network, covering the areas near the mine entrance and extend the reach of base station for faster data retrieval. \revv{We detected 16 artifacts during our four runs of the Tunnel Circuit, leading to a circuit score of 11 and the second-place finish among 11 teams (\autoref{table:competition_scores}).}

\begin{table}[]
\centering
\caption{\revv{Number of scored points for each competition run.}}
\label{table:competition_scores}
\begin{tabular}{|c||c|c|c|c|c|c|c|c|}
\hline 
Circuit & \multicolumn{4}{c|}{Tunnel}         & \multicolumn{4}{c|}{Urban}        \\ \hline
\hline
Run         & SR R1 & SR R2        & EX R1 & EX R2        & A R1 & A R2         & B R1 & B R2         \\ \hline
Run scores  & 4   & \textbf{7} & 2   & \textbf{4} & 5  & \textbf{7} & 4  & \textbf{9} \\ \hline
Course scores & \multicolumn{2}{c|}{7}      & \multicolumn{2}{c|}{4}  & \multicolumn{2}{c|}{7}         & \multicolumn{2}{c|}{9}           \\ \hline
Circuit scores & \multicolumn{4}{c|}{11}             & \multicolumn{4}{c|}{16}           \\ \hline
Circuit rank & \multicolumn{4}{c|}{2nd}             & \multicolumn{4}{c|}{1st}           \\ \hline
\end{tabular}
\end{table}

\ph{Challenges} During the Tunnel Circuit event, we faced many real-world challenges which contributed to improve our system toward the next Circuit events. The flat featureless walls of the EX course affected our localization performance which was purely LiDAR-based as of the Tunnel Circuit. This motivated the development of multi-modal method (\autoref{sec:state_estimation} and \autoref{sec:lamp}). Multi-robot operations in a comm-degraded environment also posed challenges to our networking system. Based on the analysis of the Tunnel experience, we carefully redesigned the inter-robot networking protocol and deployed it in the Urban Circuit (\autoref{sec:multirobot_networking}). The drone traveled 35 m in 43 seconds before it experienced critical state estimation error due to poor lighting. To eliminate this single-point failure in the future flights, we put more efforts on multi-modal sensing and parallel estimation (\autoref{sec:state_estimation}).

\subsubsection{Urban Circuit}
\ph{Environment} The Urban Circuit took place in February 2020 at the Satsop Business Park in Elma, WA. The unfinished power plant in the park was chosen for the place of the second circuit event where the robots were exposed to challenges from urban structures. The two courses, Alpha and Beta, covers two floors of the power plant with size around $90 \times 90 \times 15$ m$^3$. There are many small rooms and narrow corridors divided by thick walls that prevent direct wireless communications. 

\begin{figure}[!hbt]
    \centering
    \includegraphics[width=0.95\textwidth]{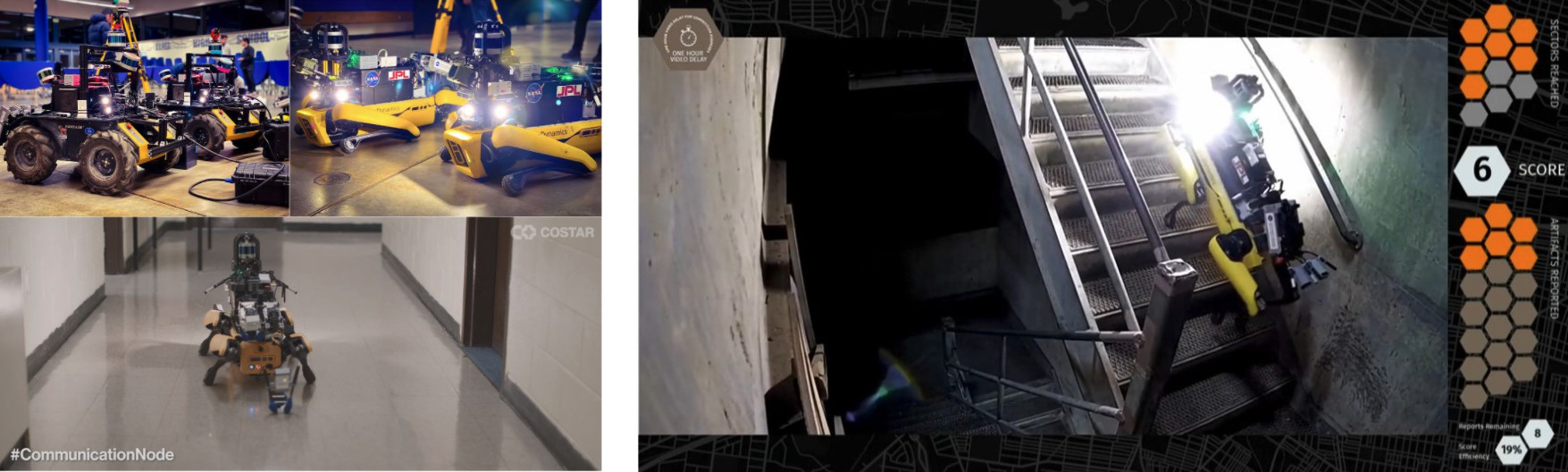}
    \caption{Robot team deployed at the Urban Circuit event. Left: Robot fleet and communication node deployment, Right: A snapshot of Spot stair climbing (DARPA TV).}
    \label{fig:urban_robots}
\end{figure}

\ph{Robots} Team CoSTAR staged seven robots and deployed four of them during this event (\autoref{fig:urban_robots}), including the newly introduced Boston Dynamics Spot quadruped robots (Spot 1 and 2). The other two robots are Clearpath Husky (Husky 1 and 4) with major upgrades in onboard electronics and sensor stack. All robots are deployed in each run, acting in the vanguard and supporting roles based on the assignment, their location, and time of the mission. One Spot was dedicated to climbing stairs to explore the floor at a different level.

\begin{figure}[!hbt]
    \centering
    \includegraphics[width=0.8\textwidth]{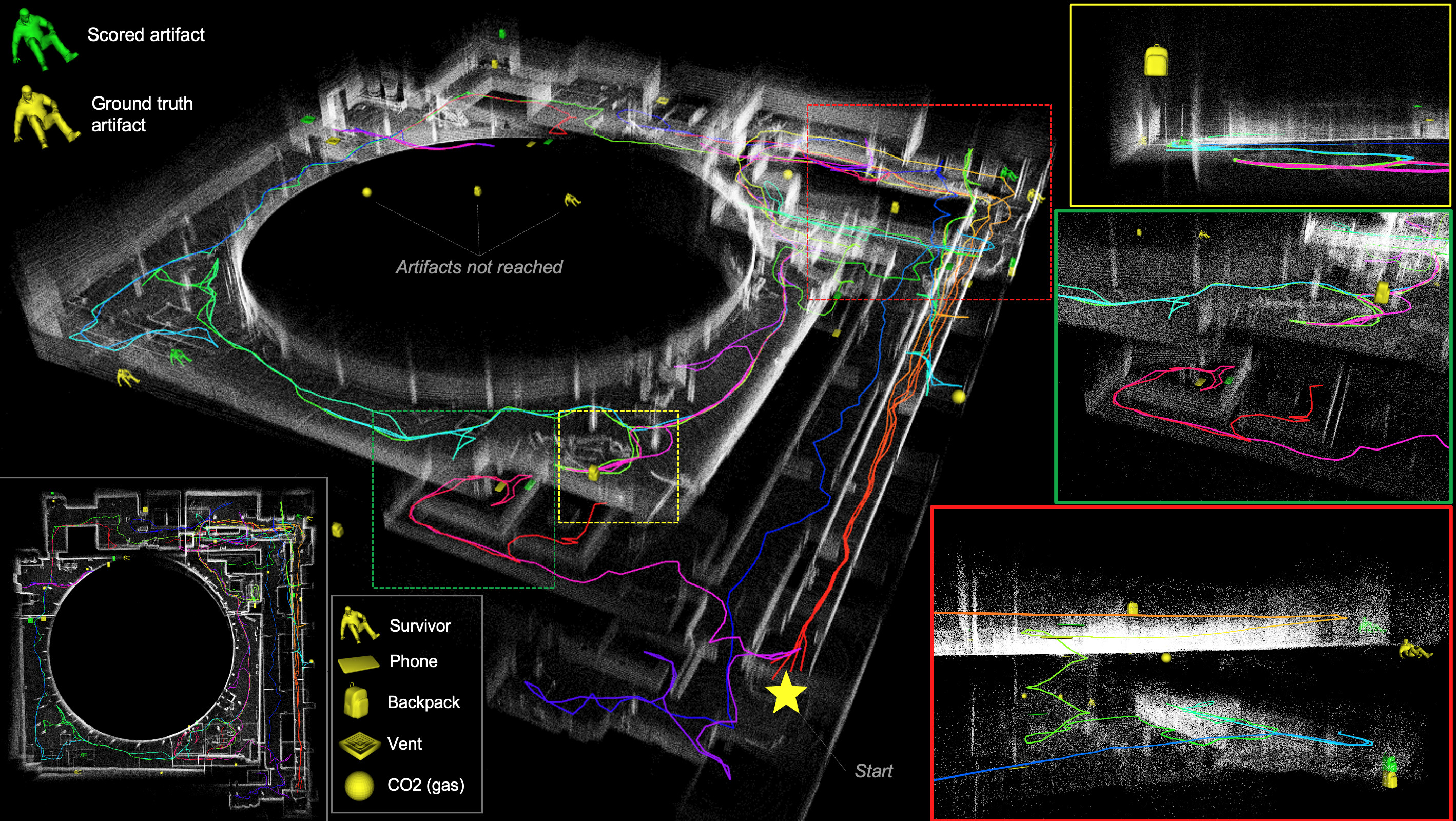}
    \caption{Multi-robot map, trajectories, scored artifacts and ground truth artifacts. All robots start from the star on the bottom right with trajectory colors cycling from red through yellow to blue and back to red based on distance travelled. RED inset shows multi-level exploration and scoring (upper-level survivor and lower-level backpack) by Spot1 with a stair descent (green part of trajectory). GREEN inset shows the furthest extent of lower-level exploration where a cell phone was scored by Spot1. YELLOW inset shows an instance of an unscored artifact, where it is placed out of view of ground robots, requiring flight. Other unscored artifacts are seen in the center of the map where no robot arrived near to them.}
    \label{fig:urban_demo}
\end{figure}

\ph{Performance} \revv{We detected 25 artifacts during our four runs of the Urban Circuit, leading to a circuit score of 16 and the first-place finish among 10 teams (\autoref{table:competition_scores}).} \autoref{fig:urban_demo} shows a multi-robot map generated at the Beta course, including details on scored artifacts at multiple levels and at the furthest extent of exploration. The four-robot team traversed a combined total of 2.3 km (\autoref{fig:tunnel_urban_distances}), including a large closed loop by the two husky robots around the central round structure. \revv{The video in~\cite{youtubespotpaper} depicts Spot robots exploring the multi-level courses autonomously.}

\begin{figure}[!hbt]
    \centering
    \includegraphics[width=0.7\textwidth]{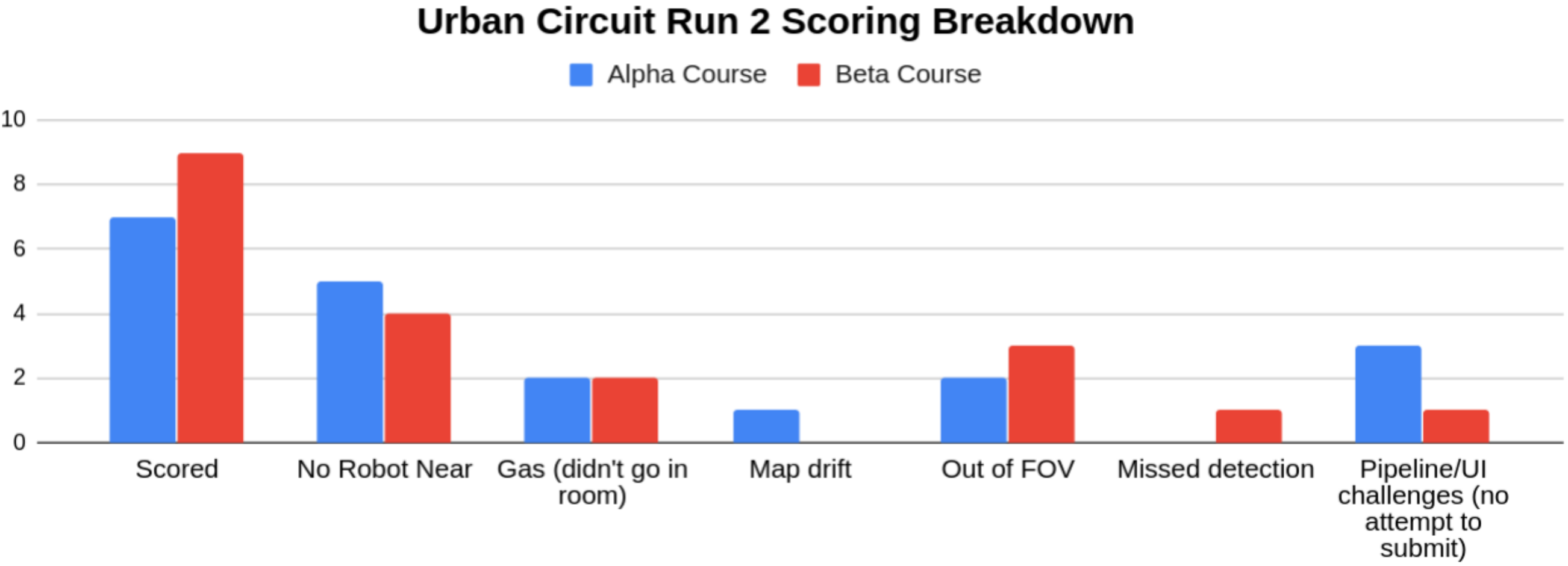}
    \caption{Scoring breakdown for the two highest-scoring Urban Circuit runs.}
    \label{fig:urban_artifacts}
\end{figure}

\autoref{fig:urban_artifacts} shows the breakdown of scoring in the two highest scoring runs of the Urban competition, summarizing the reasons for not scoring the remaining artifacts. In addition to the artifacts missed due to the robot not reaching the artifact locations (\autoref{fig:urban_demo}), we missed artifacts due to the limited field of view of the sensory suite on the robots (\autoref{fig:urban_demo}) and other challenges in the artifact detection pipeline. 

\ph{Challenges} Overall, the Team CoSTAR's system showed significant improvement from the Tunnel Circuit. Nonetheless, the circuit identified numerous ongoing challenges calling for further development. These development areas include: improved depth and breadth of coverage, enhanced sensor field-of-views for artifact detection, aerial robot developments, autonomous fault recovery, and careful attention to comm node placement/configuration to maximize wireless link quality for fast data retrieval.

\subsection{Self-Organized Cave Circuit} \label{ssec:cave_circuit}
\ph{Environment overview} We self-organized a Cave Circuit event at the Valentine Cave of the Lava Beds National Monument, Tulelake, CA. The cave, a class of lava tubes, was formed from volcanic flows, and an active research target for the future lunar/Martian cave exploration~\cite{blank2018planetary,cavesWhitepaper}. The length of the cave is approximately 300 m with an elevation change of 20 m. The tortuous and steep terrain in the cave limited the range of operations of wheeled platforms, hence tests were focused on the Spot quadrupeds (drone testing was not possible due to regulatory challenges).

\ph{Circuit Event Organization} The test was run as close to the competition setup as possible. The global frame and artifact locations were determine based on a pre-existing, high precision Faro scan of the cave. A fiducial gate was surveyed into the cave frame and used to calibrate the robots. All artifact classes other than CO2 and Vent were placed throughout the cave, with a prioritization on ropes and helmets (the cave-specific artifacts). 

\begin{figure}[!hbt]
    \centering
    \includegraphics[width=0.8\textwidth]{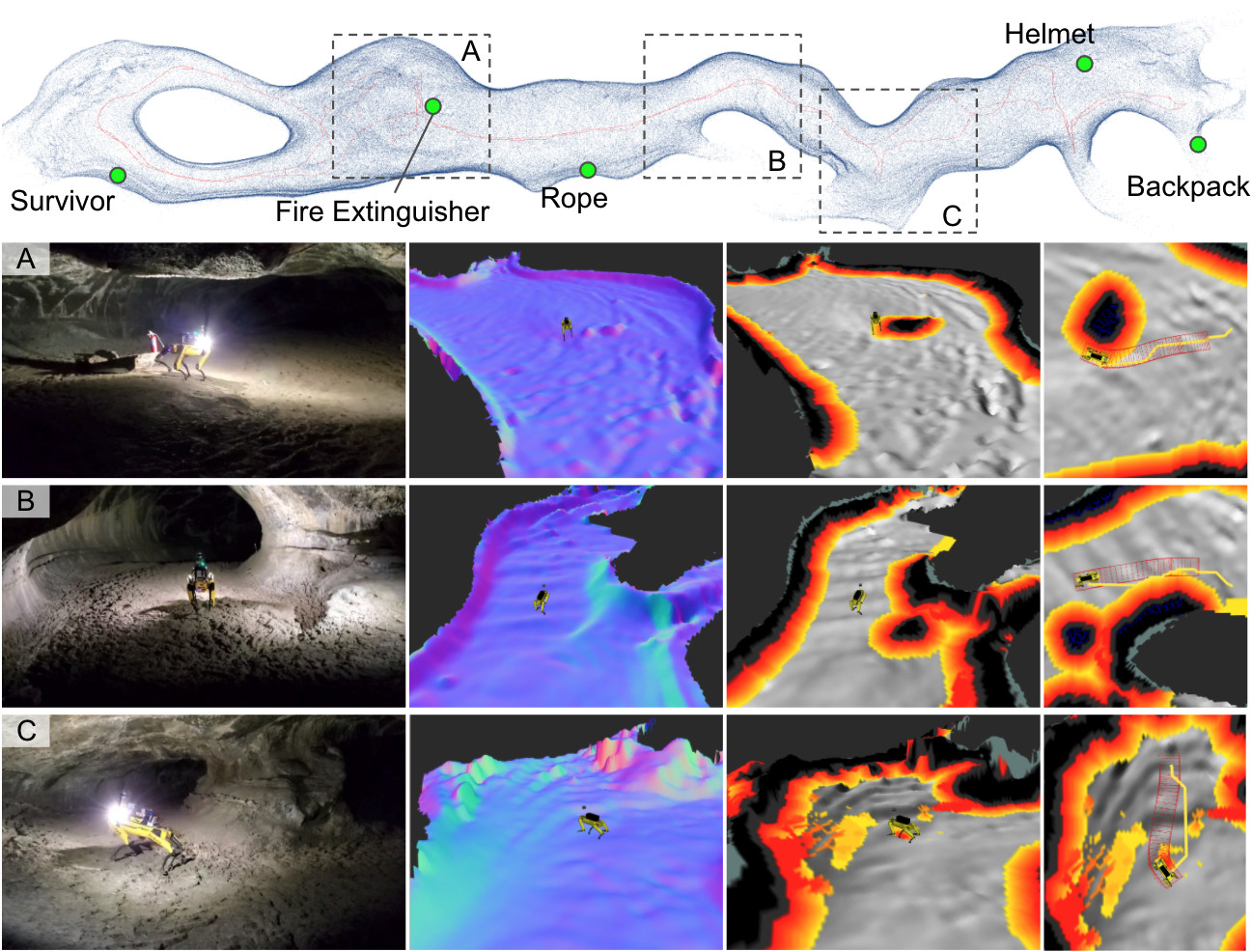}
    \caption{A pointcloud map from the SLAM module and traversability maps during autonomous exploration in the Valentine Cave at Lava Beds National Monument, Tulelake, CA.}
    \label{fig:cave_demo}
\end{figure}

\ph{Results} The cave demonstration was remarkable in that entire runs were fully-autonomous with zero human intervention: the operator only started the mission. \autoref{fig:cave_demo} shows a map from one such run. Spot traversed 400 m on average over 4 runs. Each mission was ended not by a time limit, but by the full accessible environment being covered, with side passages limited by traversability hazards (very low ceilings, cliffs). \revv{The videos of the cave demonstration are available at ~\cite{youtubeAutonomousCaveExploration,youtubeBDLavaTube, youtubeMarsDog}.}

\rev{\section{Lessons Learned}} \label{sec:lesssons_learned}
At the time of submission, NeBula's uncertainty-aware and perception-aware principles are implemented in a number of modules throughout our autonomy solution. The module interactions have been tightly co-designed and joint probability distributions across some of the modules are incorporated to bolster the overall system’s resiliency. This uncertainty-aware architecture is verified through intensive field testing campaigns using heterogeneous platforms with different variations of mobility, sensing, and computing capabilities. We observed that the joint perception-planning approach brought resiliency to the system behaviors in challenging real-world environments, where uncertainty is ubiquitous. This section summarizes the lessons learned from developing and fielding the NeBula autonomy solution.

\subsection{Heterogeneous system design and integration}
Designing a system with heterogeneous robots is a non-trivial task. Accomplishing the objectives of complex real-world operations (such as the one in the SubT) with constraints on time, resources/cost, robot size, weight, power, etc., can be too difficult or impossible for a single robot. This calls for multi-robot solutions. An important lesson learned and a major future direction is: “It is critical to have principled tools that can translate the design choices to overall mission metrics”. These design choices range from the selection of type and number of robots to the selection of sensors, instruments, and algorithms, to the type of wires and connectors, and batteries. As some concrete examples that our team has encountered: “A slight change in the weight (and hence operation time) of one single robot” or “selection of a specific wire that can slightly affect the data traffic”, can have a significant effect on the overall aggregated performance of the robot team and mission metrics. There are 100s if not thousands of such design parameters and choices (both in HW and SW), which calls for systematic tools to quickly abstract them and translate them to high-level mission metrics. 

An important observation is that for the fast-paced development of a very large autonomy architecture for a multi-robot system, the architecture should be highly modular and adaptive. Since the architecture must incorporate robots' heterogeneous capabilities (mobility, sensing, and compute), appropriate abstractions need to be implemented to operate robots in a unified framework. We have performed many iterations of the architecture redesign as our entire system capabilities grow. A key lesson learned is that integration testing at a regular interval is essential to maintain the stability of such a growing system. Team CoSTAR has organized monthly virtual and physical integration demonstrations with competition-like set-ups. Regular testing enforces the new module to comply with the system interaction rules, ensuring functionality despite constant augmentations by multiple subteams throughout the development. Regular performance tracking also allows us to understand the current technology state and highlights the next domain to be studied.

\subsection{Resilient state estimation}
In challenging real-world robotic operations, perceptually-degradation is common, and state estimation can suffer from deterioration of sensor measurements and estimation quality. In addition, physical systems are exposed to the risk of sensor failures that are made prevalent by frequent field testing. These failures are difficult to model in many cases, yet systems must be able to react to these uncertain events to ensure system stability is not lost. One key lesson from our experience is to let the system predict a failure and quickly adapt to it. We observed that adding redundancy (e.g., HeRO at Section 5) in a principled way shows a lot of potentials to improve the resiliency of an odometry system by incorporating multi-modal sensors and algorithms. HeRO’s built-in health check mechanism allows for preliminary detection of various types of failures and adapting the system to mitigate the effect. This proactive approach has been effective in our field tests and circuit events, providing the foundation to support high-level autonomous behaviors.

Another critical observation is on “how and when to trust the sensor models”. Following NeBula’s philosophy of uncertainty-aware predictive planning, the planner needs to be able to predict the joint distributions over the system’s future actions and estimated states. To predict the future performance, we divide unknowns into “known unknowns” and “unknown unknowns”. Known unknowns refer to uncertainties that can be reasonably modeled using probability distribution functions. Unknown unknowns refer to unmodeled uncertainties resulting from unexpected events in the system operation. To be resilient and robust to these uncertainties, we have observed that a cascaded framework (like HeRO) of loosely-coupled and tightly-coupled layers is promising. The cascaded framework first copes with unknown unknowns, by relying on a loosely-coupled layer and checking the consensus across different sensory modalities to detect anomalies and reject faulty estimation channels. Once left with inliers, it handles “known unknowns” by tightly fusing the estimation channels using their predicted probabilistic models.

\subsection{Large-scale positioning and mapping}
The SLAM design and deployment efforts presented in this paper highlight the importance of using complementary sensor information for localization and mapping. In complex environments, various sensors have their own advantage and disadvantages. For example, wheel-encoder-based odometry or visual odometry provides a useful source of information in long corridors where LiDAR scan matching is subject to drift; IMU data allows for accurate short-term relative rotation estimation while history-based scan matching can resolve longer-term displacement estimation. This need to fuse heterogeneous sensors also emphasizes the importance of a flexible SLAM framework where one can easily define and fuse sensor data when available. In this sense, the use of a factor graph framework 
reduced the integration efforts and allowed us to design a unified SLAM back-end that can be easily reconfigured. The importance of developing a reconfigurable framework is further exacerbated by the desire of running our SLAM solution on a heterogeneous set of platforms (e.g., robots with different sensor suites and different computational capabilities), which demanded our framework to be sensor agnostic (e.g., adjust to different sensors with minimal parameter tuning) and reconfigurable (e.g., enable and disable sensors to fuse via configuration files).

While having a reliable odometry solution is critical in large-scale mapping, even the most accurate odometry systems accumulate error over long distances across extreme terrains. This remarks the importance of loop closures to keep the localization errors bounded. However, when it comes to complex, large-scale environments, with perceptually-degraded conditions, a key lesson learned is that loop closure needs to be incorporated in a resilient manner into the overall framework. Computational constraints will limit the ability to search the map history and find correct loop closures in a reasonable time. Even after loop closures are found, it is crucial to maintain multiple hypotheses or at least remove incorrect loop closures resulting from perceptual aliasing. In this paper, we observed that \textit{graph-based outlier rejection} (e.g., a variation of \cite{mangelson2018pairwise}) is helpful in filtering some of the incorrect hypotheses. This family of methods rely on approximate max clique solvers for consistency maximization, which becomes computationally expensive due to the size and density of the SLAM graphs in competition settings. Unfortunately, they often fail to select inlier loop closures, inducing “jumps” in the robot trajectory when the solver is stuck in suboptimal maximal cliques. Future work includes improving loop closure detection (e.g., prioritizing the most informative loop closures, using different sensors for place recognition) and adopting more recent methods for outlier rejection based on graduated non-convexity.

Computational aspects can be also improved by sharing the workload across robots. Centralized multi-robot SLAM methods require increasing computational resources for larger teams of robots. Adopting a distributed SLAM system
can improve scalability and reduce communication bandwidth. In general, distributing the overall computation across various robots (based on their capabilities) can better scale to large teams of robots. 

An important lesson learned is: Augmenting geometric information with semantics can increase the resiliency of the SLAM system. For instance, incorporating semantic information such as intersections, stairs, man-made objects in the graph can increase the robustness of the loop closures. Further, identifying a set of stairs in the environment provides a readily usable prior on the geometry of the stairs; similarly, identifying doors provides actionable information for navigation. Metric-semantic mapping is an active research area and a tight integration of geometric, semantic (and physical) reasoning has the potential to improve the robustness and accuracy of the map built by the robots.

Finally, while this section is concerned with localizing the robots and building a map, the quality of the map reconstruction is highly affected by the trajectories covered by the robots. Following NeBula’s uncertainty-aware planning perspective, performing active loop closures to reduce the uncertainty in the robot location has a major impact on the SLAM accuracy. Active loop closures can be enabled by guiding robots to rendezvous points to create inter-robot measurements or by visiting parts of the environment that have unambiguous detection signatures. Towards this goal, quantifying and actively reducing uncertainty is crucial and is a fundamental trait of the NeBula framework. Active localization and mapping go beyond loop closures, and tightly couples perception, action, and communication.

Uncertainty-aware traversability analysis: 
Navigating over perceptually-degraded challenging terrains requires a radical departure from systems designed to operate in known environments with clear landmarks and easily distinguishable obstacles. Designed to detect complex geometric hazards at various scales, traversability analysis highly depends on the quality of local mapping and state estimation. As such, an important lesson has been that accounting for the “perception uncertainty” in planning is a key to building a reliable traversability system operating on challenging terrains and perceptual conditions. We learned that multi-fidelity mapping approaches (similar to the one presented in Section 8) improve the balance between computational constraints and accuracy in the presence of degraded state estimation and sensor measurement uncertainty. Uncertainty-aware estimation trusts and accumulates measurements in the world belief based on measurements’ accuracy and quality. Using this world belief, the traversability layer precisely quantifies the perception-aware traversability risks and costs when negotiating challenging terrains. Finally, a critical lesson in traversability algorithm design and development is constant field testing: The presented uncertainty-aware approach has been field-hardened in over 100 field test sites with diverse traversability-stressing elements.

\subsection{Scalable belief-space global planning}
Global planning for area coverage and exploration behaviors is one of the modules where the awareness of uncertainty plays a critical role in achieving high performance in the real world. We remark several key dilemmas encountered while developing and testing the global planner, related to uncertainty representation and uncertainty-aware decision making: i) scalability vs. information fidelity in world belief representation, ii) planning horizon vs. planning time for each receding-horizon planning episode, and iii) the consistency of plans for smooth motions vs. resiliency to sudden changes in world belief over time. We learned that hierarchical approaches (similar to the ones described in Section 9) have the potential to address the above challenges. Our method, PLGRIM, leverages i) hierarchical IRMs, ii) longer-horizon, higher resolution POMDP solvers with manageable computation load, and iii) receding horizon planning with resiliency logic. However, those are still open problems. An important future direction is “quick online adaptation” of parameters that balance the environment scales and complexity with the computing system and sensor limitations. Another important open problem is related to encoding and capturing a more accurate and reliable representation of the high-dimensional world belief.

\subsection{Semantic understanding and artifact detection}
Detecting objects and understanding the semantics of the environment in perceptually-degraded subterranean settings is a challenging task. In the DARPA subterranean challenge, artifacts have multiple signatures, ranging from visual, thermal, auditory to gas-based, and radio-based artifacts. These various signatures and payload constraints (size, weight, power, etc.) on different heterogeneous platforms call for a method that assesses the value of adding various sensors and sensing modalities to each vehicle. Further, the choice and configuration of each sensor on each robot have a significant impact on the scoring performance (see Section \ref{sec:artifacts}). For example, for artifacts that can be detected visually, the salient camera parameters include field of view, reliability, image quality, resolution, and frame rate. Hence an important lesson has been that the artifact collection system needs a tight co-design of software and hardware. This includes the whole pipeline from the sensors, cables, to the processors and algorithms. For example, a change in a USB hub or cable can significantly impact the choices of other hardware and algorithms by removing certain data chocking elements in the pipeline. A related unexpected failure mode (which was occurred in one of our competitions) at the system level was sensor data transfer and communication, caused by both USB driver and networking bandwidth limitations. Correcting this problem required a thorough analysis of the hardware configuration, including the development and testing of custom driver software by our team.

The second major lesson was related to data. While the existing standard object detection datasets are quite large and diverse, learning to detect objects in the perceptually-degraded environments (e.g., variable lighting and with obscurants), such as the ones in the DARPA subterranean challenge-like environments, are still out-of-distribution in relation to mainstream datasets.

Finally after these trade-offs, there are still many limitations in the perception side of the artifact detection architecture. So the third, and the most important, lesson learned is that the planning and perception for artifact detection need to be tightly co-designed. In other words, active perception is required where the planner needs to take actions that compensate for perception shortcomings. Examples of such actions are: (i) executing local search maneuvers where the robot sweeps the larger parts of the scene with it sensors to compensate for its limited field of view, (ii) actively getting closer to certain targets or making measurements from various angles to increase the detection confidence. This mode might also include deciding which robot and which sensor should gather more information to increase the confidence. It also includes (iii) changing perception pipeline parameters such as the camera resolution, input rate, etc. to  handle computational constraints by focusing and limiting the attention of the system on the important parts of the input channel. Active perception for semantic understanding is highly open area, with a lot of future work in these domains. 

\subsection{Bandwidth-aware communication system design}
Communications between computational units (e.g. ROS nodes) should ideally take into account the predicted available bandwidth in the link whether that link is on the same computer, via a high-speed Ethernet, a larger wired network, or radio/wireless. We took a number of successful steps in this direction: i) separating inter-agent (ROS 2) and intra-agent (ROS 1) communications, ii) using different QoS settings for different classes of topics, iii) monitoring inter-robot communications and estimated bandwidth, and iv) monitoring intra-robot communications. As future multiagent projects move away from ROS 1 and towards ROS 2 (or another middleware/communication solution) the same principles apply. We see that there is a need for continued improvements and autonomy in routing and QoS systems to optimally use communication resources in these future systems.

Another important lesson learned is that due to the communication-degraded nature of the subterranean environments, the planner needs to support and improve the communication and multi-robot networking performance by dropping communication nodes at the optimal locations and by actively carrying (mulling) data using mobile robots between the various nodes. Hence, aligned with NeBula’s philosophy, (1) taking networking uncertainties into account, (2) evaluating and predicting the potential value of the communicated information, and (3) co-designing the planning system and networking system are that critical observations to increase the robustness and performance of the overall multi-robot robot networking system in communication-degraded environments.  

\subsection{Supervised autonomy to full autonomy}
Achieving mission success depends heavily on the mission-level autonomy from both single and multi-robot perspectives. This is especially the case in contexts such as the SubT Challenge where large robot teams must intelligently coordinate themselves under strict communication, time, and resource constraints to explore and map km-scale environments and find objects of interest.

Closer to the course entrance a communication link with the human supervisor can be established. Thus the system must be capable of autonomously balancing between deeper exploration, operational risk, the value of the collected onboard information, and the cost of bringing the systems closer to the surface for the data retrieval. 
A key lesson learned is that even when robots are close to the course entrance and a communication link with the human supervisor is established, with a large heterogeneous multi-robot team, a single human supervisor becomes a bottleneck in the control loop due to excessive cognitive workloads of the supervisor. Hence, for successful operations, NeBula's mission executive and Copilot assistant are designed to support such tasks by adjusting the system autonomy levels, gradually delegating human tasks where the human is viewed as a resource, and providing suitable instructions and feedback to reduce the cognitive load of the single supervisor. Integral to the effort is maintaining a world belief that can be easily interpreted by the system or supervisor, and trigger behaviors or sequences of actions that bring the system closer towards achieving mission completion. 

However, there is a lot of future research in this area. Our ongoing efforts have been focusing on some of these areas, including efficient ways of specifying more complex tasks and missions, balancing human-machine task distribution, planning using semantic information, and scheduling under uncertainty in task execution and future event occurrence. 

\subsection{Simulator-based development} 
The development of our system was greatly accelerated by the use of different computer simulation environments, discussed herein. The challenges inherent to real-world robotics in uncertain and fully autonomous settings pose tremendous risks to naive robots and algorithms in the early stages of development. Taking development into a simulation and out of hardware has afforded us numerous opportunities to pursue and achieve high-risk-high-gain algorithmic strategies. Many safety-critical features are difficult to test with physical hardware. By carefully modeling different fleet and hardware configurations, we were able to develop, test, and eventually deploy our heterogeneous multi-robot team in the real world, with optimal networking, hardware, and software payloads. Although simulators offer an enormous benefit to robotic systems development, there are significant differences between simulations and real-world system which must be accounted for. It is important to understand simulator limitations and always perform proper validation and testing on physical hardware platforms.

\section{Conclusion} \label{sec:discussion}
Team CoSTAR's approach to the DARPA Subterranean Challenge lies in our autonomy solution, NeBula (Networked Belief-aware Perceptual Autonomy), which emphasizes resilience and intelligent decision making through uncertainty-awareness. NeBula has led to 2nd and 1st place in DARPA Subterranean Challenge's Tunnel (in 2019) and Urban (in 2020) competitions, respectively. When dealing with exploration and operation in unknown environments, uncertainty is inherent to all decisions. As the main principle, NeBula focuses on quantifying and taking advantage of uncertainty at multiple levels of our autonomy stack, including state estimation, mapping, traversability, planning, communications, and other state domains.  We combine these technologies in a synergistic way, which examines the interaction between interrelated modules.  

During this competition we have and hope to continue to demonstrate autonomous exploration in extreme environments on multiple platforms (including wheeled, legged, and aerial). Solving this problem remains of paramount interest in a wide range of applications, especially when it comes to missions to explore unknown planetary bodies beyond our home planet. We believe the uncertainty-aware and platform-agnostic nature of most NeBula components is a step towards resilient and safe robotic autonomy solutions in unknown and extreme environments with both single and multi-robot systems.

\subsubsection*{Acknowledgments}
The work is partially supported by the Jet Propulsion Laboratory, California Institute of Technology, under a contract with the National Aeronautics and Space Administration (80NM0018D0004), and Defense Advanced Research Projects Agency (DARPA).\footnote{© 2021. All rights reserved.}

We also would like to thank our collaborators and sponsors including Boston Dynamics, JPL office of strategic investments, Intel, Clearpath Robotics, Velodyne, Telerob USA, Markforged, Silvus Technologies, ARCS California State University Northridge, Vectornav, HM Engineering, ARCH Mine, Mine Safety and Health Administration (MSHA), National Institute for Occupational Safety and Health, Polytechnique Montreal, and West Virginia University. 

We would like to thank the rest of Team CoSTAR, our collaborators, and advisors for their support, fruitful discussions and their contributions to the project. In particular, we would like to thank Giulio Autelitano, Nicholas Ohi, Mamoru Sobue, Yuki Kubo, Kenny Chen, Jennifer Blank, Kevin Lu, Sahand Sabet, Sandro Berchier, Yasin Almalioglu, Micah Corah, Meriem Ben Miled, Joseph Bowkett, Navid Nasiri, Tomoki Emmei, Rianna M Jitosho, Gita Temelkova, Dave Gallagher, Leon Alkalai, Richard Volpe, Lorraine Fesq, Robert Stirbl, Randy Odle, Cagri Kilic, Leonardo Forero, Daniel Tikhomirov, Juan Nieto, Kyle Strickland, Chris Denniston, Antoni Rosinol, Alexander Nikitas Dimopoulos, Harrison Delecki, Nicolas Marchal, Oriana Claudia Peltzer, Joshua Ott, Nicholas Palomo, Adam Jaroh, Heiko Helble, Aaron Ames, Issa Nesnas, Fernando Mier-hicks, Shreyansh Daftry, Nicholas Peng, Mitra Azar, Corey Mack, Darmindra Arumugam, Jack Bush, Terry Suh, Mike Paton, Jared Beard, Jennifer Nguyen, Evangelos Theodorou, Michelle Tan, Andrew Haeffner, Pradyumna Vyshnav, Marcus Abate, Alexandra Stutt, Kartik Patath, Arghya Chatterjee, Shehryar Kattak, Chris Heckman, Kostas Alexis, Zack Jackowski, Michael Perry, Marco da Silva, Nhut Tan Ho, Barry Ridge, Brett Kennedy, Larry Matthies, Richard French, Jeffery Hall, Andila Wijekulasuriya, Paul Backes, Marco Tempest, Joe Bautista, Kayla Mesina, Erica Bettencourt, Amiel Gitai Hartman, Jeff Delaune, Daniel Pastor, Yu Gu, Jason Gross, Jacopo Villa, Soojean Han, David Chan, Rashied Amini, Jake Ketchum, Andrew Bieler, Slawek Kurdziel, Steve Zhao, Reza Radmanesh, Jose Mendez, Roberto Mendez, Christopher Patterson, Jack Dunkle, Jong Tai Jang, Petter Nilsson, Filip Claesson, Emil Fresk, Ransalu Senanayake, Mohammadjavad Khojasteh, Andrew Singletary, Alexandra Bodrova, and Tesla Wells.

The content is not endorsed by and does not necessarily reflect the position or policy of the government or our sponsors.

\section{Glossary: Acronyms} \label{sec:acronyms}
\begin{acronym}
\setlength{\parskip}{0ex}
\setlength{\itemsep}{0ex}
\acro{APE}{Absolute Position Error}
\acro{AUV}{Autonomous Underwater Vehicle}
\acro{BPMN}{Business Modelling Process Notation}
\acro{BRM}{belief roadmap}
\acro{CHORD}{Collaborative High-bandwidth Operations with Radio Droppables}
\acro{CIO}{Contact-Inertial Odometry}
\acro{Comm}{Communication}
\acro{ConOps}{Concept of Operations}
\acro{CoSTAR}{Collaborative SubTerranean Autonomous Robots}
\acro{CNN}{Convolutional Neural Network}
\acro{CVaR}{Conditional Value-at-Risk}
\acro{DARPA}{Defense Advanced Research Projects Agency}
\acro{DLT}{Direct Linear Transform}
\acro{EdgeTPU}{Edge Tensor Processing Unit}
\acro{FGA}{Flat Ground Assumption}
\acro{FIRM}{Feedback-based Information Roadmap}
\acro{FOV}{Field of View}
\acro{FPS}{Frames Per Second}
\acro{GCP}{Global Coverage Planning}
\acro{GESTALT}{Grid-based Estimation of Surface Traversability Applied to Local Terrain}
\acro{GICP}{Generalized Iterative Closest Point}
\acro{GO-LLC}{Geometric-Only LiDAR Loop Closures}
\acro{GNSS}{Global Navigation Satellite System}
\acro{GPS}{Global Positioning System}
\acro{GPU}{Graphical Processing Unit}
\acro{GTSAM}{Georgia Tech Smoothing and Mapping}
\acro{HeRO}{Heterogeneous and Resilient Odometry Estimator}
\acro{HFE}{Hierarchical Frontier-based Exploration}
\acro{IMU}{Inertial Measurement Unit}
\acro{iSAM2}{incremental Smoothing and Mapping}
\acro{IRM}{Information RoadMap} (p30,31 dups)
\acro{JPL}{Jet Propulsion Laboratory}
\acro{LAMP}{Large-scale Autonomous Mapping and Positioning}
\acro{LCP}{Local Coverage Planner}
\acro{LIO}{LiDAR-Inertial Odometry}
\acro{LION}{LiDAR-Inertial Observability-aware Navigation for vision-denied environments}
\acro{LO}{Lidar Odometry}
\acro{LOCUS}{Lidar Odometry for Consistent operations in Uncertain Settings}
\acro{MAV}{Micro Aerial Vehicle}
\acro{MDC}{Motion Distortion Correction}
\acro{MER}{Mars Exploration Rover}
\acro{MIKE}{Multi-robot Interaction assistant for unKnown cave Environments (aka `Copilot', `Copilot MIKE')}
\acro{MPC}{Model Predictive Control}
\acro{MSL}{Mars Science Laboratory}
\acro{MSR}{Mars Sample Return}
\acro{NBV}{Next-Best-View}
\acro{NeBula}{Networked Belief-aware Perceptual Autonomy}
\acro{NCDS}{NeBula Communications Deployment System}
\acro{NDB}{NeBula Diagnostics Board}
\acro{NPCC}{NeBula Power and Computing Core}
\acro{NSP}{NeBula Sensor Package}
\acro{PLGRIM}{Probabilistic Local and Global Reasoning on Information roadMaps}
\acro{POMCP}{Partially Observable Monte Carlo Planning}
\acro{POMDP}{Partially Observable Markov Decision Process}
\acro{QoS}{Quality of Service}
\acro{RA-LLC}{Range-Aided LiDAR Loop Closures}
\acro{RC}{Remote Control}
\acro{RF}{Radio Frequency}
\acro{RHC}{Receding Horizon Control}
\acro{RIO}{RaDAR-Inertial Odometry}
\acro{RRT}{Rapidly-exploring Random Trees}
\acro{QP}{Quadratic Programming}
\acro{ROAMS}{Rover Analysis Modeling and Simulation}
\acro{ROS}{Robot Operating System}
\acro{RSSI}{Received signal strength indication}
\acro{SLAM}{Simultaneous Localization and Mapping}
\acro{SLAP}{Simultaneous Localization and Planning}
\acro{SMAP}{Simultaneous Mapping and Planning}
\acro{SOG}{Sum of Gaussians}
\acro{STEP}{Stochastic Traversability Evaluation and Planning}
\acro{STIX}{SubT Integration Exercise}
\acro{TIO}{Thermal-Inertial Odometry}
\acro{TRACE}{Traceable Robotic Activity Composer and Executive}
\acro{UAV}{Unmanned Aerial Vehicle}
\acro{UWB}{Ultra Wide-Band}
\acro{VIO}{Visual-Inertial Odometry}
\acro{VO}{Visual Odometry}
\acro{WIO}{Wheel-Inertial Odometry}
\end{acronym}

\bibliographystyle{IEEEtran} 
\bibliography{main}

\end{document}